\documentclass{article}

\PassOptionsToPackage{square, numbers, compress}{natbib}
\usepackage[final]{neurips_2024}

\usepackage[utf8]{inputenc} %
\usepackage[T1]{fontenc}    %
\usepackage{hyperref}       %
\usepackage{url}            %
\usepackage{booktabs}       %
\usepackage{amsfonts}       %
\usepackage{nicefrac}       %
\usepackage{microtype}      %
\usepackage{xcolor}         %
\usepackage{subfiles}
\usepackage[frozencache,cachedir=.]{minted}
\newminted{python}{%
}

\usepackage[utf8]{inputenc} %
\usepackage[T1]{fontenc}    %
\usepackage{hyperref}       %
\usepackage{url}            %
\usepackage{booktabs}       %
\usepackage{amsfonts}       %
\usepackage{nicefrac}       %
\usepackage{microtype}      %
\usepackage{xcolor}         %
\usepackage{multirow}

\usepackage{graphicx} %
\usepackage{amsmath} %
\usepackage{fancyhdr} %
\usepackage{caption,subcaption}
\usepackage{tikz,graphics,float,epsf}
\usepackage{xcolor}
\usepackage{cancel}
\usepackage{wrapfig}

\usepackage{hyperref}
\hypersetup{colorlinks, citecolor=blue}

\usepackage[english]{babel}
\usepackage{csquotes}
\usepackage{enumitem}

\usepackage{amsfonts} 
\usepackage{natbib}
\usepackage{algorithm}
\usepackage{algpseudocode}
\usepackage{amssymb}
\usepackage{amsthm}
\usepackage{bbold}
\usepackage{amsmath}
\usepackage{mathtools}
\usepackage{xspace}
\usepackage{array}

\DeclareMathOperator*{\argmax}{arg\,max}
\DeclareMathOperator*{\argmin}{arg\,min}

\title{Going Beyond Heuristics by Imposing Policy Improvement as a Constraint}

\author{%
  Chi-Chang Lee$^1$\thanks{ indicates equal contribution. $^1$ National Taiwan University, Taiwan. $^2$ Improbable AI Lab, MIT, USA.} , Zhang-Wei Hong$^2$$^*$ , Pulkit Agrawal$^2$ \\
  Improbable AI Lab\\ 
  Massachusetts Institute of Technology
}

\begin{document}

\maketitle

\begin{abstract}

In many reinforcement learning (RL) applications, augmenting the task rewards with heuristic rewards that encode human priors about how a task should be solved is crucial for achieving desirable performance.  
However, because such heuristics are usually not optimal, much human effort and computational resources are wasted in carefully balancing tasks and heuristic rewards. Theoretically rigorous ways of incorporating heuristics rely on the idea of \textit{policy invariance}, which guarantees that the performance of a policy obtained by maximizing heuristic rewards is the same as the optimal policy with respect to the task reward.
However, in practice, policy invariance doesn't result in policy improvement, and such methods are known to empirically perform poorly. 
We propose a new paradigm to mitigate reward hacking and effectively use heuristics based on the practical goal of maximizing policy improvement instead of policy improvement. 
Our framework, Heuristic Enhanced Policy Optimization (HEPO), effectively leverages heuristics while avoiding the pitfall of prior methods for mitigating reward hacking. 
HEPO achieves superior performance on standard benchmarks with well-engineered reward functions. More surprisingly, HEPO allows policy optimization to achieve good performance even when heuristics are not well-engineered and designed by non-expert humans, showcasing HEPO's ability to reduce human effort in reward design. %
HEPO is a plug-and-play optimization method for leveraging heuristics in reinforcement learning. 
Code is available at \url{https://github.com/Improbable-AI/hepo}.

\end{abstract}

\section{Introduction}
\label{sec:intro}

Reinforcement learning (RL) \citep{sutton2018reinforcement} is a powerful framework for learning policies that can exceed human performance on complex tasks \citep{wurman2022outracing,silver2017mastering,mnih2015human}. Let $\pi_J$ be a policy learned by optimizing the task reward $J$ that indicates whether the agent is successful on the given task. However, when the task reward is sparse or delayed for many steps, its estimation becomes noisy and lacks a strong learning signal, making it ineffective for training policies with RL. Rather than relying solely on a sparse task reward, it is common to design a heuristic reward function composed of multiple terms that provide denser feedback. Heuristic rewards are typically easier to estimate accurately and are designed to align with increases in the true task reward. This facilitates more efficient training and often leads to better performance across a range of applications~\citep{chen2021system,margolis2022rapid,margolis2023walk,kumar21rma,lee2020learning}, as they offer less noisy and more informative learning signals~\cite{ng1999policy,gupta2022unpacking,parkposition}. Let $\pi_H$ denote the policy trained with the heuristic reward. A well-designed heuristic $H$ should guide the agent toward high task performance, satisfying $J(\pi_H) > J(\pi_J)$.

Because heuristics incentivize an agent differently from task rewards $J$, they introduce a bias that can limit the agent's performance or result in \textit{reward hacking}~\cite{russell2002artificial,agrawal2022task}, where an RL algorithm discovers unintended strategies that maximize the heuristic reward $H$ but fail to align with the original task objective $J$. An example of heuristic rewards restricting performance is: consider training a humanoid robot to walk fast. A heuristic reward function that encourages the robot to walk like a human might prevent the discovery of faster walking behaviors that do not resemble human gait.
The most common workaround for going beyond the heuristic rewards and achieving high task performance is combining heuristics and task rewards in a joint optimization objective $J(\pi) + \lambda H(\pi)$, where $\lambda$ controls the trade-off between the task and heuristic rewards. While proper tuning of $\lambda$ can yield strong task performance, finding the right value often requires extensive hyperparameter search, making the process tedious and computationally expensive.

The importance of heuristics in policy optimization has motivated a growing body of work in developing principled methods to use heuristics without their downsides. A majority of the work relies on the principle of \textit{policy invariance} \citep{ng1999policy,devlin2012dynamic,cheng2021heuristic,devlin2014potential,eck2016potential,badnava2023new,forbes2024potential} that constructs heuristic reward functions in a manner that ensures that the policy obtained by optimizing the joint objective $J(\pi) + H(\pi)$ remains unbiased with respect to the policy obtained by optimizing only the task reward $J(\pi)$. However, policy invariance does not imply policy improvement nor guarantee better task performance compared to a policy directly optimizing task $J$ or heuristic rewards $H$. In fact, previous work \citep{cheng2021heuristic} and our results in Section~\ref{subsec:exp:bench} show that methods with policy invariance guarantees often perform worse than simply optimizing the heuristic reward function. 

We posit that using heuristics to achieve policy improvement, rather than the long-standing view in the field of achieving policy invariance, is more aligned with the practical need of achieving high task performance with RL algorithms. 
Policy invariance merely guarantees that the optimization process is unbiased—it does not ensure that the learned policy will outperform the policy trained only with task rewards. However, in practice, when training a new policy, the reason for using heuristics is to improve over the policy learned using only the task rewards. In tasks where the true task reward is sparse or difficult to optimize, the policy trained with heuristic rewards, $\pi_H$, typically achieves higher task performance than the policy trained only with task rewards, but its performance is limited due to potential misalignment in the objectives as described above. Our goal is to develop a policy optimization framework that can leverage heuristics as needed, but performs better than the policy trained with heuristics. 
We formalize our intuition by training a policy $\pi$ using both task and heuristic rewards while ensuring its task performance is at least as good as that of the heuristic policy $\pi_H$, i.e., $J(\pi) \geq J(\pi_H)$. We argue that this constraint provides a more practical guideline based on policy improvement for mitigating reward hacking instead of the widely used principle of policy invariance.

We call the resulting algorithm \textbf{Heuristic Enhanced Policy Optimization (HEPO)} -- a general policy optimization method that goes beyond the performance of policies trained with any given set of heuristic rewards. 
HEPO builds on ideas from conservative policy iteration \citep{kakade2002approximately} to guarantee policy improvement by enforcing the constraint $J(\pi) \geq J(\pi_H)$. 
HEPO concurrently trains both $\pi$ and $\pi_H$, and the policy improvement scheme can be implemented using standard deep RL algorithms~\citep{schulman2017proximal, schulman2015trust, haarnoja2018soft, mnih2015human}. The objective is:
\begin{align*}
    \max_{\pi} J(\pi) + H(\pi) \quad \text{subject to} \quad J(\pi) \geq J(\pi_H)
\end{align*}
By enforcing this constraint, HEPO ensures that the learned policy performs at least as well as the heuristic baseline, mitigating reward hacking. Unlike traditional approaches that require manually tuning the trade-off ($\lambda$) between task and heuristic rewards (e.g., optimizing $J(\pi) + \lambda H(\pi)$), HEPO avoids the need for such balancing by explicitly preserving task performance.

Results in Section~\ref{subsec:exp:bench} show that HEPO outperforms PPO at utilizing well-engineered heuristic rewards provided as part of the standard benchmark of IssacGym\citep{makoviychuk2021isaac}. Furthermore, to test if HEPO can reduce human effort in designing reward functions, we ran a human study where we invited non-experts to design reward functions for solving simulated robotic tasks. The rewards designed by non-experts are worse than experts, and PPO~\cite{schulman2017proximal} optimization often fails. However, HEPO successfully uses heuristics as needed and prevents over-optimization of badly designed heuristics. While HEPO can't design better heuristics, its superior use of heuristics can lead to good performance even when the provided heuristics are inferior (e.g., from non-experts). Results in Section~\ref{subsec:exp:application} show that with HEPO, even non-expert heuristic rewards can attain expert-level task performance.

\section{Preliminaries: Reinforcement Learning with Heuristic}
\label{sec:prelim}

\paragraph{Reinforcement Learning (RL):}
RL is a popular paradigm for solving sequential decision-making problems~\citep{sutton2018reinforcement} modeled as an interaction between an agent and an unknown environment~\cite{sutton2018reinforcement}.
The agent aims to improve its performance through repeated interactions with the environment. 
At each round of interaction, the agent starts from the environment's initial state $s_0$. At each timestep $t$, the agent observes the state $s_t$, takes an action $a_t \sim \pi(.|s_t)$ according to the current policy $\pi$, receives a \textit{task} reward $r_t = r(s_t, a_t)$, and transitions to a next state $s_{t+1}$ until reaching terminal states, after which a new trajectory is initialized from $s_0$, and the cycle repeats. 
The agent's goal is to learn a policy $\pi$ that maximizes the expected return $J(\pi)$ in a trajectory as below:
\begin{align}
    J(\pi) = \mathbb{E}_{\pi}[\sum^{\infty}_{t=0} \gamma^{t} r(s_t, a_t)],
\end{align}
where $\gamma$ denotes a discount factor~\citep{sutton2018reinforcement} and $\mathbb{E}_{\pi}\left[.\right]$ denotes taking expectation over the trajectories sampled by $\pi$.
$J$ is the true \textit{task} objective indicating the policy's performance on the task.

\paragraph{RL with Heuristic:}
In many tasks, learning a policy to maximize the true objective $J$ is challenging because rewards may be sparse or delayed. 
This lack of feedback makes policy optimization difficult for RL algorithms. 
To address this, practitioners often use a heuristic reward function $h$ with denser reward signals to facilitate optimization, aiming to learn a policy that performs better in $J$. 
The policy trained to maximize the expected return of heuristic rewards is called the \textit{heuristic} policy $\pi_H$. 
The expected return of heuristic rewards, termed the \textit{heuristic} objective $H$, is defined as:
\begin{align}
    H(\pi_H) = \mathbb{E}_{\pi_H} \left[\sum_{t=0}^{\infty} \gamma^t h(s_t, a_t)\right],
\end{align}
where $h(s_t, a_t)$ is the heuristic reward at timestep $t$ for state $s_t$ and action $a_t$.

\section{Method: Improving Heuristic Policy via Constrained Optimization}
\label{sec:method}

\paragraph{Problem statement:}
Optimizing both task $J$ and heuristic $H$ objectives jointly could lead to better task performance than training solely with $J$ or $H${, but needs careful tuning on the weight coefficient $\lambda$ in $\max_\pi J(\pi) + \lambda H(\pi)$. Without careful tuning, the policy $\pi$ may learn to exploit heuristic rewards $H$ and compromise task rewards $J$. Our aim is to mitigate the need to manually tune $\lambda$ and obtain performant policies that successfully leverage heuristic rewards.

\paragraph{Key insight - Leveraging Heuristic with Constraint:} 
We aim to use the heuristic objective $H$ for training only when it improves task performance $J$ and ignore it otherwise. Rather than manually tuning the weight coefficient $\lambda$ to balance both rewards, we introduce a key insight: impose a \textit{policy improvement} constraint (i.e., $J(\pi) \geq J(\pi_H)$) during training. This prevents RL algorithms from exploiting heuristic rewards $H$ at the expense of task rewards $J$. To achieve this goal, we introduce the following constrained optimization objective:
\begin{align}
\label{eq:constrained_obj}
\max_{\pi} J(\pi) + H(\pi) \text{ subject to } J(\pi) \geq J(\pi_{{H}}).
\end{align}
This constrained objective (Equation~\ref{eq:constrained_obj}) results in an improved policy $\pi$ over the heuristic policy $\pi_H$, leading us to call this framework \textit{Heuristic-Enhanced Policy Optimization (HEPO)}. 
A practical algorithm to optimize this objective is presented in Section~\ref{subsec:alg}, and its implementation on a widely-used RL algorithm~\citep{schulman2017proximal} is detailed in Section~\ref{subsec:impl}.

\subsection{Algorithm: Heuristic-Enhanced Policy Optimization (HEPO)}
\label{subsec:alg}
Finding feasible solutions for the constrained optimization problem in Equation~\ref{eq:constrained_obj} is challenging when the objective $J(\pi)$ and $H(\pi)$ are undifferentiable. One practical approach is to convert it into the following unconstrained min-max optimization problem using Lagrangian duality:
\begin{align}
\label{eq:unconstrained_obj}
\min_{\alpha \geq 0} \max_{\pi} \mathcal{L}(\pi, \alpha), \text{ where } \mathcal{L}(\pi, \alpha) := J(\pi) + H(\pi) + \alpha \left( J(\pi) - J(\pi_{\text{H}}) \right), 
\end{align}
where the Lagrangian multiplier is $\alpha \in \mathbb{R}^{+}$. 
We can optimize the policy $\pi$ and the multiplier $\alpha$ for this min-max problem by a gradient descent-ascent strategy, alternating between optimizing $\pi$ and $\alpha$.

\textbf{Enhanced policy $\pi$:}
The optimization objective for the policy $\pi$ can be obtained by rearranging Equation~\ref{eq:unconstrained_obj} as follows:
\begin{align}
\begin{split}
\label{eq:policy_obj}
\max_{\pi} ~&~  (1+\alpha) J(\pi) + H(\pi), \\ 
\text{ where } &  (1+\alpha) J(\pi) + H(\pi) = \mathbb{E}_{\pi} \left[ \sum^{\infty}_{t=0} \gamma^{t} \big( (1+\alpha) r(s_t, a_t)) + h(s_t, a_t)\big)\right]. 
\end{split}
\end{align}
This represents an unconstrained regular RL objective with the modified reward at each step as $(1+\alpha) r(s_t, a_t)+h(s_t, a_t)$, which can be optimized using any off-the-shelf deep RL algorithm.
In this modified reward, the task reward $r(s_t, a_t)$ is weighted by the Lagrangian multiplier $\alpha$, reflecting the potential variation in the task reward's importance during training as $\alpha$ evolves.
The interaction between the update of the Lagrangian multiplier and the policy will be elaborated upon next.

\paragraph{Lagrangian Multiplier $\alpha$:} The Lagrangian multiplier $\alpha$ is optimized for Equation~\ref{eq:unconstrained_obj} by stochastic gradient descent, with the gradient defined as:
\begin{align}
\label{eq:alpha_grad}
    \nabla_\alpha \mathcal{L}(\pi, \alpha) = J(\pi) - J(\pi_{\text{H}}).
\end{align}
Notably, $\nabla_\alpha \mathcal{L}(\pi, \alpha)$ is exactly the performance gain of the policy $\pi$ over the heuristic policy $\pi_H$ on the task objective $J$. 
When $J(\pi) > J(\pi_H)$ it implies that $\nabla_\alpha \mathcal{L}(\pi, \alpha) > 0$ which decreases the  Lagrangian multiplier $\alpha$. 
As $\alpha$ represents the weight of the task reward in Equation~\ref{eq:policy_obj}, it indicates that when $\pi$ outperforms $\pi_H$, the importance of the task reward diminishes because $\pi$ already achieves superior performance compared to the heuristic policy $\pi_H$ regarding the task objective $J$. Conversely, when $J(\pi) < J(\pi_H)$, $\alpha$ increases, thereby emphasizing the importance of task rewards in optimization. The update procedure for the Lagrangian multiplier $\alpha$ offers an adaptive reconciliation between the heuristic reward $h$ and the task reward $r$.

\subsection{Implementation}
\label{subsec:impl}
We present a practical approach to optimize the min-max problem in Equation~\ref{eq:unconstrained_obj} using Proximal Policy Optimization (PPO)~\citep{schulman2017proximal}. 
We selected PPO because it is widely used in robotic applications involving heuristic rewards, although our HEPO framework is not restricted to PPO. 
The standard PPO implementation involves iterative stochastic gradient descent updates over numerous iterations, alternating between collecting trajectories with policies and updating those policies. 
We outline the optimization process for each iteration and provide a summary of our implementation in Algorithm~\ref{alg:hepo}.

\textbf{Training policies $\pi$ and $\pi_H$:}
Instead of pre-training the heuristic policy $\pi_H$, which requires additional data and reduces data efficiency, we concurrently train both the enhanced policy $\pi$ and the heuristic policy $\pi_H$, allowing them to share data.
For each iteration $i$, we gather trajectories $\tau$ and $\tau_H$ using the enhanced policy $\pi^i$ and the heuristic policy $\pi^i_H$, respectively. 
Following PPO's implementation, we compute the advantages $A^{\pi^i}_r(s_t, a_t)$, $A^{\pi^i_H}_r(s_t, a_t)$, $A^{\pi^i}_h(s_t, a_t)$, and $A^{\pi^i_H}_h(s_t, a_t)$ for the task reward $r$ and heuristic reward $h$ with respect to $\pi^i$ and $\pi^i_H$. 
We then weight the advantage with the action probability ratio between the new policies being optimized (i.e., $\pi^{i+1}$ and $\pi^{i+1}_{H}$) and the policies collecting the trajectories (i.e., $\pi^{i}$ or $\pi^i_{H}$). 
Finally, we optimize the policies at the next iteration $i+1$ for the objectives in Equations~\ref{eq:impl_pi} and~\ref{eq:impl_pi_h}:
\begin{align}
\label{eq:impl_pi}
    \pi^{i+1}\leftarrow \argmax_{\pi} ~& \mathbb{E}_{\tau \sim \pi^{i}}\left[\dfrac{\pi(a_t|s_t)}{\pi^{i}(a_t|s_t)}\left((1+\alpha) A_r^{\pi^{i}}(s_t, a_t)+A_h^{\pi^{i}}(s_t, a_t)\right)\right] + \\  
    & \mathbb{E}_{\tau_H \sim \pi^{i}_H}\left[\dfrac{\pi(a_t|s_t)}{\pi^{i}_H(a_t|s_t)}\left((1+\alpha)  A_r^{\pi^{i}_H}(s_t, a_t) + A_h^{\pi^{i}_H}(s_t, a_t) + \right)\right] \nonumber & \text{(Enhanced policy)}\\
\label{eq:impl_pi_h}
    \pi^{i+1}_H\leftarrow\argmax_{\pi} ~& \mathbb{E}_{\tau_H \sim \pi^{i}_H}\left[\dfrac{\pi(a_t|s_t)}{\pi^{i}_H(a_t|s_t)}A_h^{\pi^{i}}(s_t, a_t) \right] + \\ &\mathbb{E}_{\tau \sim \pi^{i}}\left[\dfrac{\pi(a_t|s_t)}{\pi^{i}(a_t|s_t)}A_h^{\pi^{i}_H}(s_t, a_t) \right] \nonumber& \text{(Heuristic policy)}.
\end{align}
Maximizing the advantages will result in a policy that maximizes the expected return for a chosen reward function, as demonstrated in PPO~\citep{schulman2017proximal}. 
This enables us to maximize the objective $J$ and $H$. 
We estimate the advantages $A^{\pi^i}_r(s_t, a_t)$ and $A^{\pi^i}_h(s_t, a_t)$ (or $A^{\pi^i_H}_r(s_t, a_t)$ and $A^{\pi^i_H}_h(s_t, a_t)$) using the standard PPO implementation with different reward functions. Further details are provided in  Appendix~\ref{app:derivation}.

Although PPO is an on-policy algorithm, the use of off-policy importance ratio correction (i.e., the action probability ratios between two policies) allows us to use states and actions generated by another policy. 
This enables us to train $\pi$ using data from $\pi_H$ and vice versa. 
Both policies $\pi$ and $\pi_H$ are trained using the same data but with different reward functions. 
Note that collecting trajectories from both policies does not require more data than the standard PPO implementation. 
We collect half the trajectories with each policy, $\pi$ and $\pi_H$, for a total of $B$ trajectories (see Algorithm~\ref{alg:hepo}). 
Then, we update both $\pi$ and $\pi_H$ using all $B$ trajectories.

\textbf{Optimizing the Lagrangian multiplier $\alpha$:} 
To update the Lagrangian multiplier $\alpha$, we need to compute the gradient in Equation~\ref{eq:alpha_grad}, which corresponds to the performance gain of the enhanced policy $\pi$ over the heuristic policy $\pi_H$ on the task objective $J$.
Utilizing the performance difference lemma~\citep{kakade2002approximately, schulman2015trust}, we relate this improvement to the expected advantages over trajectories sampled by the enhanced policy $\pi$ as $J(\pi) - J(\pi_{H}) = \mathbb{E}_{\pi} \left[ A^{\pi_{H}}r(s_t, a_t) \right]$. 
However, this approach only utilizes half of the trajectories at each iteration since it exclusively relies on trajectories from the enhanced policy $\pi$. To leverage trajectories from both policies, we also consider the performance gain in the reverse direction as $-(J(\pi_H) - J(\pi)) = - \mathbb{E}_{\pi_H} \left[ A^{\pi}r(s_t, a_t) \right]$. 
Consequently, we can estimate the gradient of $\alpha$ using trajectories from both policies, as illustrated below:
\begin{align}
\label{eq:impl_alpha}
\nabla_\alpha \mathcal{L}(\pi, \alpha) &= J(\pi) - J(\pi_{H}) = \mathbb{E}_{\pi} \left[ A^{\pi_{H}}_r(s_t, a_t) \right] \\
&= -(J(\pi_H) - J(\pi)) = -\mathbb{E}_{\pi_H} \left[ A^{\pi}_r(s_t, a_t) \right].
\end{align}
At each iteration $i$, we estimate the gradient of $\alpha$ using the advantage $A^{\pi_{H}}_r(s_t, a_t)$ and $A^{\pi}_r(s_t, a_t)$ on the trajectories sampled from both $\pi^i$ and $\pi^i_H$, and update $\alpha$ with stochastic gradient descent as follows:
\begin{align}
\label{eq:impl_alpha_update}
\alpha \leftarrow \alpha - \dfrac{\eta}{2} \left(\mathbb{E}_{\tau \sim \pi^i} \left[ A^{\pi^{i}_{H}}_r(s_t, a_t) \right] - \mathbb{E}_{\tau \sim \pi^i_H} \left[ A^{\pi^{i}}_r(s_t, a_t) \right]\right),
\end{align}
where $\eta \in \mathbb{R}^+$ is the step size. The expected advantage in Equation~\ref{eq:impl_alpha_update} are estimated using the generalized advantage estimator (GAE)~\citep{schulman2015high}.

\begin{algorithm}[htp!]
\caption{Heuristic-Enhanced Policy Optimization (HEPO)}
\label{alg:hepo}
\begin{algorithmic}[1]

\State \textbf{Input:} Number of trajectories per iteration $B$
\State Initialize policy $\pi^0$, the heuristic policy $\pi^0_H$, and the Lagrangian multiplier $\alpha$

\For{$i = 0 \cdots $} \Comment{$i$ denotes iteration index}
    \State Rollout $B / 2$ trajectories $\tau$ by $\pi^i$
    \State Rollout $B / 2$ trajectories $\tau_H$ by $\pi^i_{H}$
    \State $\pi^{i+1} \longleftarrow$ Train the policy $\pi^i$ for optimizing Equation~\ref{eq:impl_pi} using both $\tau$ and $\tau_H$
    \State $\pi^{i+1}_H \longleftarrow$ Train the policy $\pi^i_H$ for optimizing Equation~\ref{eq:impl_pi_h} using both $\tau$ and $\tau_H$
    \State $\alpha \longleftarrow$ Update the Lagrangian multiplier $\alpha$ by gradient descent (Equation~\ref{eq:impl_alpha}) using $\tau$ and $\tau_H$
\EndFor

\end{algorithmic}
\end{algorithm}

\subsection{Connection to Extrinsic-Intrinsic Policy Optimization~\cite{chen2022redeeming}}
\label{subsec:hepo_vs_eipo}
Closely related to our HEPO framework, \citet{chen2022redeeming} proposes Extrinsic-Intrinsic Policy Optimization (EIPO), which trains a policy to maximize both task rewards and exploration bonuses~\citep{burda2018exploration} subject to the constraint that the learned policy $\pi$ must outperform the \textit{task} policy $\pi_J$ trained solely on task rewards. HEPO and EIPO differ in their objective functions and implementation of the constrained optimization problem. Additional information can be found in the Appendix, covering the objective formulation (Appendix~\ref{app:derivation}), implementation details (Appendix~\ref{app:trick}), and detailed pseudocode (Appendix~\ref{alg:minimmax}).

Exploration bonuses~\citep{burda2018exploration} can be viewed as heuristic rewards. 
The main difference between HEPO and EIPO's optimization objectives lies in constraint design. 
Both frameworks require the learned policy $\pi$ to outperform a reference policy $\pi_{\text{ref}}$ (i.e., $J(\pi) \geq J(\pi_{\text{ref}})$) but use a different reference policy. 
EIPO uses the task policy $\pi_J$ as the reference policy $\pi_{\text{ref}}$ because they aim for asymptotic optimality in task rewards. 
If the constraint is satisfied with $\pi_J$ being the optimal policy for task rewards, the learned policy $\pi$ will also be optimal for task rewards. 
In contrast, HEPO uses the heuristic policy $\pi_H$ trained solely on heuristic rewards since HEPO aims to improve upon it.

HEPO simplifies the implementation. 
Both HEPO and EIPO train two policies with shared data, but EIPO alternates the policy used for trajectory collection each iteration and has a complex switching rule, which introduces more hyperparameters. 
HEPO collects trajectories using both policies together at each iteration, simplifying implementation and avoiding extra hyperparameters.

\section{Experiments}
\label{sec:exp}
We evaluate whether HEPO enhances the performance of RL algorithms in maximizing task rewards while training with heuristic rewards.
We conduct experiments on 9 tasks from IsaacGym (\textsc{Isaac})~\citep{makoviychuk2021isaac} and 20 tasks from the Bidexterous Manipulation (\textsc{Bi-Dex}) benchmark~\citep{chen2022towards}.
These tasks rely on heavily engineered reward functions for training RL algorithms. Each task has a task reward function $r$ that defines the task objective $J$ to be maximized, and a heuristic reward function $h$ that defines the heuristic objective $H$, provided in the benchmarks to facilitate the optimization of task objectives $J$.
We implement HEPO based on PPO~\citep{schulman2017proximal} and compare it with the following baselines:
\begin{itemize}[leftmargin=*]
    \item \textbf{H-only (heuristic only)}: This is the standard PPO baseline provided in \textsc{Isaac}. The policy is trained solely using the heuristic reward: $\max_{\pi} H(\pi)$. The heuristic reward functions in \textsc{Isaac} and \textsc{Bi-Dex} are designed to help RL algorithms maximize the task objective $J$. This baseline is crucial to determine if an algorithm can surpass a policy trained with highly engineered heuristic rewards.
    \item \textbf{J-only (task only)}: 
    The policy is trained using only the task reward: $\max_{\pi} J(\pi)$. This baseline demonstrates the performance achievable without heuristics. Ideally, algorithms that incorporate heuristics should outperform this baseline.
    \item \textbf{J+H (mixture of task and heuristic)}: 
    The policy is trained using a mixture of task and heuristic rewards: $\max_{\pi} J(\pi) + \lambda H(\pi)$, with $\lambda$ balancing the two rewards. As~\citep{chen2022redeeming} shows, proper tuning of $\lambda$ can enhance task performance by balancing both training objectives.
    \item \textbf{Potential-based Reward Shaping (PBRS)~\citep{ng1999policy}}: 
    The policy is trained to maximize $\mathbb{E}_{\pi} [\sum_{t=0}^{\infty}\gamma^t r_t + \gamma h_{t+1} - h_t]$, where $r_t$ and $h_t$ are the task and heuristic rewards at timestep $t$. PBRS guarantees that the optimal policy is invariant to the task reward function. We include it as a baseline to examine if these theoretical guarantees hold in practice.
    \item \textbf{HuRL~\citep{cheng2021heuristic}}: 
    The policy is trained to maximize $\mathbb{E}_{\pi} [\sum_{t=0}^{\infty}\gamma^t r_t + (1 - \beta_i)\gamma h_{t+1}]$, where $\beta_i$ is a coefficient updated at each iteration to balance heuristic rewards during different training stages. The scheduling mechanism is detailed in~\citep{cheng2021heuristic} and our {source code} provided in the {Supplementary Material}.
    \item \textbf{EIPO~\citep{chen2022redeeming}}: The policy is trained using the constrained objective: $\max_\pi J(\pi) + H(\pi) \quad \text{s.t.} \quad J(\pi) \geq J(\pi_J)$, where $\pi_J$ is the policy trained with task rewards only. EIPO is similar to HEPO but differs in formulation and implementation, as detailed in Section~\ref{subsec:exp:ablation}.
\end{itemize}
Each method is trained for $5$ random seeds and implemented based on the open-sourced implementation~\citep{rl-games2021}, where the detailed training hyperparameters can be found in Appendix~\ref{app:sec:training}.

\textbf{Metrics:}
Based on the task success criteria in \textsc{Isaac} and \textsc{Bi-Dex}, we consider two types of task reward functions $r$: (i) Progressing (for locomotion or helicopter robots) and (ii) Goal-reaching (for manipulation).
In progressing tasks, robots aim to maximize their traveling distance or velocity from an initial point to a destination.
Thus, movement progress is defined as the task reward.
In goal-reaching tasks, robots aim to complete assigned goals by reaching specific goal states.
Here, task rewards are binary, with a value of $1$ indicating successful attainment of the goal and $0$ otherwise.
Detailed descriptions of our task objectives and total reward definitions are provided in {Appendix~\ref{appendix: task_desc}}.

\subsection{Benchmark results}
\label{subsec:exp:bench}
\textbf{Setup:} We aim to determine if HEPO achieves higher task returns and improves upon the policy trained with only heuristic rewards (H-only) in the majority of tasks.
In this experiment, we use the heuristic reward functions from the \textsc{Isaac} and \textsc{Bi-Dex} benchmarks.
To measure performance improvement over the heuristic policy, we normalize the return of each algorithm \(X\) using the formula \((J_X - J_\text{random}) / (J_\text{H-only} - J_\text{random})\), where \(J_X\), \(J_\text{H-only}\), and \(J_\text{random}\) denote the task returns of algorithm \(X\), the heuristic policy, and the random policy, respectively.
In Figure~\ref{fig:all_iqm_combined}, we present the interquartile mean (IQM) of the normalized return and the probability of improvement for each method across 29 tasks, following \cite{agarwal2021deep}. IQM, also known as the 25\% trimmed mean, is a robust estimate against outliers. It discards the bottom and top 25\% of runs and calculates the mean score of the remaining 50\%. The probability of improvement measures whether an algorithm performs better than another, regardless of the margin of improvement. Both approaches prevent outliers from dominating the performance estimate.

\textbf{Results:} The results in Figure~\ref{fig:all_iqm_combined} indicate that policies trained with task rewards only (J-only) generally perform worse than those trained with heuristics, both in terms of IQM of normalized return and probability of improvement.
PBRS does not improve upon J-only, demonstrating that the optimal policy invariance guarantee rarely holds in practice. Both EIPO and HuRL outperform J-only but do not surpass H-only, demonstrating that neither approach can improve upon the heuristic policy.
Policies trained with both task and heuristic rewards (J+H) perform slightly worse than those trained with heuristics only (H-only), possibly because the weight coefficient balancing both rewards is too task-sensitive to work across all tasks.
HEPO, however, outperforms all other methods in both IQM of normalized returns and shows a probability of improvement over the heuristic policy greater than 50\%, indicating statistically significant improvements as suggested by \citet{agarwal2021deep}.
Complete learning curves are presented in the Appendix~\ref{app:all_curve}. Additional results on various benchmarks and RL algorithms are provided in Appendix~\ref{app:additional_results}, demonstrating that HEPO is effective in hard-exploration tasks using exploration bonuses \citep{burda2018exploration} and with RL algorithms beyond PPO.

\begin{figure}[t!]
    \centering
    \begin{subfigure}[b]{0.9\textwidth}
        \centering
        \includegraphics[width=\textwidth]{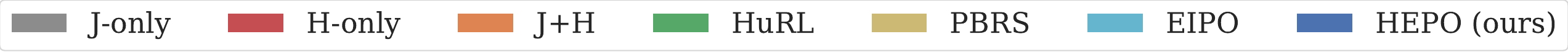}
        \label{fig:all_legend}
    \end{subfigure}%
    \hfill
    \begin{subfigure}[b]{0.5\textwidth}
        \centering
        \includegraphics[width=\textwidth]{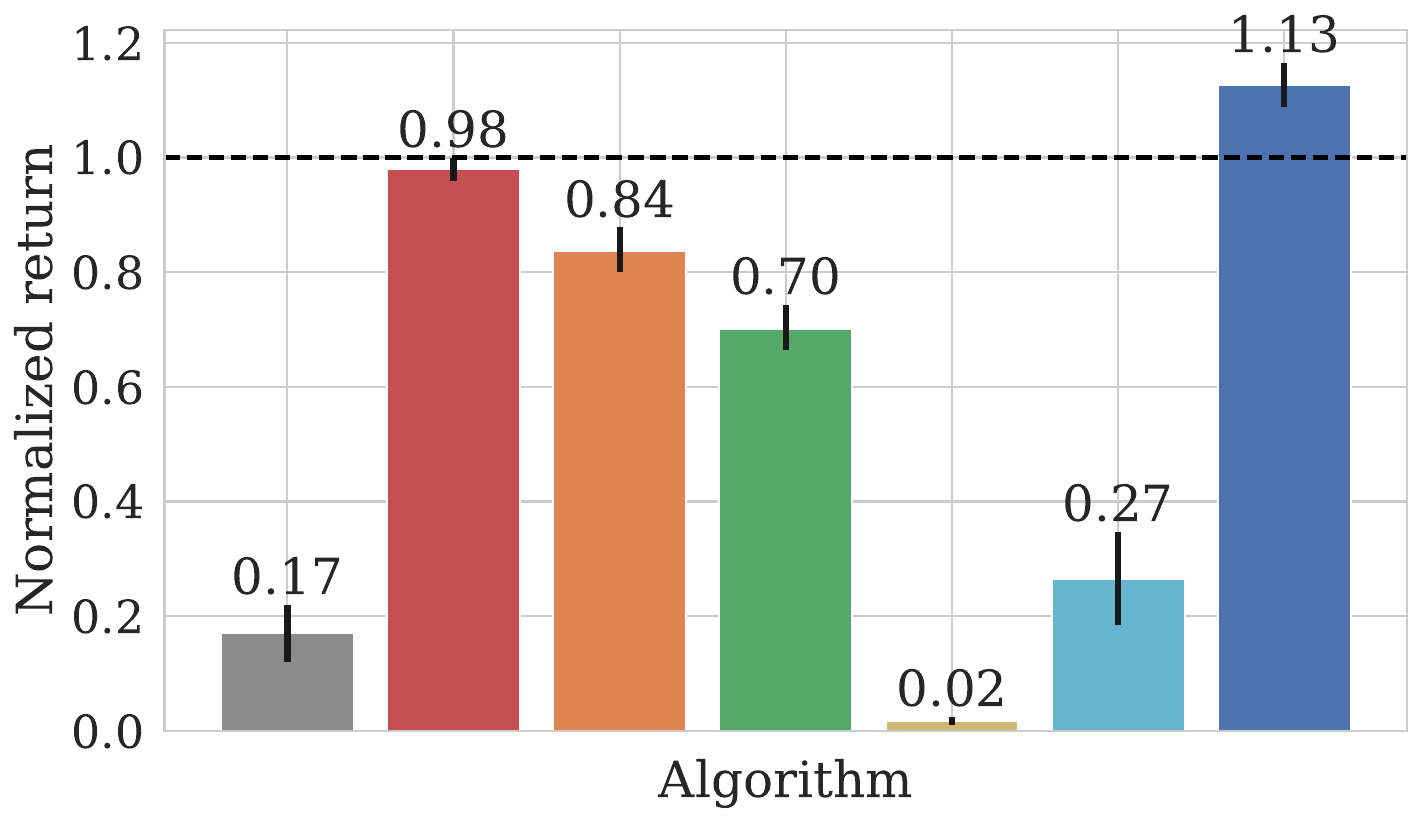}
        \caption{IQM over \textsc{Isaac} + \textsc{Bi-Dex}}
        \label{fig:all_iqm_bar}
    \end{subfigure}%
    \begin{subfigure}[b]{0.5\textwidth}
        \centering
        \includegraphics[width=\textwidth]{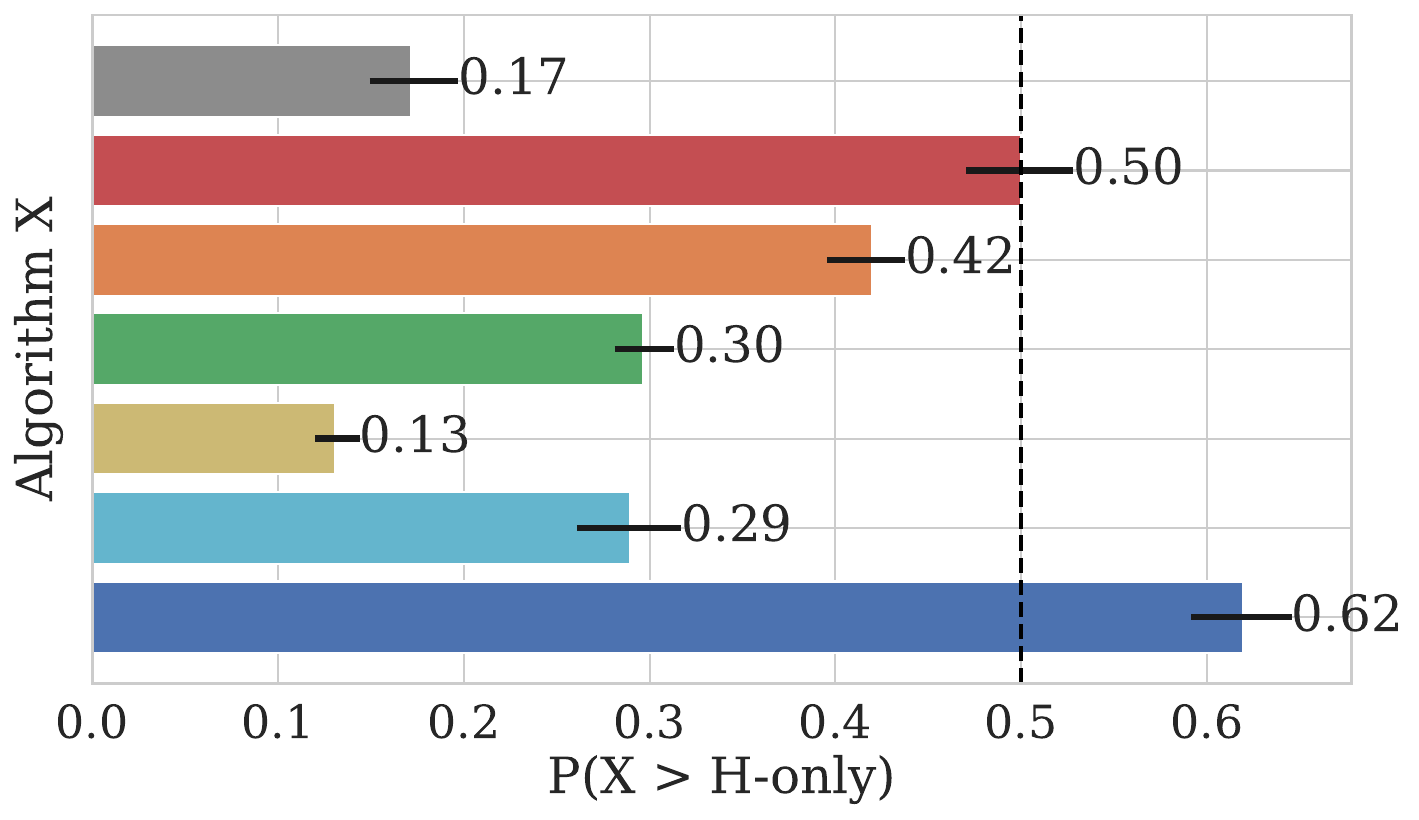}
        \caption{Probability of Improvement over \textsc{Isaac} + \textsc{Bi-Dex}}
        \label{fig:all_iqm_pi}
    \end{subfigure}
    \caption{\textbf{(a)} The vertical axis represents the interquartile mean (IQM)~\citep{agarwal2021deep} of normalized task return across 29 tasks. HEPO outperforms the policy trained solely with a heavily engineered heuristic reward function (H-only) and other methods, demonstrating that HEPO makes better use of heuristic rewards for learning. 
    \textbf{(b)} The horizontal axis shows the probability that algorithm X outperforms the policy trained solely with heuristic rewards (H-only). HEPO achieves a 62\% probability of improvement over the heuristic policy on average, with the lower bound of the confidence interval above 50\%, indicating statistically significant improvements over the heuristic policy.}
    \label{fig:all_iqm_combined}
\end{figure}

\subsection{Can HEPO achieve good performance using reward functions designed by non-experts?}
\label{subsec:exp:application}

\textbf{Setup:} 
We evaluate HEPO's performance when trained with heuristic reward functions designed by non-expert human participants on the same tasks as in the \textsc{Issac} benchmark.  Unlike the highly engineered reward functions in \textsc{Isaac}, participants were asked to design their heuristic reward functions within a short time frame. This approach assesses the algorithm's effectiveness when trained with less engineered heuristic rewards.
Participants were asked to iterate on their reward design by writing the reward function, training a policy with a given RL algorithm, reviewing the videos and learning curves, refining the reward, and repeating the process. 
We selected the \textit{FrankaCabinet} task for this study because its original heuristic reward function is heavily engineered with many terms. This task requires a robot arm to open a cabinet. We recruited twelve graduate students with varying machine learning and robotics proficiency and divided them into two groups. One group used HEPO to iterate on heuristic rewards, while the other used PPO.
This approach ensures that the designed heuristic reward functions are not specialized for one algorithm and ineffective for another.
Each participant was instructed to edit the heuristic reward function to help RL algorithms maximize the task return. We used the same task reward metric in Section~\ref{subsec:exp:bench}. 
We then used the final versions of their heuristic reward functions to train HEPO and PPO and computed the normalized return of the policies using the scheme outlined in Section~\ref{subsec:exp:bench} and using the performance of policy trained with the original heuristic reward function as the baseline for normalization. 

\begin{wraptable}{r}{0.5\textwidth}    
    \footnotesize
    \centering
    \begin{tabular}{lcc}
        \toprule
        \textbf{Algo. X}    & \textbf{IQM}  & \textbf{P(X > H-only)} \\
        \midrule
        H-only & 0.44 (0.37, 0.52) & 0.50 (0.49, 0.51) \\
        HuRL   & 0.00 (0.00, 0.00) & 0.33 (0.32, 0.33) \\
        PBRS   & 0.00 (0.00, 0.00) & 0.22 (0.21, 0.22) \\
        HEPO   & \textbf{0.94 (0.85, 1.03)} & \textbf{0.73 (0.73, 0.74)} \\
        \bottomrule
    \end{tabular}
    \caption{IQM and probability of improvement (PI) over H-only (P(X > H-only)) with 95\% confidence intervals across 12 heuristic reward functions (Section~\ref{subsec:exp:application}). H-only uses PPO with heuristic rewards. HEPO achieves higher normalized returns and a statistically significant PI greater than 0.5, indicating it significantly outperforms H-only.}    
    \label{table:human_agg}
\end{wraptable}

\textbf{Quantitative results:} Figure ~\ref{fig:human_study_bar} shows that HEPO achieves significantly higher (lower confidence bound above the baselines' upper confidence bound) average normalized returns than PPO trained only on heuristic rewards (PPO (H-only)) in 9 out of 12 heuristic reward functions. Additionally, Table ~\ref{table:human_agg} indicates that across all heuristic functions, HEPO achieves a higher interquartile mean (IQM) of normalized returns and has a statistically significant probability of outperforming PPO (H-only) with lower confidence bound greater than $0.5$. 
This suggests that even when trained with poorly designed heuristic reward functions, HEPO performs better than PPO (H-only). Notably, PPO (H-only) that is trained with $H2$, $H5$, and $H6$ achieves normalized returns below $1$, while HEPO achieves returns greater than or close to $1$. Since returns are normalized using the performance of the PPO policy trained with the well-designed heuristic reward function in \textsc{Isaac}, a return below $1.0$ indicates a performance drop for PPO (H-only) when using potentially ill-designed heuristic rewards. In contrast, HEPO can improve upon policies trained with carefully engineered heuristic reward functions, even when trained with possibly ill-designed heuristic reward functions.

\textbf{Qualitative observation:} We aim to understand why PPO's performance declines when trained with heuristic reward functions $H2$, $H5$, and $H6$. These functions are similar to the original heuristic reward in \textit{FrankaCabinet}, but with different weights for each term. For example, in $H5$, the weight of action penalty is $1$, whereas in the original heuristic reward function it is $7.5$. This suggests that HEPO might handle poorly scaled heuristic reward terms better than PPO, which is sensitive to these weights. The heuristic reward functions $H12$ and $H9$ had an incorrect sign for the distance component, which caused the policy to be rewarded for moving away from the cabinet instead of toward it, making the learning task more challenging.

\begin{figure}[t!]
    \centering
    \includegraphics[width=\textwidth]{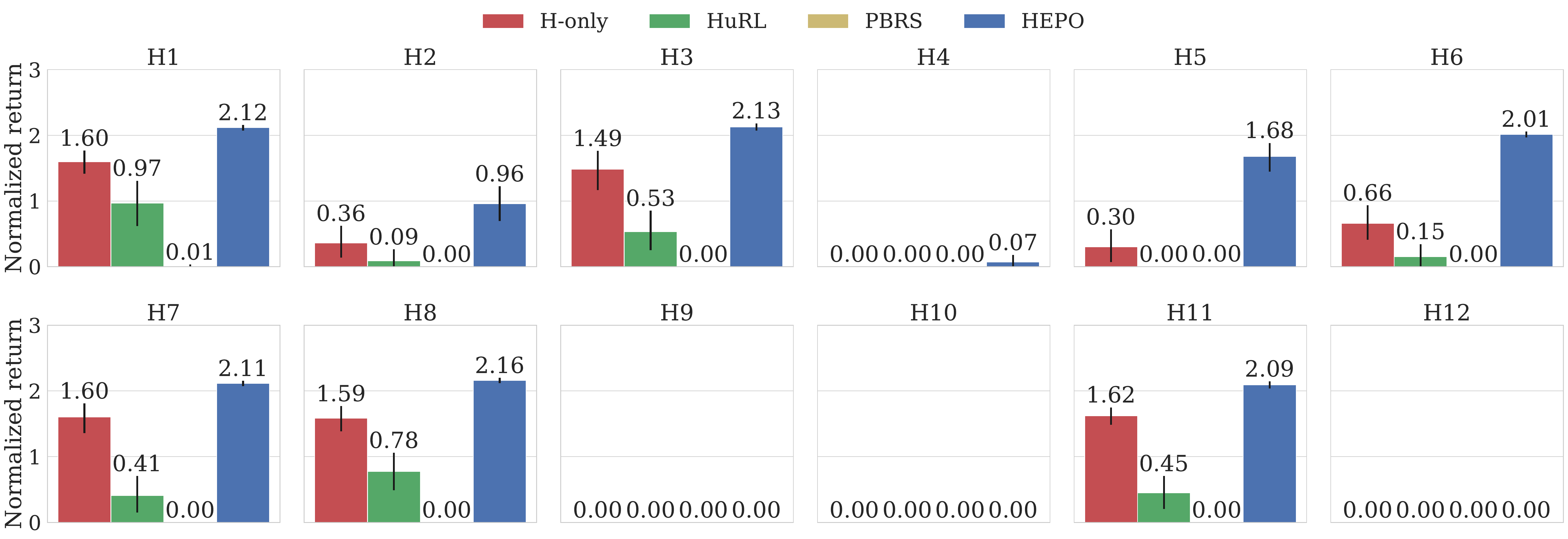}
    \caption{Normalized task return of PPO (H-only) and HEPO that are trained with heuristic reward function $H1$ to $H12$ designed by human subjects in the real world reward design condition. HEPO achieves higher task return than PPO (H-only) in 9 out of 12 tasks. This shows HEPO is robust to possibly ill-designed heuristic reward functions and can leverage them to improve performance. 
    }
    \label{fig:human_study_bar}
\end{figure}

\subsection{Ablation Studies}
\label{subsec:exp:ablation}
Expanding on the discussion of relation to relevant work EIPO \citep{chen2022redeeming} in Section~\ref{subsec:hepo_vs_eipo}, our goal is to examine the implementation choices of HEPO and illustrate the efficacy of each modification in this section. HEPO differs from EIPO primarily in two aspects: (1) the selection of a reference policy $\pi_{\text{ref}}$ in the constraint $J(\pi) \geq J(\pi_{\text{ref}})$, and (2) the strategy for utilizing policies to gather trajectories. Both studies are conducted on standard locomotion and manipulation tasks, such as \textit{Ant}, \textit{FrankaCabinet}, and \textit{AllegroHand}. In addition, we provide further studies on the sensitivity to hyperparameters in Appendix~\ref{subsec:exp:sensitivity}.

\textbf{Selection of reference policy in constraint:}
HEPO and EIPO both enforce a performance improvement constraint $J(\pi) \geq J(\pi_{\text{ref}})$ during training. HEPO uses a heuristic policy $\pi_H$ as the reference ($\pi_\text{ref} = \text{H-only}$), while EIPO uses a task-only policy ($\pi_\text{ref}=\text{J-only}$). However, relying solely on policies trained with task rewards as references may not suffice for complex robotic tasks, as they often perform much worse than those trained with heuristic rewards. We compared the performance of HEPO with different reference policies in Figure~\ref{fig:abl_ref_policy}. The result shows that setting $\pi_\text{ref}=\text{J-only}$ (EIPO) improves the performance over the task-only policy \textit{J-only} while notably degrading performance, sometimes even worse than \textit{{H-only}}, suggesting it's insufficient for surpassing the heuristic policy.

\paragraph{Strategy of collecting trajectories:}
We use both the enhanced policy $\pi$ and the heuristic policy $\pi_H$ simultaneously to sample half of the environment's trajectories (referred to as \textit{Joint}). Conversely, EIPO switches between $\pi$ and $\pi_H$ using a specified mechanism, where only one selected policy samples trajectories for updating both $\pi$ and $\pi_H$ within the same episode (referred to as \textit{Alternating}). This study compares the performance of these two trajectory rollout methods. We modify HEPO to gather trajectories using the \textit{Alternating} strategy and present the results in Figure~\ref{fig:abl_traj}. The findings indicate that \textit{Alternating} results in a performance drop during mid-training and fails to match the performance of \textit{HEPO(Joint)}. We hypothesize that this occurs because the batch of trajectories collected solely by one policy deviates significantly from those that another policy can generate (i.e., high off-policy error), leading to less effective PPO policy updates. In contrast, \textit{Joint} samples trajectories using both policies, preventing the collected trajectories from deviating too much from each other.

\begin{figure}[t!]
    \vspace{-10ex}
    \centering
    \begin{subfigure}[b]{0.9\textwidth}
        \centering
        \includegraphics[width=\textwidth]{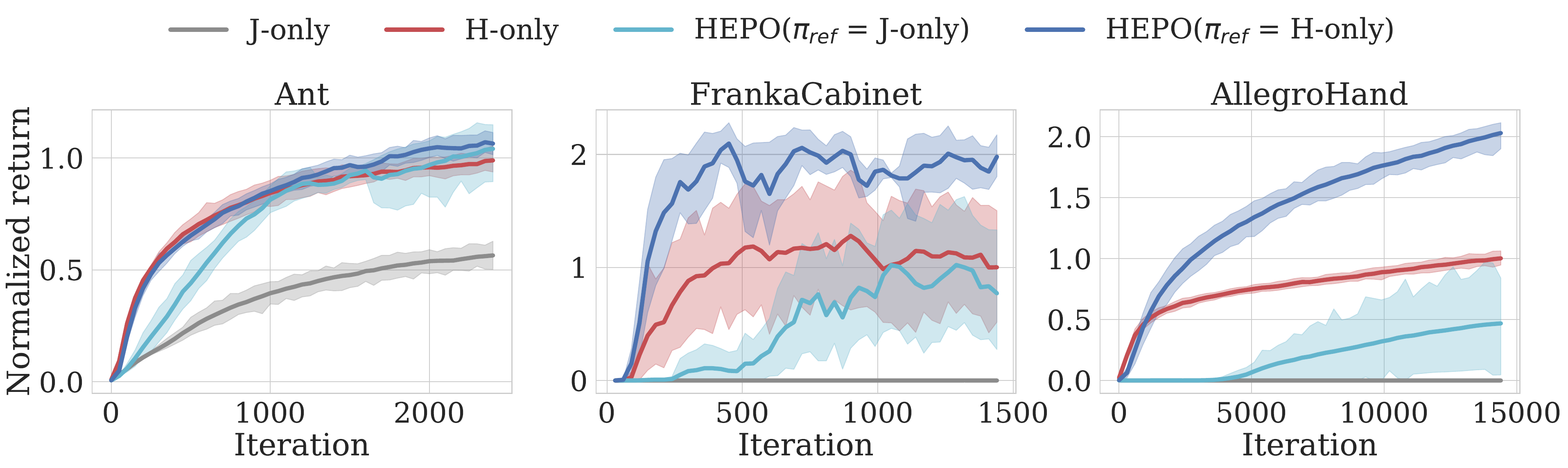}
        \caption{Comparison of reference policy $\pi_{\text{ref}}$ choice in HEPO's constraint $J(\pi) \geq J(\pi_{\text{ref}})$}
        \label{fig:abl_ref_policy}
    \end{subfigure} 
    \hfill
    \begin{subfigure}[b]{0.9\textwidth}
        \centering
        \includegraphics[width=\textwidth]{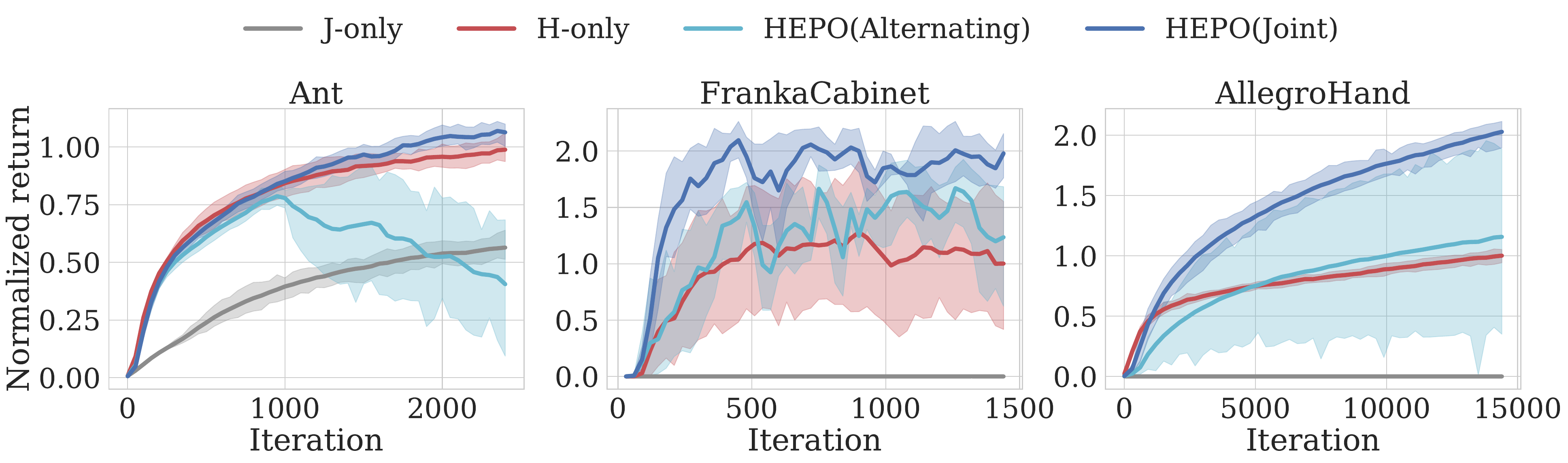}
        \caption{Comparison of trajectory collecting strategies 
        }
        \label{fig:abl_traj}
    \end{subfigure}
    \label{fig:abl_combined}
    \caption{\textbf{(a)}  We show that using the policies trained with heuristic rewards (J-only) is better than using the policies trained with task rewards (J-only) when training HEPO. \textbf{(b)} \textit{HEPO(Joint)} that collects trajectories using both policies leads to better performance than \textit{HEPO(Alternating)} that alternates between two policies to collect trajectories. See Section~\ref{subsec:exp:ablation} for details}
\end{figure}

\section{Related Works}
\label{sec:related}
\textbf{Reward shaping:} Reward shaping has been a significant area, including potential-based reward shaping (PBRS) \citep{ng1999policy, Devlin2012DynamicPR}, bilevel optimization approaches \citep{hu2020learning, gupta2024behavior, zheng2018learning} on reward model learning, and heuristic-guided methods (HuRL) \citep{cheng2021heuristic} that schedule heuristic rewards. Our method differs as it is a policy optimization method agnostic to the heuristic reward function and can be applied to those shaped or learned rewards.

\textbf{Constrained policy optimization:} Recent work like Extrinsic-Intrinsic Policy Optimization (EIPO) \citep{chen2022redeeming} proposes constrained optimization by tuning exploration bonuses to prevent exploiting them at the cost of task rewards. Extensions \citep{pmlr-v202-shenfeld23a} balance imitating a teacher model and reward maximization. Our work differs in balancing human-designed heuristic rewards and task rewards, improving upon policies trained with engineered heuristic rewards. We also propose implementation enhancements over EIPO \citep{chen2022redeeming} (Section~\ref{subsec:exp:ablation}).

\section{Discussion \& Limitation}
\label{sec:discussion}
\textbf{HEPO for RL practitioners:} In this paper, we showed that HEPO is robust to the possibly ill-designed heuristic reward function in Section~\ref{subsec:exp:application} and also exhibit high-probability improvement over PPO when training with heavily engineered heuristic rewards in robotic tasks in Section~\ref{subsec:exp:bench}. Moving forward, when users need to integrate heuristic reward functions into RL algorithms, HEPO can potentially be a useful tool to reduce users' time on designing rewards since it can improve performance even with under-engineered heuristic rewards.

\textbf{Limitations:} While HEPO shows high-probability performance improvement over heuristic policies trained with well-designed heuristic reward, one limitation is that HEPO does not have a guarantee to converge to the optimal policy theoretically. One future work can be incorporating the insight in recent theoretical advances on reward engineering \citep{gupta2022unpacking} to make a convergence guarantee. In addition, the requirement to train two policies in HEPO may hinder its adoption for large language models (LLMs) due to high memory demands. A promising direction for future work is to simplify HEPO to enable training with a single policy.

\section*{Acknowledgements}
We thank members of the Improbable AI Lab for helpful discussions and feedback. We are grateful to MIT Supercloud and the Lincoln Laboratory Supercomputing Center for providing HPC resources. This research was supported in part by Hyundai Motor Company,  Quanta Computer Inc., an AWS MLRA research grant, ARO MURI under Grant Number W911NF-23-1-0277, DARPA Machine Common Sense Program, ARO MURI under Grant Number W911NF-21-1-0328, and ONR MURI under Grant Number N00014-22-1-2740. The views and conclusions contained in this document are those of the authors and should not be interpreted as representing the official policies, either expressed or implied, of the Army Research Office or the United States Air Force or the U.S. Government. The U.S. Government is authorized to reproduce and distribute reprints for Government purposes, notwithstanding any copyright notation herein.

\section*{Author Contributions}
\begin{itemize}[leftmargin=*]
    \item \textbf{Chi-Chang Lee:} Co-led the project, led the implementation of the proposed algorithms and baselines, and conducted the experiments.    
    \item \textbf{Zhang-Wei Hong:} Co-led the project, led the writing of the paper, scaled up the experiment infrastructure, and conducted the experiments.   
    \item \textbf{Pulkit Agrawal:} Played a key role in overseeing the project, editing the manuscript, and the presentation of the work.
\end{itemize}

\setcitestyle{numbers}
\bibliography{main.bib}

\appendix

\newpage
\section{Implementation Details}
\subsection{Full Derivation}
\label{app:derivation}
We will detailedly describe the update of the enhanced policy ($\pi$ in Equation~\ref{eq:impl_pi}) and the heuristic policy ($\pi_H$ in Equation~\ref{eq:impl_pi_h}) at each iteration.

\subsubsection{Notations}
\label{app:notation}
\begin{itemize}
    \item $V^{\pi}_{r}(s_{t}) \coloneqq \mathbb{E}_{(s_t, a_t) \sim \pi}{\Big[\sum^{\infty}_{t=0}\gamma^t r(s_t, a_t) | s_0 = s_t\Big]}$
    \item $V^{\pi}_{h}(s_{t}) \coloneqq \mathbb{E}_{(s_t, a_t) \sim \pi}{\Big[\sum^{\infty}_{t=0}\gamma^t h(s_t, a_t) | s_0 = s_t\Big]}$
    \item $A^{\pi}_r(s_t, a_t) \coloneqq r(s_t, a_t) + V^{\pi}_r(s_{t+1}) - V^{\pi}_r(s_t)$
    \item $A^{\pi}_h(s_t, a_t) \coloneqq h(s_t, a_t) + V^{\pi}_h(s_{t+1}) - V^{\pi}_h(s_t)$
    \item $B_\text{HEPO}$: the buffer to store samples collected by $\pi^i$
    \item $B_H$: the buffer to store samples collected $\pi^i_H$. 
\end{itemize}

\subsubsection{Enhanced Policy $\pi$ Update}
Given a $\alpha$ value, $\pi^{i+1}$ is derived using the arguments of the maxima in Equation~\ref{eq:unconstrained_obj}, which can be re-written as follows:
\begin{align}
\begin{split}
\label{eq:mixed_obj}
    \pi^{i+1} &= \argmax_{\pi} \Big\{J(\pi) + H(\pi) - \alpha \Big(J(\pi) - J(\pi_H^i)\Big)\Big\} \\
        &= \argmax_{\pi} \Big\{(1 + \alpha) J(\pi) + H(\pi)\Big\} \\
        &= \argmax_{\pi} \Big\{\Big((1 + \alpha) J(\pi) + H(\pi)\Big) - \frac{1}{2}\Big((1 + \alpha) J(\pi^i) + H(\pi^i)\Big) \\
        &\text{ }\text{ }\text{ }\text{ }\text{ }\text{ }\text{ }\text{ }\text{ }\text{ }\text{ }\text{ }\text{ }\text{ }\text{ }\text{ }\text{ }\text{ }\text{ }\text{ }- \frac{1}{2}\Big((1 + \alpha)J(\pi_H^i) + H(\pi_H^i)\Big)\Big\}\\
        &= \argmax_{\pi} \Big\{\frac{1}{2}\mathbb{E}_{\pi} \Big[(1+\alpha) A^{\pi^i}_r(s_t, a_t) + A^{\pi^i}_h(s_t, a_t) \Big] \\
        &\text{ }\text{ }\text{ }\text{ }\text{ }\text{ }\text{ }\text{ }\text{ }\text{ }\text{ }\text{ }\text{ }\text{ }\text{ }\text{ }\text{ }\text{ }\text{ }\text{ } + \frac{1}{2}\mathbb{E}_{\pi} \Big[(1+\alpha) A^{\pi^i_H}_r(s_t, a_t)+A^{\pi^i_H}_h(s_t, a_t) \Big]\\
        &= \argmax_{\pi} \Big\{\frac{1}{2}\mathbb{E}_{\pi} \Big[U^{\pi^i}_{\alpha}(s_t, a_t)\Big] + \frac{1}{2}\mathbb{E}_{\pi} \Big[U^{\pi^i_H}_{\alpha}(s_t, a_t)\Big]\Big\} \\
        &= \argmax_{\pi} \Big\{\mathbb{E}_{\pi} \Big[U^{\pi^i}_{\alpha}(s_t, a_t)\Big] + \mathbb{E}_{\pi} \Big[U^{\pi^i_H}_{\alpha}(s_t, a_t)\Big]\Big\}
\end{split}
\end{align}
where $U^{\pi^i}_{\alpha}$ and $U^{\pi^i_H}_{\alpha}$ are defined as follows:
\begin{align}
    U^{\pi^i}_{\alpha}(s_t, a_t) &\coloneqq (1+\alpha) A^{\pi^i}_r(s_t, a_t) + A^{\pi^i}_h(s_t, a_t)  \nonumber\\
     U^{\pi^i_H}_{\alpha}(s_t, a_t)  &\coloneqq (1 + \alpha) A^{\pi^i_H}_r(s_t, a_t) + A^{\pi^i_H}_h(s_t, a_t)
\end{align}
To efficiently achieve the update process in Equation~\ref{eq:mixed_obj}, we aim to utilize previously collected trajectories for optimization, outlined in Equation~\ref{eq:impl_pi}.
Here, we refer to \cite{schulman2017proximal}, using those previously collected trajectories to form a lower bound surrogate objectives, $\hat{J}^{\pi^i}_{\alpha}(\pi)$ and $\hat{J}^{\pi_H^i}_{\alpha}(\pi)$, as alternatives of $\mathbb{E}_{\pi} [U^{\pi^i}_{\alpha}(s_t, a_t)]$ and $\mathbb{E}_{\pi} [U^{\pi^i_H}_{\alpha}(s_t, a_t)]$ to derive $\pi^{i+1}$:
\begin{align}
    \label{eq:HEPO_clipped}
    \begin{split}
    \hat{J}^{\pi^i}_\text{HEPO}(\pi) &\coloneqq \frac{1}{|B_\text{HEPO}|}\sum_{(s_t, a_t) \in B_\text{HEPO}} \Big[ \sum^\infty_{t=0} \gamma^{t} \min \Big\{  \dfrac{\pi(a_t|s_t)}{\pi^i(a_t|s_t)} U^{\pi^i}_{\alpha}(s_t,a_t), \\ 
    &\text{ }\text{ }\text{ }\text{ }\text{ }\text{ }\text{ }\text{ }\text{ }\text{ }\text{ }\text{ }\text{ }\text{ }\text{ }\text{ }\text{ }\text{ }\text{ }\text{ }\textrm{clip}\left(\dfrac{\pi(a_t|s_t)}{\pi^i(a_t|s_t)}, 1- \epsilon, 1+\epsilon  \right) U^{\pi^i}_{\alpha}(s_t,a_t) \Big\} \Big]\\
    \hat{J}^{\pi_H^i}_\text{HEPO}(\pi) &\coloneqq \frac{1}{|B_H|}\sum_{(s_t, a_t) \in B_H} \Big[ \sum^\infty_{t=0} \gamma^{t} \min \Big\{  \dfrac{\pi(a_t|s_t)}{\pi^i_H(a_t|s_t)} U^{\pi^i_H}_{\alpha}(s_t,a_t), \\ 
    &\text{ }\text{ }\text{ }\text{ }\text{ }\text{ }\text{ }\text{ }\text{ }\text{ }\text{ }\text{ }\text{ }\text{ }\text{ }\text{ }\text{ }\text{ }\text{ }\text{ }\textrm{clip}\left(\dfrac{\pi(a_t|s_t)}{\pi^i_H(a_t|s_t)}, 1- \epsilon, 1+\epsilon  \right) U^{\pi^i_H}_{\alpha}(s_t,a_t) \Big\} \Big],
    \end{split}
\end{align}
where $\mathbb{E}_{\pi} [U^{\pi^i}_{\alpha}(s_t, a_t)] \geq \hat{J}^{\pi^i}_\text{HEPO}(\pi)$ 
and $\mathbb{E}_{\pi} [U^{\pi^i_H}_{\alpha}(s_t, a_t)] \geq \hat{J}^{\pi_H^i}_\text{HEPO}(\pi)$ always hold; 
$\epsilon \in [0, 1]$ denotes a threshold.
Intuitively, this clipped objective (Eq.~\ref{eq:HEPO_clipped}) penalizes the policy $\pi$ that behaves differently from $\pi^i$ or $\pi^i_H$ because overly large or small the action probability ratios between two policies are clipped.

\subsubsection{Heuristic Policy $\pi_H$ Update}
$\pi^{i+1}_H$ is derived using the arguments of the maxima of $H(\pi)$, which can be re-written as follows:
\begin{align}
\begin{split}
\label{eq:ref_obj}
    \pi_H^{i+1} &= \argmax_{\pi} \Big\{H(\pi) \Big\} \\
        &= \argmax_{\pi} \Big\{H(\pi) - \frac{1}{2}H(\pi^i) - \frac{1}{2}H(\pi_H^i)\Big\} \\
        &= \argmax_{\pi} \Big\{ \frac{1}{2}\mathbb{E}_{\pi} \Big[A^{\pi^i}_h(s_t, a_t)\Big] + \frac{1}{2}\mathbb{E}_{\pi} \Big[A^{\pi^i_H}_h(s_t, a_t)\Big]\Big\}\\
        &= \argmax_{\pi} \Big\{ \mathbb{E}_{\pi} \Big[A^{\pi^i}_h(s_t, a_t)\Big] + \mathbb{E}_{\pi} \Big[A^{\pi^i_H}_h(s_t, a_t)\Big]\Big\}
\end{split}
\end{align}
Similarly, we again rely on the approximation from~\cite{schulman2017proximal} to derive a lower bound surrogate objective for both $\mathbb{E}_{\pi}[A^{\pi^i}_h(s_t, a_t)]$ and $\mathbb{E}_{\pi} [A^{\pi^i_H}_h(s_t, a_t)]$ as follows:
\begin{align}
    \label{eq:H_clipped}
    \begin{split}
    \hat{H}^{\pi^i}(\pi) \coloneqq \frac{1}{|B_\text{HEPO}|}\sum_{(s_t, a_t) \in B_\text{HEPO}} \Big[ \sum^\infty_{t=0} \gamma^{t} \min \Big\{  \dfrac{\pi(a_t|s_t)}{\pi^i(a_t|s_t)} A^{\pi^i}_h(s_t,a_t), \\ \textrm{clip}\left(\dfrac{\pi(a_t|s_t)}{\pi^i(a_t|s_t)}, 1- \epsilon, 1+\epsilon  \right) A^{\pi^i}_h(s_t,a_t) \Big\} \Big],
    \end{split}\\
    \begin{split}
    \hat{H}^{\pi_H^i}(\pi) \coloneqq \frac{1}{|B_H|}\sum_{(s_t, a_t) \in B_H} \Big[ \sum^\infty_{t=0} \gamma^{t} \min \Big\{  \dfrac{\pi(a_t|s_t)}{\pi^i_H(a_t|s_t)} A^{\pi^i_H}_h(s_t,a_t), \\ \textrm{clip}\left(\dfrac{\pi(a_t|s_t)}{\pi^i_H(a_t|s_t)}, 1- \epsilon, 1+\epsilon  \right) A^{\pi^i_H}_h(s_t,a_t) \Big\} \Big]
    \end{split}
\end{align}
where $\mathbb{E}_{\pi} [A^{\pi^i}_h(s_t, a_t)] \geq \hat{H}^{\pi^i}(\pi)$ and $\mathbb{E}_{\pi} [A^{\pi^i_H}_h(s_t, a_t)] \geq \hat{H}^{\pi_H^i}(\pi)$ always hold.
Different from vanilla heuristic training, instead of solely collecting trajectories from $\pi^i_H$, we collect trajectories from both $\pi$ and $\pi_H$ to enrich sample efficiency.

\subsection{Implementation Tricks}
\label{app:trick}
\subsubsection{Sample Sharing for Value Function Update}
\label{app:value_update}
In practice, obtaining real value functions for training is not feasible. We estimate the value function using collected trajectories, but this approach tends to fail because the value function becomes biased toward the policy responsible for trajectory collection.

To prevent error information from the estimated value function interfering with the training procedure, we share the trajectory samples within the $B_\text{HEPO}$ and $B_H$ buffers to update our value functions:
\begin{align}
\label{app:eq:vf}
V^{\pi^{i+1}}_r \leftarrow \argmin_{V}\Big\{&\sum_{\big(s_t, r(s_t, a_t), s_{t+1}\big)\in B_\text{HEPO} \cup B_H}\frac{|r(s_t, a_t) + \gamma V^{\pi^i}_r(s_{t+1}) - V(s_t)|^2}{|B_\text{HEPO}| + |B_H|} \Big\}\\
V^{\pi^{i+1}}_h \leftarrow \argmin_{V}\Big\{&\sum_{\big(s_t, h(s_t, a_t), s_{t+1}\big) \in B_\text{HEPO} \cup B_H}\frac{|h(s_t, a_t) + \gamma V^{\pi^i}_h(s_{t+1}) - V(s_t)|^2}{|B_\text{HEPO}| + |B_H|} \Big\}\\
V^{\pi^{i+1}_H}_r \leftarrow \argmin_{V}\Big\{&\sum_{\big(s_t, r(s_t, a_t), s_{t+1}\big)\in B_\text{HEPO} \cup B_H}\frac{|r(s_t, a_t) + \gamma V^{\pi^i_H}_r(s_{t+1}) - V(s_t)|^2}{|B_\text{HEPO}| + |B_H|} \Big\}\\
V^{\pi^{i+1}_H}_h \leftarrow \argmin_{V}\Big\{&\sum_{\big(s_t, h(s_t, a_t), s_{t+1}\big) \in B_\text{HEPO} \cup B_H}\frac{|h(s_t, a_t) + \gamma V^{\pi^i_H}_h(s_{t+1}) - V(s_t)|^2}{|B_\text{HEPO}| + |B_H|} \Big\}
\end{align}

\subsubsection{Smoothing Lagrangian Multiplier $\alpha$ Update}
The Lagrangian multiplier $\alpha$ determines the desired constraint information during training. However, in practice the gradient $\alpha$ tends to become explosive. 
To stabilize the $\alpha$ update procedure, we accumulate previous gradients and adopt the Adam optimizer~\citep{adam} as follows:
\begin{align}
\label{eq:alpha_update}
\begin{split}
    g(\alpha) &\leftarrow \text{med}\Big\{\frac{1}{|B_\text{HEPO}|}\sum_{(s_t, a_t) \in B_\text{HEPO}} \left[ A^{\pi^{i}_{H}}_r(s_t, a_t) \right] - \frac{1}{|B_H|}\sum_{(s_t, a_t) \in B_H} \left[ A^{\pi^{i}}_r(s_t, a_t) \right]\Big\}_{i-K}^i\\
\alpha &\leftarrow  \text{AdamOpt}\Big[g(\alpha) \Big]
\end{split}
\end{align}
where $K$ is the number of previous $K$ advantage expectation records that we take into account. 
To smooth the current $\alpha$ gradient for each update, we calculate the median of the previous $K$ records. 
In our experiments, we assigned $K$ a value of $8$.

\subsection{Overall Workflow}
\label{alg:minimmax}
\begin{algorithm}[H]
\caption{Detailed Heuristic-Enhanced Policy Optimization (HEPO)}
\begin{algorithmic}[1]

\State Initialize policies ($\pi^1$, $\pi^1_H$) and values ($V^{\pi^1}_r$, $V^{\pi^1}_h$, $V^{\pi^1_H}_r$, $V^{\pi^1_H}_h$)
\For{$i = 1 \cdots $} \Comment{$i$ denotes iteration index}
        \State \# ROLLOUT STAGE
        \State Collect trajectory buffers $(B_\text{HEPO}, B_H)$ using $(\pi^i, \pi^i_H)$
        \State Compute $\big(A^{\pi^i}_r(s_t, a_t), A^{\pi^i}_h(s_t, a_t)\big)$ via GAE with $\big(V^{\pi^i}_r, V^{\pi^i}_h\big)$ $\forall (s_t, a_t) \in B_\text{HEPO}$
        \State Compute $\big(A^{\pi^i_H}_r(s_t, a_t), A^{\pi^i_H}_h(s_t, a_t)\big)$ via GAE with $\big(V^{\pi^i_H}_r, V^{\pi^i_H}_h\big)$ $\forall (s_t, a_t) \in B_H$
        \State Compute $\big(\hat{J}^{\pi^i}_\text{HEPO}, \hat{J}^{\pi^i_H}_\text{HEPO}\big)$ based on Equation~\ref{eq:HEPO_clipped}
        \State Compute $\big(\hat{H}^{\pi^i}, \hat{H}^{\pi^i_H}\big)$ based on Equation~\ref{eq:H_clipped}\\
        \State \# UPDATE STAGE
        \State $\pi^{i+1} \leftarrow \argmax_\pi\Big\{\hat{J}^{\pi^i}_\text{HEPO}(\pi) + \hat{J}^{\pi^i_H}_\text{HEPO}(\pi)\Big\}$
        \State $\pi^{i+1}_H \leftarrow \argmax_\pi\Big\{\hat{H}^{\pi^i}(\pi) + \hat{H}^{\pi^i_H}(\pi)\Big\}$
        \State Update $(V^{\pi^i}_r, V^{\pi^i}_h, V^{\pi^i_H}_r, V^{\pi^i_H}_h)$ based on Equation~\ref{app:eq:vf}
        \State Update $\alpha$ based on Equation~\ref{eq:alpha_update}
\EndFor

\end{algorithmic}
\end{algorithm}
\subsection{Training details}
\label{app:sec:training}
Following the PPO framework \citep{schulman2017proximal}, our experiments are based on a continuous action actor-critic algorithm implemented in \texttt{rl\_games}~\cite{rl-games2021}, using Generalized Advantage Estimation (GAE) \citep{schulman2015high} to compute advantages for policy optimization. For PPO, we employed the same policy network and value network architecture, and the same hyperparameters used in \texttt{IsaacGymEnvs} \citep{makoviychuk2021isaac}. We also include our source code in the supplementary material. In HEPO, we use two policies for optimization, with each policy maintaining the same model configurations as those used in PPO. 
Below we introduce HEPO-specific hyperparameters used in our experiments in Section~\ref{subsec:exp:bench}.
The hyperparameters for updating the Lagrangian multiplier $\alpha$ in HEPO are listed as follows:
\begin{table}[H]
\centering
\caption{HEPO Hyperparameters}
\label{app:tab:eipo_hype}
\begin{tabular}{|c|c|}
\hline
\textbf{Name  }                                 & \textbf{Value}  \\ \hline
Initial $\alpha$                 & 0.0   \\ \hline
Step size $\eta$ of $\alpha$ (learning rate)                 & 0.01   \\ \hline
Clipping range of $\delta \alpha$ $(-\epsilon_{\alpha}, \epsilon_{\alpha})$                 & 1.0  \\ \hline
Range of the $\alpha$ value                 & $[0, \infty)$  \\ \hline
\end{tabular}
\end{table}

For baselines, we search for hyperparamters $\lambda$ for $J+H$ in \textit{Ant}, \textit{FrankaCabinet}, and \textit{AllegroHand}, as shown in Section~\ref{subsec:exp:sensitivity}. We set $\lambda = 1$ for all the experiments because it shows better performance on the three chosen environments. For HuRL \citep{cheng2021heuristic}, we follow the scheduling setting provided in their paper.

\newpage
\section{Supplementary Experimental Results}
\subsection{All Learning Curves on the task objective $J$}
\label{app:all_curve}
We present all the learning curves in Figure~\ref{app:fig:all_curves}.
\begin{figure}[htb!]
    \centering
    \includegraphics[width=\textwidth]{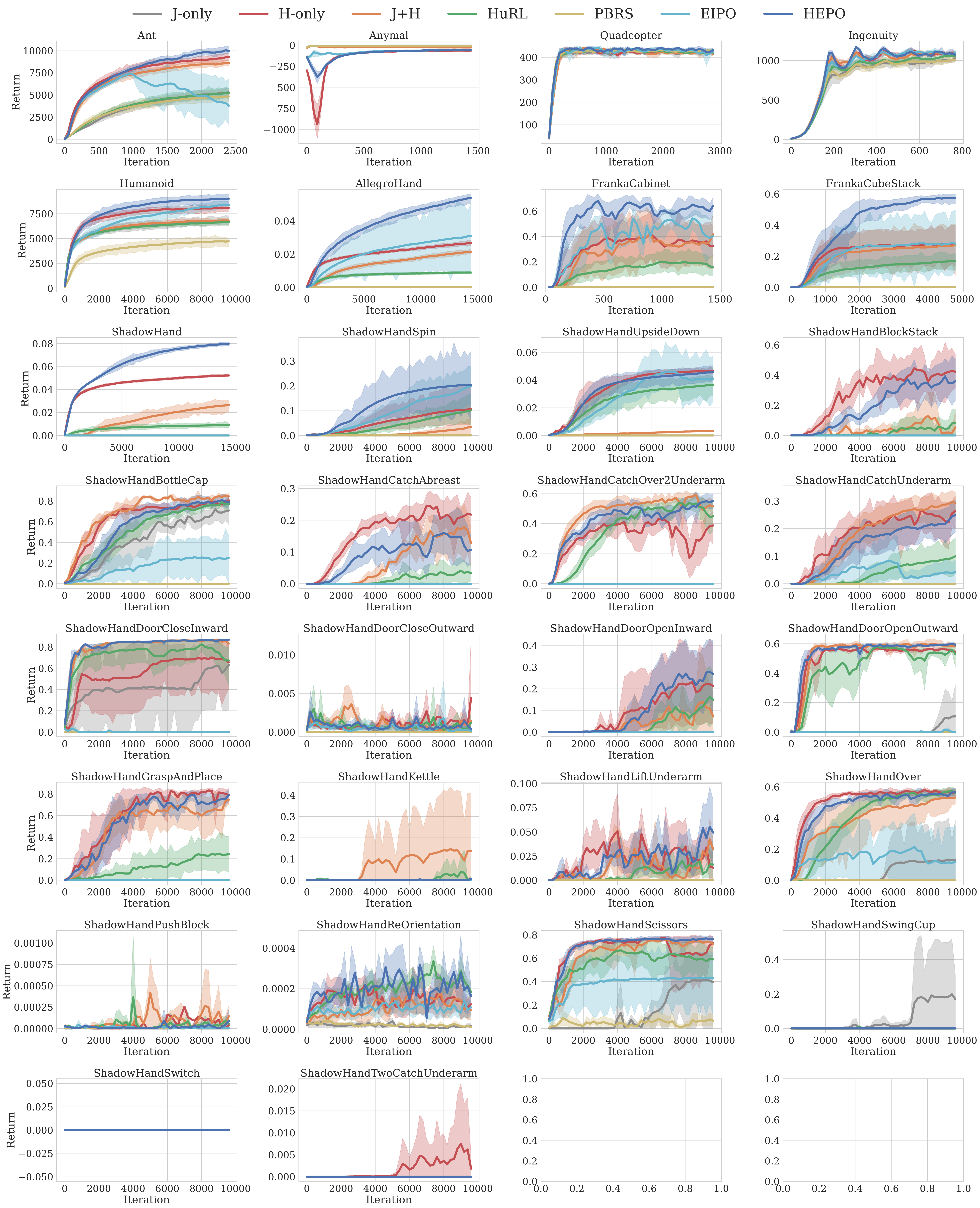}
    \caption{All learning curves in Section~\ref{subsec:exp:bench}}
    \label{app:fig:all_curves}
\end{figure}

\subsection{Sensitivity to hyperparameters}
\label{subsec:exp:sensitivity}
In this section, we aim to verify HEPO's sensitivity to two main types of hyperparameters: (1) the weight of the heuristic reward in optimization (denoted as $\lambda$) and (2) the learning rate for updating $\alpha$. 
Similar to Section~\ref{subsec:exp:ablation}, we conducted our experiments on the {\textit{Ant}}, {\textit{FrankaCabinet}}, and {\textit{AllegroHand}} tasks.

\subsubsection{Sensitivity to the $\lambda$ Value}
\label{app:lmbd}
Both \textbf{HEPO} and \textbf{J+H} can set a scaling coefficient to weight the heuristic reward in optimization, such that the objective becomes $J(\pi) + \lambda H(\pi)$. This scaling coefficient can be used to balance both objectives. 
In this study, we compare \textbf{HEPO} and \textbf{J+H} on their performance sensitivity to the choice of $\lambda$, exhaustively training both \textbf{HEPO} and \textbf{J+H} with varying $\lambda$ values. Note that though the formulation of HEPO does not depend on $\lambda$, one can still set a $\lambda$ coefficient to scale the heuristic reward in HEPO. In our experiments, we did not optimize $\lambda$ for HEPO but for the baselines trained with both rewards (J+H). In Figure~\ref{fig:sens_lmbd}, we found that \textbf{J+H} is sensitive to $\lambda$ in all selected tasks, while \textbf{HEPO} performs well across a wide range of $\lambda$ values. This indicates that \textbf{HEPO} is robust to the choice of $\lambda$.

\begin{figure}
    \centering
    \includegraphics[width=\textwidth=0.5]{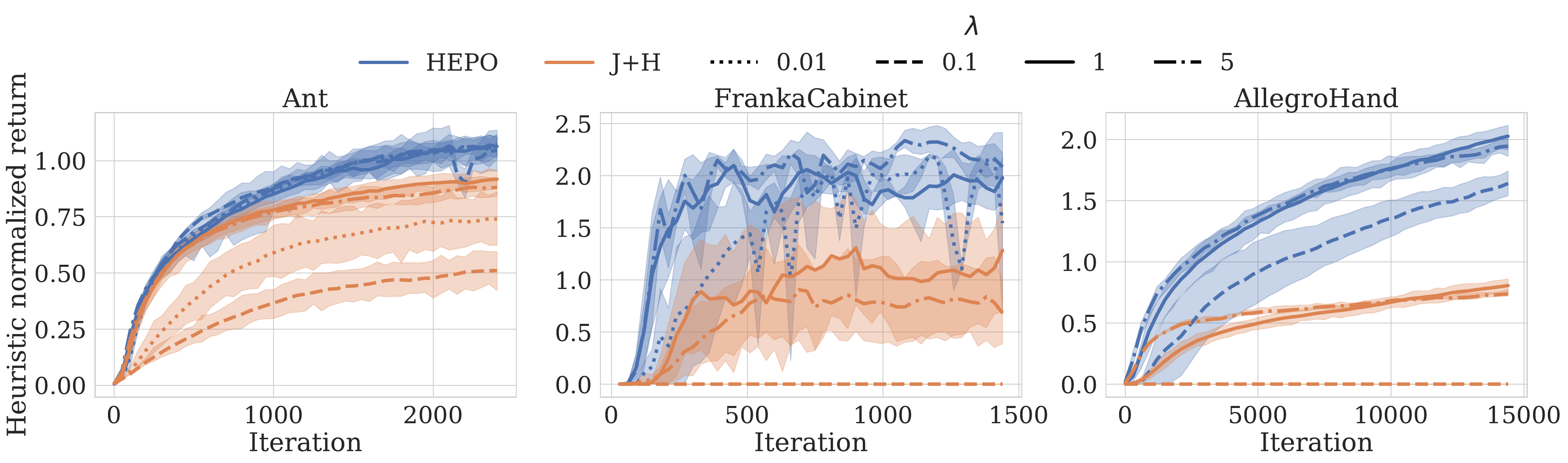}
    \caption{Sensitivity to $\lambda$}
    \label{fig:sens_lmbd}
\end{figure}

\subsubsection{Sensitivity to the Learning Rate for Updating $\alpha$} 
HEPO's robustness relies on the $\alpha$ update, as it reflects the necessary constraint information at each iteration. Similar to Appendix~\ref{app:lmbd}, setting different initial values of $\alpha$ is equivalent to using different $\lambda$ values for our estimation, since both can be rewritten as the ratio between $H(\pi)$ and $J(\pi)$. Both of these parameters indicate the necessary constraint information for conducting multi-objective optimization.
In this study, we aim to verify whether \textbf{HEPO} can yield comparable improvement gaps under different initial values of $\alpha$, thus providing a more robust optimization procedure.

As shown in Figure~\ref{fig:sens_alphalr}, we observe that \textbf{HEPO} is also robust to the choice of $\alpha$'s initial values, similar to the results in Figure~\ref{fig:sens_lmbd}.
\begin{figure}
    \centering
    \includegraphics[width=\textwidth=0.5]{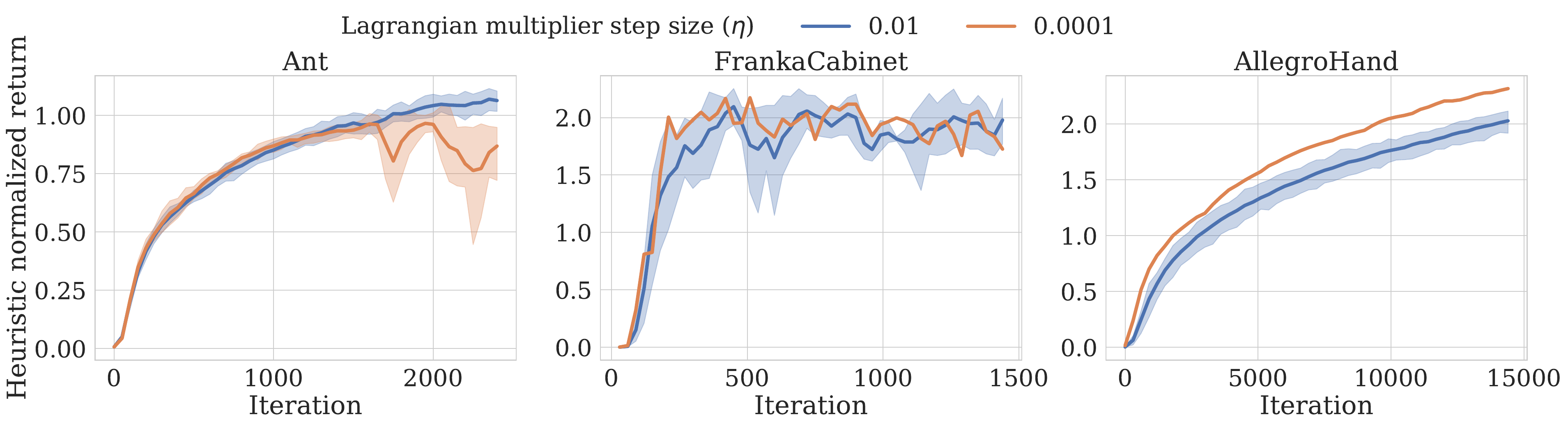}
    \caption{Sensitivity to alpha learning rate}
    \label{fig:sens_alphalr}
\end{figure}

\subsection{Additional results}
\label{app:additional_results}

\textbf{Generality of HEPO:} We also demonstrated HEPO can be implemented over the other RL algorithms in addition to PPO. We integrated HEPO into HuRL's SAC codebase. Despite HuRL using tuned hyperparameters as reported in its paper \citep{cheng2021heuristic}, HEPO outperformed SAC and matched HuRL on the most challenging task in Figure~\ref{fig:rebuttal:sparse_reacher} using the same hyperparameters from Section~\ref{sec:exp} of our manuscript, showing the generality of HEPO on different RL algorithms. 

\textbf{Better than EIPO on the most challenging task:} Comparing HEPO and EIPO on the most challenging task, Montezuma's Revenge, reported in EIPO's paper \citep{chen2022redeeming} using RND exploration bonuses \citep{burda2018exploration} (as used in the EIPO paper), Figure \ref{fig:rebuttal:mrevenge} shows that HEPO performs better than EIPO. Also, HEPO matched PPO trained with RND bonuses at convergence (2 billion frames) reported in \citep{burda2018exploration} using only 20\% of the training data, demonstrating drastically improved sample efficiency.

\textbf{How quality of heuristic reward functions impact HEPO?}  We believe that Figure \ref{fig:human_study_bar} reveals the relationship between HEPO's performance and the quality of the heuristic reward. The policy trained with only heuristic rewards (H-only) represents both the asymptotic performance of in HEPO and the quality of the heuristic itself. We found a positive correlation (Pearson coefficient of 0.9) between the average performances of H-only and HEPO in Figure \ref{fig:human_study_bar} results, suggesting that better heuristics lead to improved HEPO performance. Figure \ref{fig:rebuttal:corr} provides more details.

\begin{figure}[h!]
    \centering
    \begin{subfigure}[b]{0.3\textwidth}
        \centering
        \includegraphics[width=\textwidth]{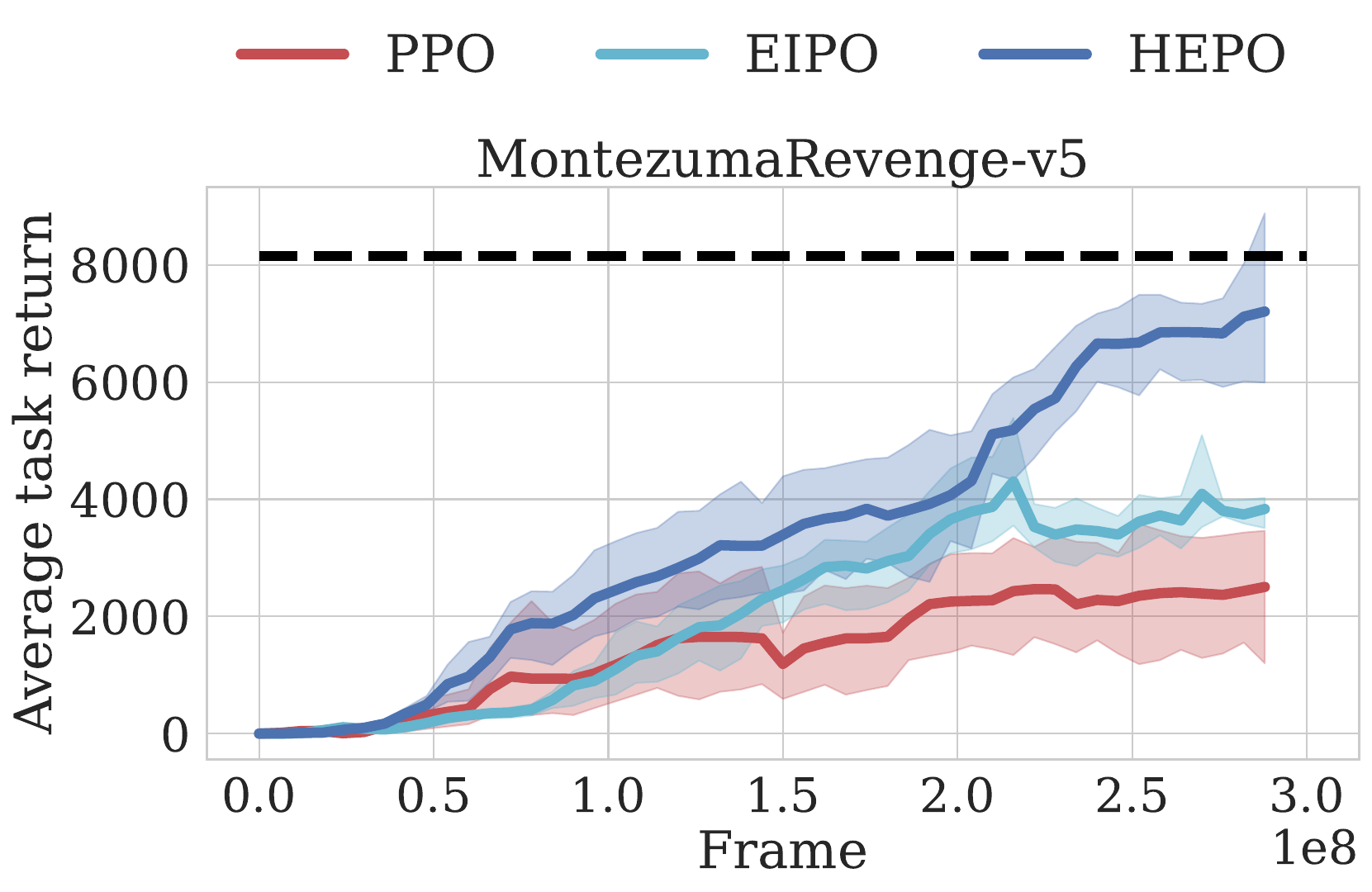}
        \caption{EIPO v.s. HEPO on PPO}
        \label{fig:rebuttal:mrevenge}
    \end{subfigure}%
    \hfill
    \begin{subfigure}[b]{0.33\textwidth}
        \centering
        \includegraphics[width=\textwidth]{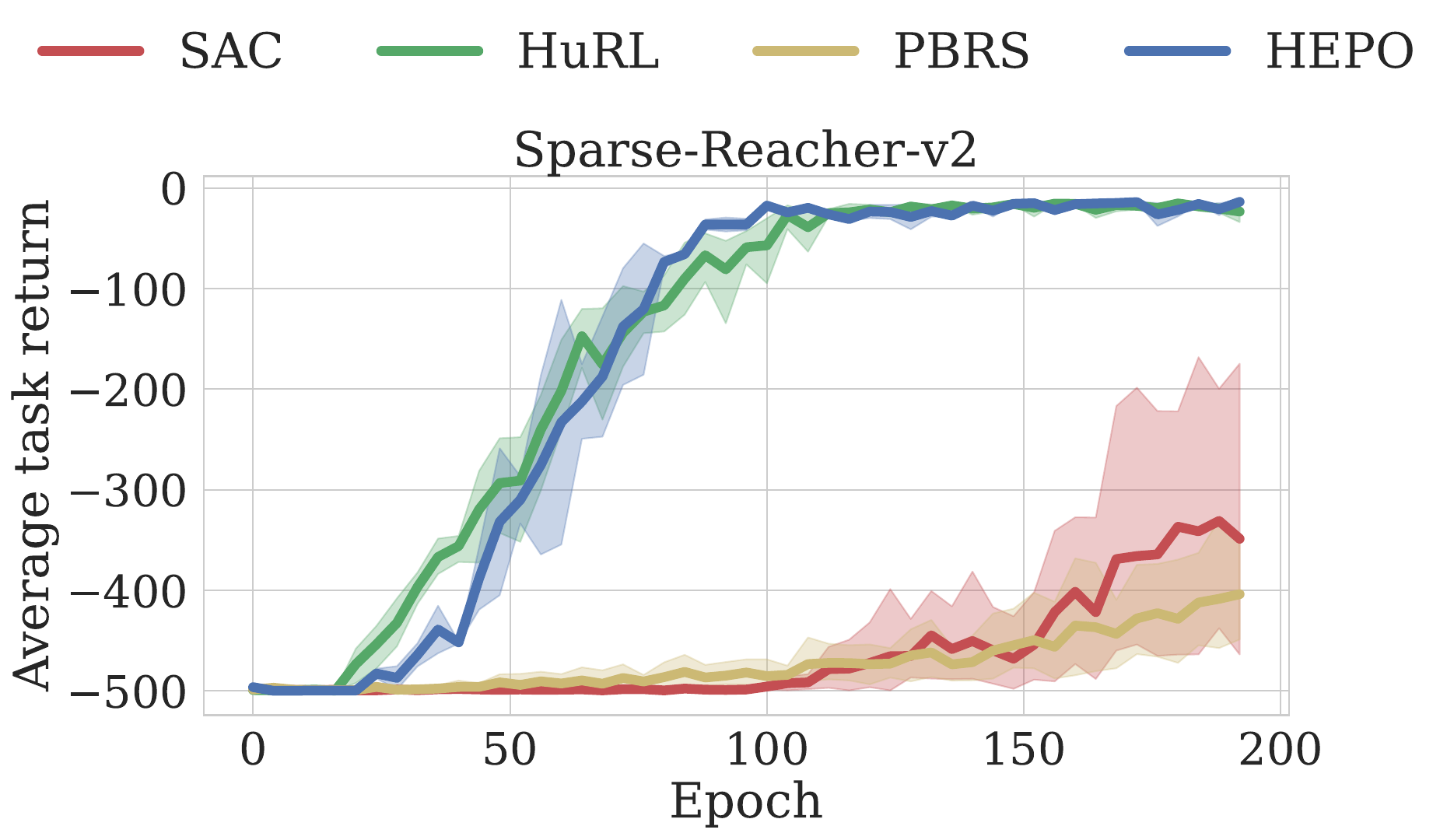}
        \caption{HuRL v.s. HEPO on SAC}
        \label{fig:rebuttal:sparse_reacher}
    \end{subfigure}%
    \hfill
    \begin{subfigure}[b]{0.27\textwidth}
        \centering
        \includegraphics[width=\textwidth]{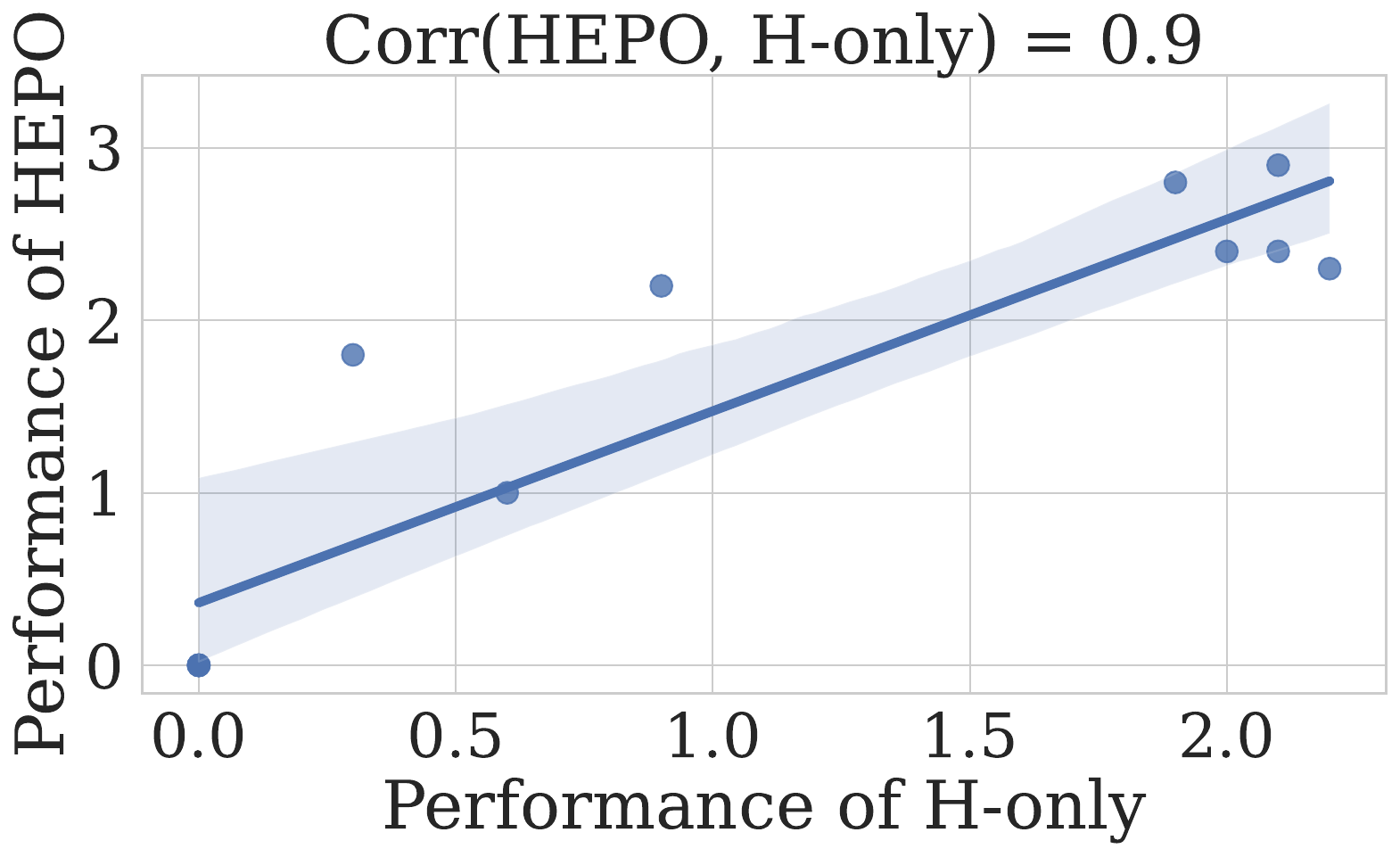}
        \caption{Correlation between the performance of HEPO and H-only.}
        \label{fig:rebuttal:corr}
    \end{subfigure}%
    \caption{\textbf{(a)} Comparison of HEPO and EIPO \citep{chen2022redeeming} on the most challenging Atari task, \texttt{Montezuma's Revenge}, shown in the EIPO paper \citep{chen2022redeeming}. Both are implemented on top of EIPO's PPO codebase using RND exploration bonuses \citep{burda2018exploration} as heuristic rewards $H$, as suggested in \cite{chen2022redeeming}. HEPO outperforms EIPO, achieving the performance (denoted as dashed line) similar to PPO trained with RND at convergence (2 billion frames) reported in \cite{burda2018exploration} in five times fewer frames. \textbf{(b)} HEPO matches HuRL's performance on the most challenging \texttt{Sparse-Reacher} task using HuRL's SAC codebase \citep{cheng2021heuristic}, despite HuRL being tuned for this task and HEPO using the same hyperparameters from our Section \ref{sec:exp}. This also highlights HEPO's generality in different RL algorithms. \textbf{(c)} HEPO's performance is positively correlated with that of the heuristic policy trained with heuristic rewards only (H-only), suggesting that HEPO's effectiveness will improve as the quality of heuristic rewards increases.
    }
    \label{fig:rebuttal:curve_corr}
\end{figure}

\section{Environment Details}
\label{appendix: task_desc}
As depicted in Section~\ref{sec:exp}, we conducted our experiments based on the Isaac Gym (\textbf{\textsc{Isaac}}) simulator~\citep{makoviychuk2021isaac} and the Bi-DexHands (\textbf{\textsc{Bi-Dex}}) benchmark~\citep{chen2022towards}.
The selected task classes in \textsc{Isaac} can be partitioned into 4 groups - Locomotion Tracking (\textbf{\textit{Anymal}}), Locomotion Progressing (\textbf{\textit{Ant}} and \textbf{\textit{Humanoid}}), Helicopter Progressing (\textbf{\textit{Ingenuity}} and \textbf{\textit{Quadcopter}}), and Manipulation Tasks (\textbf{\textit{FrankaCabinet}}, \textbf{\textit{FrankaCubeStack}}, \textbf{\textit{ShadowHand}}, and \textbf{\textit{AllegroHand}}).
In addition, \textsc{Bi-Dex} provides dual dexterous hand manipulation tasks through \textsc{Isaac}, reaching human-level sophistication of hand dexterity and bimanual coordination.
Their tasks include \textbf{\textit{ShadowHandOver}}, \textbf{\textit{ShadowHandCatchUnderarm}}, \textbf{\textit{ShadowHandCatchOver2Underarm}}, \textbf{\textit{ShadowHandCatchAbreast}}, \textbf{\textit{ShadowHandTwoCatchUnderarm}}, \textbf{\textit{ShadowHandLiftUnderarm}}, \textbf{\textit{ShadowHandDoorOpenInward}}, \textbf{\textit{ShadowHandDoorOpenOutward}}, \textbf{\textit{ShadowHandDoorCloseInward}}, \textbf{\textit{ShadowHandDoorCloseOutward}}, \textbf{\textit{ShadowHandSpin}}, \textbf{\textit{ShadowHandUpsideDown}}, \textbf{\textit{ShadowHandBlockStack}}, \textbf{\textit{ShadowHandBottleCap}}, \textbf{\textit{ShadowHandGraspAndPlace}}, \textbf{\textit{ShadowHandKettle}}, \textbf{\textit{ShadowHandPen}}, \textbf{\textit{ShadowHandPushBlock}}, \textbf{\textit{ShadowHandReOrientation}}, \textbf{\textit{ShadowHandScissors}}, \textbf{\textit{ShadowHandSwingCup}}.

For Locomotion Tracking, our emphasis lies in assessing the precision of velocities in linear and angular motions, ensuring that the robot responds closely to the assigned values. 
To this end, we define tracking errors as our task rewards.
For Locomotion Progressing and Helicopter Progressing, our emphasis lies in evaluating the progress made by the robots in reaching the assigned destination from a given start point. 
To this end, we define movement progress as our task rewards.
For Manipulation tasks and all tasks within the \textsc{Bi-Dex} benchmark, our emphasis lies in whether and how quickly the robotic hands can successfully complete the assigned missions, reaching the desired goal states. 
To this end, we define task rewards using a binary label, assigning a value of 1 to indicate successful attainment of the goal state and 0 otherwise.

The following are our reward function definitions, which include heuristic and task reward terms in Python style:

\begin{figure}[H]
\centering
\caption*{Isaac Gym - Locomotion Tracking Task: \textbf{\textit{Anymal}}}
\begin{minted}[mathescape, breaklines,frame=single, fontsize=\scriptsize]{python}
def compute_anymal_reward(
    root_states,
    commands,
    torques,
    contact_forces,
    knee_indices,
    episode_lengths,
    rew_scales,
    base_index,
    max_episode_length
):
    # (reward, reset, feet_in air, feet_air_time, episode sums)
    # type: (Tensor, Tensor, Tensor, Tensor, Tensor, Tensor, Dict[str, float], int, int) -> Tuple[Tensor, Tensor, Tensor]

    # prepare quantities (TODO: return from obs ?)
    base_quat = root_states[:, 3:7]
    base_lin_vel = quat_rotate_inverse(base_quat, root_states[:, 7:10])
    base_ang_vel = quat_rotate_inverse(base_quat, root_states[:, 10:13])

    # velocity tracking reward
    lin_vel_error = torch.sum(torch.square(commands[:, :2] - base_lin_vel[:, :2]), dim=1)
    ang_vel_error = torch.square(commands[:, 2] - base_ang_vel[:, 2])
    rew_lin_vel_xy = torch.exp(-lin_vel_error/0.25) * rew_scales["lin_vel_xy"]
    rew_ang_vel_z = torch.exp(-ang_vel_error/0.25) * rew_scales["ang_vel_z"]

    # torque penalty
    rew_torque = torch.sum(torch.square(torques), dim=1) * rew_scales["torque"]

    total_reward = rew_lin_vel_xy + rew_ang_vel_z + rew_torque
    total_reward = torch.clip(total_reward, 0., None)
    tracking_reward = -(lin_vel_error + ang_vel_error)

    # reset agents
    reset = torch.norm(contact_forces[:, base_index, :], dim=1) > 1.
    reset = reset | torch.any(torch.norm(contact_forces[:, knee_indices, :], dim=2) > 1., dim=1)
    time_out = episode_lengths >= max_episode_length - 1  # no terminal reward for time-outs
    reset = reset | time_out
    heuristic_reward, task_reward = total_reward.detach(), tracking_reward
    return heuristic_reward, task_reward, reset
\end{minted}
\end{figure}

\begin{figure}[H]
\centering
\caption*{Isaac Gym - Locomotion Progressing Task: \textbf{\textit{Ant}}}
\begin{minted}[mathescape, breaklines,frame=single, fontsize=\scriptsize]{python}
def compute_ant_reward(
    obs_buf,
    reset_buf,
    progress_buf,
    actions,
    up_weight,
    heading_weight,
    potentials,
    prev_potentials,
    actions_cost_scale,
    energy_cost_scale,
    joints_at_limit_cost_scale,
    termination_height,
    death_cost,
    max_episode_length
):
    # type: (Tensor, Tensor, Tensor, Tensor, float, float, Tensor, Tensor, float, float, float, float, float, float) -> Tuple[Tensor, Tensor, Tensor]

    # reward from direction headed
    heading_weight_tensor = torch.ones_like(obs_buf[:, 11]) * heading_weight
    heading_reward = torch.where(obs_buf[:, 11] > 0.8, heading_weight_tensor, heading_weight * obs_buf[:, 11] / 0.8)

    # aligning up axis of ant and environment
    up_reward = torch.zeros_like(heading_reward)
    up_reward = torch.where(obs_buf[:, 10] > 0.93, up_reward + up_weight, up_reward)

    # energy penalty for movement
    actions_cost = torch.sum(actions ** 2, dim=-1)
    electricity_cost = torch.sum(torch.abs(actions * obs_buf[:, 20:28]), dim=-1)
    dof_at_limit_cost = torch.sum(obs_buf[:, 12:20] > 0.99, dim=-1)

    # reward for duration of staying alive
    alive_reward = torch.ones_like(potentials) * 0.5
    progress_reward = potentials - prev_potentials

    total_reward = progress_reward + alive_reward + up_reward + heading_reward - \
        actions_cost_scale * actions_cost - energy_cost_scale * electricity_cost - dof_at_limit_cost * joints_at_limit_cost_scale

    # adjust reward for fallen agents
    total_reward = torch.where(obs_buf[:, 0] < termination_height, torch.ones_like(total_reward) * death_cost, total_reward)
    
    # reset agents
    reset = torch.where(obs_buf[:, 0] < termination_height, torch.ones_like(reset_buf), reset_buf)
    reset = torch.where(progress_buf >= max_episode_length - 1, torch.ones_like(reset_buf), reset)
    heuristic_reward, task_reward = total_reward, progress_reward
    return heuristic_reward, task_reward, reset
\end{minted}
\end{figure}

\begin{figure}[H]
\centering
\caption*{Isaac Gym - Locomotion Progressing Task: \textbf{\textit{Humanoid}}}
\begin{minted}[mathescape, breaklines,frame=single, fontsize=\scriptsize]{python}
def compute_humanoid_reward(
    obs_buf,
    reset_buf,
    progress_buf,
    actions,
    up_weight,
    heading_weight,
    potentials,
    prev_potentials,
    actions_cost_scale,
    energy_cost_scale,
    joints_at_limit_cost_scale,
    max_motor_effort,
    motor_efforts,
    termination_height,
    death_cost,
    max_episode_length
):
    # type: (Tensor, Tensor, Tensor, Tensor, float, float, Tensor, Tensor, float, float, float, float, Tensor, float, float, float) -> Tuple[Tensor, Tensor, Tensor]

    # reward from the direction headed
    heading_weight_tensor = torch.ones_like(obs_buf[:, 11]) * heading_weight
    heading_reward = torch.where(obs_buf[:, 11] > 0.8, heading_weight_tensor, heading_weight * obs_buf[:, 11] / 0.8)

    # reward for being upright
    up_reward = torch.zeros_like(heading_reward)
    up_reward = torch.where(obs_buf[:, 10] > 0.93, up_reward + up_weight, up_reward)

    actions_cost = torch.sum(actions ** 2, dim=-1)

    # energy cost reward
    motor_effort_ratio = motor_efforts / max_motor_effort
    scaled_cost = joints_at_limit_cost_scale * (torch.abs(obs_buf[:, 12:33]) - 0.98) / 0.02
    dof_at_limit_cost = torch.sum((torch.abs(obs_buf[:, 12:33]) > 0.98) * scaled_cost * motor_effort_ratio.unsqueeze(0), dim=-1)

    electricity_cost = torch.sum(torch.abs(actions * obs_buf[:, 33:54]) * motor_effort_ratio.unsqueeze(0), dim=-1)

    # reward for duration of being alive
    alive_reward = torch.ones_like(potentials) * 2.0
    progress_reward = potentials - prev_potentials

    total_reward = progress_reward + alive_reward + up_reward + heading_reward - \
        actions_cost_scale * actions_cost - energy_cost_scale * electricity_cost - dof_at_limit_cost

    # adjust reward for fallen agents
    total_reward = torch.where(obs_buf[:, 0] < termination_height, torch.ones_like(total_reward) * death_cost, total_reward)
    
    # reset agents
    reset = torch.where(obs_buf[:, 0] < termination_height, torch.ones_like(reset_buf), reset_buf)
    reset = torch.where(progress_buf >= max_episode_length - 1, torch.ones_like(reset_buf), reset)
    heuristic_reward, task_reward = total_reward, progress_reward
    return heuristic_reward, task_reward, reset
\end{minted}
\end{figure}

\begin{figure}[H]
\centering
\caption*{Isaac Gym - Helicopter Progressing Task: \textbf{\textit{Ingenuity}}}
\begin{minted}[mathescape, breaklines,frame=single, fontsize=\scriptsize]{python}
def compute_ingenuity_reward(root_positions, target_root_positions, root_quats, root_linvels, root_angvels, reset_buf, progress_buf, max_episode_length):
    # type: (Tensor, Tensor, Tensor, Tensor, Tensor, Tensor, Tensor, float) -> Tuple[Tensor, Tensor, Tensor]

    # distance to target
    target_dist = torch.sqrt(torch.square(target_root_positions - root_positions).sum(-1))
    pos_reward = 1.0 / (1.0 + target_dist * target_dist)

    # uprightness
    ups = quat_axis(root_quats, 2)
    tiltage = torch.abs(1 - ups[..., 2])
    up_reward = 5.0 / (1.0 + tiltage * tiltage)

    # spinning
    spinnage = torch.abs(root_angvels[..., 2])
    spinnage_reward = 1.0 / (1.0 + spinnage * spinnage)

    # combined reward
    # uprigness and spinning only matter when close to the target
    reward = pos_reward + pos_reward * (up_reward + spinnage_reward)

    # resets due to misbehavior
    ones = torch.ones_like(reset_buf)
    die = torch.zeros_like(reset_buf)
    die = torch.where(target_dist > 8.0, ones, die)
    die = torch.where(root_positions[..., 2] < 0.5, ones, die)

    # resets due to episode length
    reset = torch.where(progress_buf >= max_episode_length - 1, ones, die)
    heuristic_reward, task_reward = reward, pos_reward
    return heuristic_reward, task_reward, reset
\end{minted}
\end{figure}

\begin{figure}[H]
\centering
\caption*{Isaac Gym - Helicopter Progressing Task: \textbf{\textit{Quadcopter}}}
\begin{minted}[mathescape, breaklines,frame=single, fontsize=\scriptsize]{python}
def compute_quadcopter_reward(root_positions, root_quats, root_linvels, root_angvels, reset_buf, progress_buf, max_episode_length):
    # type: (Tensor, Tensor, Tensor, Tensor, Tensor, Tensor, float) -> Tuple[Tensor, Tensor, Tensor]

    # distance to target
    target_dist = torch.sqrt(root_positions[..., 0] * root_positions[..., 0] +
                             root_positions[..., 1] * root_positions[..., 1] +
                             (1 - root_positions[..., 2]) * (1 - root_positions[..., 2]))
    pos_reward = 1.0 / (1.0 + target_dist * target_dist)

    # uprightness
    ups = quat_axis(root_quats, 2)
    tiltage = torch.abs(1 - ups[..., 2])
    up_reward = 1.0 / (1.0 + tiltage * tiltage)

    # spinning
    spinnage = torch.abs(root_angvels[..., 2])
    spinnage_reward = 1.0 / (1.0 + spinnage * spinnage)

    # combined reward
    # uprigness and spinning only matter when close to the target
    reward = pos_reward + pos_reward * (up_reward + spinnage_reward)
    
    # resets due to misbehavior
    ones = torch.ones_like(reset_buf)
    die = torch.zeros_like(reset_buf)
    die = torch.where(target_dist > 3.0, ones, die)
    die = torch.where(root_positions[..., 2] < 0.3, ones, die)

    # resets due to episode length
    reset = torch.where(progress_buf >= max_episode_length - 1, ones, die)
    heuristic_reward, task_reward = reward, pos_reward
    return heuristic_reward, task_reward, reset
\end{minted}
\end{figure}

\begin{figure}[H]
\centering
\caption*{Isaac Gym - Manipulation Task: \textbf{\textit{FrankaCabinet}}}
\begin{minted}[mathescape, breaklines,frame=single, fontsize=\scriptsize]{python}
def compute_franka_reward(
    reset_buf, progress_buf, reset_goal_buf, successes, consecutive_successes, actions, cabinet_dof_pos,
    franka_grasp_pos, drawer_grasp_pos, franka_grasp_rot, drawer_grasp_rot,
    franka_lfinger_pos, franka_rfinger_pos,
    gripper_forward_axis, drawer_inward_axis, gripper_up_axis, drawer_up_axis,
    num_envs, dist_reward_scale, rot_reward_scale, around_handle_reward_scale, open_reward_scale,
    finger_dist_reward_scale, action_penalty_scale, distX_offset, max_episode_length
):
    # type: (Tensor, Tensor, Tensor, Tensor, Tensor, Tensor, Tensor, Tensor, Tensor, Tensor, Tensor, Tensor, Tensor, Tensor, Tensor, Tensor, Tensor, int, float, float, float, float, float, float, float, float) -> Tuple[Tensor, Tensor, Tensor, Tensor]

    # distance from hand to the drawer
    d = torch.norm(franka_grasp_pos - drawer_grasp_pos, p=2, dim=-1)
    dist_reward = 1.0 / (1.0 + d ** 2)
    dist_reward *= dist_reward
    dist_reward = torch.where(d <= 0.02, dist_reward * 2, dist_reward)

    axis1 = tf_vector(franka_grasp_rot, gripper_forward_axis)
    axis2 = tf_vector(drawer_grasp_rot, drawer_inward_axis)
    axis3 = tf_vector(franka_grasp_rot, gripper_up_axis)
    axis4 = tf_vector(drawer_grasp_rot, drawer_up_axis)

    dot1 = torch.bmm(axis1.view(num_envs, 1, 3), axis2.view(num_envs, 3, 1)).squeeze(-1).squeeze(-1)  # alignment of forward axis for gripper
    dot2 = torch.bmm(axis3.view(num_envs, 1, 3), axis4.view(num_envs, 3, 1)).squeeze(-1).squeeze(-1)  # alignment of up axis for gripper
    # reward for matching the orientation of the hand to the drawer (fingers wrapped)
    rot_reward = 0.5 * (torch.sign(dot1) * dot1 ** 2 + torch.sign(dot2) * dot2 ** 2)

    # bonus if left finger is above the drawer handle and right below
    around_handle_reward = torch.zeros_like(rot_reward)
    around_handle_reward = torch.where(franka_lfinger_pos[:, 2] > drawer_grasp_pos[:, 2],
                                       torch.where(franka_rfinger_pos[:, 2] < drawer_grasp_pos[:, 2],
                                                   around_handle_reward + 0.5, around_handle_reward), around_handle_reward)
    # reward for distance of each finger from the drawer
    finger_dist_reward = torch.zeros_like(rot_reward)
    lfinger_dist = torch.abs(franka_lfinger_pos[:, 2] - drawer_grasp_pos[:, 2])
    rfinger_dist = torch.abs(franka_rfinger_pos[:, 2] - drawer_grasp_pos[:, 2])
    finger_dist_reward = torch.where(franka_lfinger_pos[:, 2] > drawer_grasp_pos[:, 2],
                                     torch.where(franka_rfinger_pos[:, 2] < drawer_grasp_pos[:, 2],
                                                 (0.04 - lfinger_dist) + (0.04 - rfinger_dist), finger_dist_reward), finger_dist_reward)

    # regularization on the actions (summed for each environment)
    action_penalty = torch.sum(actions ** 2, dim=-1)

    # how far the cabinet has been opened out
    open_reward = cabinet_dof_pos[:, 3] * around_handle_reward + cabinet_dof_pos[:, 3]  # drawer_top_joint

    rewards = dist_reward_scale * dist_reward + rot_reward_scale * rot_reward \
        + around_handle_reward_scale * around_handle_reward + open_reward_scale * open_reward \
        + finger_dist_reward_scale * finger_dist_reward - action_penalty_scale * action_penalty

    # bonus for opening drawer properly
    rewards = torch.where(cabinet_dof_pos[:, 3] > 0.01, rewards + 0.5, rewards)
    rewards = torch.where(cabinet_dof_pos[:, 3] > 0.2, rewards + around_handle_reward, rewards)
    rewards = torch.where(cabinet_dof_pos[:, 3] > 0.39, rewards + (2.0 * around_handle_reward), rewards)

    # prevent bad style in opening drawer
    rewards = torch.where(franka_lfinger_pos[:, 0] < drawer_grasp_pos[:, 0] - distX_offset,
                          torch.ones_like(rewards) * -1, rewards)
    rewards = torch.where(franka_rfinger_pos[:, 0] < drawer_grasp_pos[:, 0] - distX_offset,
                          torch.ones_like(rewards) * -1, rewards)
    
    # reset if drawer is open or max length reached
    successes = torch.where(cabinet_dof_pos[:, 3] > 0.39, torch.ones_like(successes), successes)
    goal_reach = torch.where(cabinet_dof_pos[:, 3] > 0.39, torch.ones_like(reset_goal_buf), torch.zeros_like(reset_goal_buf))
    reset_buf = torch.where(progress_buf >= max_episode_length - 1, torch.ones_like(reset_buf), reset_buf)

    consecutive_successes = torch.where(reset_buf > 0, successes * reset_buf, consecutive_successes)
    heuristic_reward, task_reward = rewards, goal_reach
    return heuristic_reward, reset_buf, task_reward, consecutive_successes
\end{minted}
\end{figure}

\begin{figure}[H]
\centering
\caption*{Isaac Gym - Manipulation Task: \textbf{\textit{FrankaCubeStack}}}
\begin{minted}[mathescape, breaklines,frame=single, fontsize=\scriptsize]{python}
def compute_franka_reward(
    reset_buf, progress_buf, reset_goal_buf, actions, states, reward_settings, max_episode_length
):
    # type: (Tensor, Tensor, Tensor, Tensor, Dict[str, Tensor], Dict[str, float], float) -> Tuple[Tensor, Tensor, Tensor]

    # Compute per-env physical parameters
    target_height = states["cubeB_size"] + states["cubeA_size"] / 2.0
    cubeA_size = states["cubeA_size"]
    cubeB_size = states["cubeB_size"]

    # distance from hand to the cubeA
    d = torch.norm(states["cubeA_pos_relative"], dim=-1)
    d_lf = torch.norm(states["cubeA_pos"] - states["eef_lf_pos"], dim=-1)
    d_rf = torch.norm(states["cubeA_pos"] - states["eef_rf_pos"], dim=-1)
    dist_reward = 1 - torch.tanh(10.0 * (d + d_lf + d_rf) / 3)

    # reward for lifting cubeA
    cubeA_height = states["cubeA_pos"][:, 2] - reward_settings["table_height"]
    cubeA_lifted = (cubeA_height - cubeA_size) > 0.04
    lift_reward = cubeA_lifted

    # how closely aligned cubeA is to cubeB (only provided if cubeA is lifted)
    offset = torch.zeros_like(states["cubeA_to_cubeB_pos"])
    offset[:, 2] = (cubeA_size + cubeB_size) / 2
    d_ab = torch.norm(states["cubeA_to_cubeB_pos"] + offset, dim=-1)
    align_reward = (1 - torch.tanh(10.0 * d_ab)) * cubeA_lifted

    # Dist reward is maximum of dist and align reward
    dist_reward = torch.max(dist_reward, align_reward)

    # final reward for stacking successfully (only if cubeA is close to target height and corresponding location, and gripper is not grasping)
    cubeA_align_cubeB = (torch.norm(states["cubeA_to_cubeB_pos"][:, :2], dim=-1) < 0.02)
    cubeA_on_cubeB = torch.abs(cubeA_height - target_height) < 0.02
    gripper_away_from_cubeA = (d > 0.04)
    stack_reward = cubeA_align_cubeB & cubeA_on_cubeB & gripper_away_from_cubeA

    # Compose rewards

    # We either provide the stack reward or the align + dist reward
    rewards = torch.where(
        stack_reward,
        reward_settings["r_stack_scale"] * stack_reward,
        reward_settings["r_dist_scale"] * dist_reward + reward_settings["r_lift_scale"] * lift_reward + reward_settings[
            "r_align_scale"] * align_reward,
    )

    # Compute resets
    reset_buf = torch.where((progress_buf >= max_episode_length - 1), torch.ones_like(reset_buf), reset_buf)
    goal_reach = torch.where(stack_reward > 0, torch.ones_like(reset_goal_buf), torch.zeros_like(reset_goal_buf))
    heuristic_reward, task_reward = rewards, goal_reach
    return heuristic_reward, task_reward, reset_buf
\end{minted}
\end{figure}

\begin{figure}[H]
\centering
\caption*{Isaac Gym - Manipulation Task: \textbf{\textit{ShadowHand}}}
\begin{minted}[mathescape, breaklines,frame=single, fontsize=\scriptsize]{python}
def compute_hand_reward(
    rew_buf, reset_buf, reset_goal_buf, progress_buf, successes, consecutive_successes,
    max_episode_length: float, object_pos, object_rot, target_pos, target_rot,
    dist_reward_scale: float, rot_reward_scale: float, rot_eps: float,
    actions, action_penalty_scale: float,
    success_tolerance: float, reach_goal_bonus: float, fall_dist: float,
    fall_penalty: float, max_consecutive_successes: int, av_factor: float, ignore_z_rot: bool
):
    # Distance from the hand to the object
    goal_dist = torch.norm(object_pos - target_pos, p=2, dim=-1)

    if ignore_z_rot:
        success_tolerance = 2.0 * success_tolerance

    # Orientation alignment for the cube in hand and goal cube
    quat_diff = quat_mul(object_rot, quat_conjugate(target_rot))
    rot_dist = 2.0 * torch.asin(torch.clamp(torch.norm(quat_diff[:, 0:3], p=2, dim=-1), max=1.0))

    dist_rew = goal_dist * dist_reward_scale
    rot_rew = 1.0/(torch.abs(rot_dist) + rot_eps) * rot_reward_scale

    action_penalty = torch.sum(actions ** 2, dim=-1)

    # Total reward is: position distance + orientation alignment + action regularization + success bonus + fall penalty
    reward = dist_rew + action_penalty * action_penalty_scale

    # Find out which envs hit the goal and update successes count
    goal_reach = torch.where(torch.abs(rot_dist) <= success_tolerance, torch.ones_like(reset_goal_buf), reset_goal_buf)
    successes = successes + goal_reach

    # Success bonus: orientation is within `success_tolerance` of goal orientation
    reward = torch.where(goal_reach == 1, reward + reach_goal_bonus, reward)

    # Fall penalty: distance to the goal is larger than a threshold
    reward = torch.where(goal_dist >= fall_dist, reward + fall_penalty, reward)

    # Check env termination conditions, including maximum success number
    resets = torch.where(goal_dist >= fall_dist, torch.ones_like(reset_buf), reset_buf)
    if max_consecutive_successes > 0:
        # Reset progress buffer on goal envs if max_consecutive_successes > 0
        progress_buf = torch.where(torch.abs(rot_dist) <= success_tolerance, torch.zeros_like(progress_buf), progress_buf)
        resets = torch.where(successes >= max_consecutive_successes, torch.ones_like(resets), resets)
    resets = torch.where(progress_buf >= max_episode_length - 1, torch.ones_like(resets), resets)

    # Apply penalty for not reaching the goal
    if max_consecutive_successes > 0:
        reward = torch.where(progress_buf >= max_episode_length - 1, reward + 0.5 * fall_penalty, reward)

    num_resets = torch.sum(resets)
    finished_cons_successes = torch.sum(successes * resets.float())

    cons_successes = torch.where(num_resets > 0, av_factor*finished_cons_successes/num_resets + (1.0 - av_factor)*consecutive_successes, consecutive_successes)
    heuristic_reward, task_reward = reward, goal_reach
    return heuristic_reward, resets, task_reward, progress_buf, successes, cons_successes
\end{minted}
\end{figure}

\begin{figure}[H]
\centering
\caption*{Isaac Gym - Manipulation Task: \textbf{\textit{AllegroHand}}}
\begin{minted}[mathescape, breaklines,frame=single, fontsize=\scriptsize]{python}
def compute_hand_reward(
    rew_buf, reset_buf, reset_goal_buf, progress_buf, successes, consecutive_successes,
    max_episode_length: float, object_pos, object_rot, target_pos, target_rot,
    dist_reward_scale: float, rot_reward_scale: float, rot_eps: float,
    actions, action_penalty_scale: float,
    success_tolerance: float, reach_goal_bonus: float, fall_dist: float,
    fall_penalty: float, max_consecutive_successes: int, av_factor: float, ignore_z_rot: bool
):
    # Distance from the hand to the object
    goal_dist = torch.norm(object_pos - target_pos, p=2, dim=-1)

    if ignore_z_rot:
        success_tolerance = 2.0 * success_tolerance

    # Orientation alignment for the cube in hand and goal cube
    quat_diff = quat_mul(object_rot, quat_conjugate(target_rot))
    rot_dist = 2.0 * torch.asin(torch.clamp(torch.norm(quat_diff[:, 0:3], p=2, dim=-1), max=1.0))

    dist_rew = goal_dist * dist_reward_scale
    rot_rew = 1.0/(torch.abs(rot_dist) + rot_eps) * rot_reward_scale

    action_penalty = torch.sum(actions ** 2, dim=-1)

    # Total reward is: position distance + orientation alignment + action regularization + success bonus + fall penalty
    reward = dist_rew + rot_rew + action_penalty * action_penalty_scale

    # Find out which envs hit the goal and update successes count
    goal_reach = torch.where(torch.abs(rot_dist) <= success_tolerance, torch.ones_like(reset_goal_buf), reset_goal_buf)
    successes = successes + goal_reach

    # Success bonus: orientation is within `success_tolerance` of goal orientation
    reward = torch.where(goal_reach == 1, reward + reach_goal_bonus, reward)

    # Fall penalty: distance to the goal is larger than a threshold
    reward = torch.where(goal_dist >= fall_dist, reward + fall_penalty, reward)

    # Check env termination conditions, including maximum success number
    resets = torch.where(goal_dist >= fall_dist, torch.ones_like(reset_buf), reset_buf)
    if max_consecutive_successes > 0:
        # Reset progress buffer on goal envs if max_consecutive_successes > 0
        progress_buf = torch.where(torch.abs(rot_dist) <= success_tolerance, torch.zeros_like(progress_buf), progress_buf)
        resets = torch.where(successes >= max_consecutive_successes, torch.ones_like(resets), resets)

    timed_out = progress_buf >= max_episode_length - 1
    resets = torch.where(timed_out, torch.ones_like(resets), resets)

    # Apply penalty for not reaching the goal
    if max_consecutive_successes > 0:
        reward = torch.where(timed_out, reward + 0.5 * fall_penalty, reward)

    num_resets = torch.sum(resets)
    finished_cons_successes = torch.sum(successes * resets.float())

    cons_successes = torch.where(num_resets > 0, av_factor*finished_cons_successes/num_resets + (1.0 - av_factor)*consecutive_successes, consecutive_successes)
    heuristic_reward, task_reward = reward, goal_reach
    return heuristic_reward, resets, task_reward, progress_buf, successes, cons_successes
\end{minted}
\end{figure}

\begin{figure}[H]
\centering
\caption*{Bi-DexHands: \textbf{\textit{ShadowHandOver}}}
\begin{minted}[mathescape, breaklines,frame=single, fontsize=\scriptsize]{python}
def compute_hand_reward(
    rew_buf, reset_buf, reset_goal_buf, progress_buf, successes, consecutive_successes,
    max_episode_length: float, object_pos, object_rot, target_pos, target_rot,
    dist_reward_scale: float, rot_reward_scale: float, rot_eps: float,
    actions, action_penalty_scale: float,
    success_tolerance: float, reach_goal_bonus: float, fall_dist: float,
    fall_penalty: float, max_consecutive_successes: int, av_factor: float, ignore_z_rot: bool
):
    # Distance from the hand to the object
    goal_dist = torch.norm(target_pos - object_pos, p=2, dim=-1)
    if ignore_z_rot:
        success_tolerance = 2.0 * success_tolerance

    # Orientation alignment for the cube in hand and goal cube
    quat_diff = quat_mul(object_rot, quat_conjugate(target_rot))
    rot_dist = 2.0 * torch.asin(torch.clamp(torch.norm(quat_diff[:, 0:3], p=2, dim=-1), max=1.0))

    dist_rew = goal_dist
    
    # Total reward is: position distance + orientation alignment + action regularization + success bonus + fall penalty
    reward = torch.exp(-0.2*(dist_rew * dist_reward_scale + rot_dist))

    # Find out which envs hit the goal and update successes count
    goal_resets = torch.where(torch.abs(goal_dist) <= 0, torch.ones_like(reset_goal_buf), reset_goal_buf)
    successes = torch.where(successes == 0, 
                    torch.where(goal_dist < 0.03, torch.ones_like(successes), successes), successes)

    # Success bonus: orientation is within `success_tolerance` of goal orientation
    reward = torch.where(goal_resets == 1, reward + reach_goal_bonus, reward)

    # Fall penalty: distance to the goal is larger than a threashold
    reward = torch.where(object_pos[:, 2] <= 0.2, reward + fall_penalty, reward)

    # Check env termination conditions, including maximum success number
    resets = torch.where(object_pos[:, 2] <= 0.2, torch.ones_like(reset_buf), reset_buf)
    if max_consecutive_successes > 0:
        # Reset progress buffer on goal envs if max_consecutive_successes > 0
        progress_buf = torch.where(torch.abs(rot_dist) <= success_tolerance, torch.zeros_like(progress_buf), progress_buf)
        resets = torch.where(successes >= max_consecutive_successes, torch.ones_like(resets), resets)
    resets = torch.where(progress_buf >= max_episode_length, torch.ones_like(resets), resets)

    # Apply penalty for not reaching the goal
    if max_consecutive_successes > 0:
        reward = torch.where(progress_buf >= max_episode_length, reward + 0.5 * fall_penalty, reward)

    cons_successes = torch.where(resets > 0, successes * resets, consecutive_successes).mean()
    
    goal_reach = torch.where(goal_dist <= 0.03, torch.ones_like(successes), torch.zeros_like(successes))
    heuristic_reward, task_reward = reward, goal_reach
    return heuristic_reward, task_reward, resets, goal_resets, progress_buf, successes, cons_successes
\end{minted}
\end{figure}

\begin{figure}[H]
\centering
\caption*{Bi-DexHands: \textbf{\textit{ShadowHandCatchUnderarm}}}
\begin{minted}[mathescape, breaklines,frame=single, fontsize=\scriptsize]{python}
def compute_hand_reward(
    rew_buf, reset_buf, reset_goal_buf, progress_buf, successes, consecutive_successes,
    max_episode_length: float, object_pos, object_rot, target_pos, target_rot,
    dist_reward_scale: float, rot_reward_scale: float, rot_eps: float,
    actions, action_penalty_scale: float,
    success_tolerance: float, reach_goal_bonus: float, fall_dist: float,
    fall_penalty: float, max_consecutive_successes: int, av_factor: float, ignore_z_rot: bool
):
    # Distance from the hand to the object
    goal_dist = torch.norm(target_pos - object_pos, p=2, dim=-1)

    if ignore_z_rot:
        success_tolerance = 2.0 * success_tolerance

    # Orientation alignment for the cube in hand and goal cube
    quat_diff = quat_mul(object_rot, quat_conjugate(target_rot))
    rot_dist = 2.0 * torch.asin(torch.clamp(torch.norm(quat_diff[:, 0:3], p=2, dim=-1), max=1.0))

    dist_rew = goal_dist

    # Total reward is: position distance + orientation alignment + action regularization + success bonus + fall penalty
    reward = torch.exp(-0.2*(dist_rew * dist_reward_scale + rot_dist))

    # Find out which envs hit the goal and update successes count
    goal_resets = torch.where(torch.abs(goal_dist) <= 0, torch.ones_like(reset_goal_buf), reset_goal_buf)
    successes = torch.where(successes == 0, 
                    torch.where(goal_dist < 0.03, torch.ones_like(successes), successes), successes)

    # Fall penalty: distance to the goal is larger than a threashold
    reward = torch.where(object_pos[:, 2] <= 0.1, reward + fall_penalty, reward)

    # Check env termination conditions, including maximum success number
    resets = torch.where(object_pos[:, 2] <= 0.1, torch.ones_like(reset_buf), reset_buf)
    if max_consecutive_successes > 0:
        # Reset progress buffer on goal envs if max_consecutive_successes > 0
        progress_buf = torch.where(torch.abs(rot_dist) <= success_tolerance, torch.zeros_like(progress_buf), progress_buf)
        resets = torch.where(successes >= max_consecutive_successes, torch.ones_like(resets), resets)
    resets = torch.where(progress_buf >= max_episode_length, torch.ones_like(resets), resets)

    # Apply penalty for not reaching the goal
    if max_consecutive_successes > 0:
        reward = torch.where(progress_buf >= max_episode_length, reward + 0.5 * fall_penalty, reward)

    cons_successes = torch.where(resets > 0, successes * resets, consecutive_successes)
    goal_reach = torch.where(goal_dist <= 0.03, 
                             torch.ones_like(successes), torch.zeros_like(successes))
    heuristic_reward, task_reward = reward, goal_reach
    return heuristic_reward, task_reward, resets, goal_resets, progress_buf, successes, cons_successes
\end{minted}
\end{figure}

\begin{figure}[H]
\centering
\caption*{Bi-DexHands: \textbf{\textit{ShadowHandCatchOver2Underarm}}}
\begin{minted}[mathescape, breaklines,frame=single, fontsize=\scriptsize]{python}
def compute_hand_reward(
    rew_buf, reset_buf, reset_goal_buf, progress_buf, successes, consecutive_successes,
    max_episode_length: float, object_pos, object_rot, target_pos, target_rot, left_hand_base_pos, right_hand_base_pos,
    dist_reward_scale: float, rot_reward_scale: float, rot_eps: float,
    actions, action_penalty_scale: float,
    success_tolerance: float, reach_goal_bonus: float, fall_dist: float,
    fall_penalty: float, max_consecutive_successes: int, av_factor: float, ignore_z_rot: bool, device: str
):    
    # Distance from the hand to the object
    goal_dist = torch.norm(target_pos - object_pos, p=2, dim=-1)
    if ignore_z_rot:
        success_tolerance = 2.0 * success_tolerance

    # Orientation alignment for the cube in hand and goal cube
    quat_diff = quat_mul(object_rot, quat_conjugate(target_rot))
    rot_dist = 2.0 * torch.asin(torch.clamp(torch.norm(quat_diff[:, 0:3], p=2, dim=-1), max=1.0))

    dist_rew = goal_dist
   
    # Total reward is: position distance + orientation alignment + action regularization + success bonus + fall penalty
    reward = torch.exp(-0.2*(dist_rew * dist_reward_scale + rot_dist))

    # Find out which envs hit the goal and update successes count
    goal_resets = torch.where(torch.abs(goal_dist) <= 0, torch.ones_like(reset_goal_buf), reset_goal_buf)
    successes = torch.where(successes == 0, 
                    torch.where(goal_dist < 0.03, torch.ones_like(successes), successes), successes)

    # Check env termination conditions, including maximum success number
    right_hand_base_dist = torch.norm(right_hand_base_pos - torch.tensor([0.0, 0.0, 0.5], dtype=torch.float, device=device), p=2, dim=-1)
    left_hand_base_dist = torch.norm(left_hand_base_pos - torch.tensor([0.0, -0.8, 0.5], dtype=torch.float, device=device), p=2, dim=-1)

    resets = torch.where(right_hand_base_dist >= 0.1, torch.ones_like(reset_buf), reset_buf)
    resets = torch.where(left_hand_base_dist >= 0.1, torch.ones_like(resets), resets)
    resets = torch.where(object_pos[:, 2] <= 0.3, torch.ones_like(resets), resets)
    if max_consecutive_successes > 0:
        # Reset progress buffer on goal envs if max_consecutive_successes > 0
        progress_buf = torch.where(torch.abs(rot_dist) <= success_tolerance, torch.zeros_like(progress_buf), progress_buf)
        resets = torch.where(successes >= max_consecutive_successes, torch.ones_like(resets), resets)
    resets = torch.where(progress_buf >= max_episode_length, torch.ones_like(resets), resets)

    # Apply penalty for not reaching the goal
    if max_consecutive_successes > 0:
        reward = torch.where(progress_buf >= max_episode_length, reward + 0.5 * fall_penalty, reward)

    cons_successes = torch.where(resets > 0, successes * resets, consecutive_successes)
    goal_reach = torch.where(goal_dist <= 0.03, 
                             torch.ones_like(successes), torch.zeros_like(successes))
    heuristic_reward, task_reward = reward, goal_reach
    return heuristic_reward, task_reward, resets, goal_resets, progress_buf, successes, cons_successes
\end{minted}
\end{figure}

\begin{figure}[H]
\centering
\caption*{Bi-DexHands: \textbf{\textit{ShadowHandCatchAbreast}}}
\begin{minted}[mathescape, breaklines,frame=single, fontsize=\scriptsize]{python}
def compute_hand_reward(
    rew_buf, reset_buf, reset_goal_buf, progress_buf, successes, consecutive_successes,
    max_episode_length: float, object_pos, object_rot, target_pos, target_rot, left_hand_pos, right_hand_pos, left_hand_base_pos, right_hand_base_pos,
    dist_reward_scale: float, rot_reward_scale: float, rot_eps: float,
    actions, action_penalty_scale: float,
    success_tolerance: float, reach_goal_bonus: float, fall_dist: float,
    fall_penalty: float, max_consecutive_successes: int, av_factor: float, ignore_z_rot: bool, device: str
):
    # Distance from the hand to the object
    goal_dist = torch.norm(target_pos - object_pos, p=2, dim=-1)
    
    if ignore_z_rot:
        success_tolerance = 2.0 * success_tolerance

    # Orientation alignment for the cube in hand and goal cube
    quat_diff = quat_mul(object_rot, quat_conjugate(target_rot))
    rot_dist = 2.0 * torch.asin(torch.clamp(torch.norm(quat_diff[:, 0:3], p=2, dim=-1), max=1.0))

    dist_rew = goal_dist
    
    # Total reward is: position distance + orientation alignment + action regularization + success bonus + fall penalty
    reward = torch.exp(-0.2*(dist_rew * dist_reward_scale + rot_dist))
    
    # Find out which envs hit the goal and update successes count
    goal_resets = torch.where(torch.abs(goal_dist) <= 0, torch.ones_like(reset_goal_buf), reset_goal_buf)

    successes = torch.where(successes == 0, 
                    torch.where(goal_dist < 0.03, torch.ones_like(successes), successes), successes)

    # Fall penalty: distance to the goal is larger than a threashold
    reward = torch.where(object_pos[:, 2] <= 0.2, reward + fall_penalty, reward)

    # Check env termination conditions, including maximum success number
    right_hand_base_dist = torch.norm(right_hand_base_pos - torch.tensor([-0.3, -0.55, 0.5], dtype=torch.float, device=device), p=2, dim=-1)
    left_hand_base_dist = torch.norm(left_hand_base_pos - torch.tensor([-0.3, -1.15, 0.5], dtype=torch.float, device=device), p=2, dim=-1)

    resets = torch.where(right_hand_base_dist >= 0.1, torch.ones_like(reset_buf), reset_buf)
    resets = torch.where(left_hand_base_dist >= 0.1, torch.ones_like(resets), resets)

    resets = torch.where(object_pos[:, 2] <= 0.2, torch.ones_like(resets), resets)
    if max_consecutive_successes > 0:
        # Reset progress buffer on goal envs if max_consecutive_successes > 0
        progress_buf = torch.where(torch.abs(rot_dist) <= success_tolerance, torch.zeros_like(progress_buf), progress_buf)
        resets = torch.where(successes >= max_consecutive_successes, torch.ones_like(resets), resets)
    resets = torch.where(progress_buf >= max_episode_length, torch.ones_like(resets), resets)

    # Apply penalty for not reaching the goal
    if max_consecutive_successes > 0:
        reward = torch.where(progress_buf >= max_episode_length, reward + 0.5 * fall_penalty, reward)

    cons_successes = torch.where(resets > 0, successes * resets, consecutive_successes)
    goal_reach = torch.where(goal_dist <= 0.03, 
                             torch.ones_like(successes), torch.zeros_like(successes))
    heuristic_reward, task_reward = reward, goal_reach
    return heuristic_reward, task_reward, resets, goal_resets, progress_buf, successes, cons_successes
\end{minted}
\end{figure}

\begin{figure}[H]
\centering
\caption*{Bi-DexHands: \textbf{\textit{ShadowHandTwoCatchUnderarm}}}
\begin{minted}[mathescape, breaklines,frame=single, fontsize=\scriptsize]{python}
def compute_hand_reward(
    rew_buf, reset_buf, reset_goal_buf, progress_buf, successes, consecutive_successes,
    max_episode_length: float, object_pos, object_rot, target_pos, target_rot, object_another_pos, object_another_rot, target_another_pos, target_another_rot,
    dist_reward_scale: float, rot_reward_scale: float, rot_eps: float,
    actions, action_penalty_scale: float,
    success_tolerance: float, reach_goal_bonus: float, fall_dist: float,
    fall_penalty: float, max_consecutive_successes: int, av_factor: float, ignore_z_rot: bool
):
    # Distance from the hand to the object
    goal_dist = torch.norm(target_pos - object_pos, p=2, dim=-1)
    if ignore_z_rot:
        success_tolerance = 2.0 * success_tolerance

    goal_another_dist = torch.norm(target_another_pos - object_another_pos, p=2, dim=-1)
    if ignore_z_rot:
        success_tolerance = 2.0 * success_tolerance

    # Orientation alignment for the cube in hand and goal cube
    quat_diff = quat_mul(object_rot, quat_conjugate(target_rot))
    rot_dist = 2.0 * torch.asin(torch.clamp(torch.norm(quat_diff[:, 0:3], p=2, dim=-1), max=1.0))

    quat_another_diff = quat_mul(object_another_rot, quat_conjugate(target_another_rot))
    rot_another_dist = 2.0 * torch.asin(torch.clamp(torch.norm(quat_another_diff[:, 0:3], p=2, dim=-1), max=1.0))

    dist_rew = goal_dist
 
    # Total reward is: position distance + orientation alignment + action regularization + success bonus + fall penalty
    reward = torch.exp(-0.2*(dist_rew * dist_reward_scale + rot_dist)) + torch.exp(-0.2*(goal_another_dist * dist_reward_scale + rot_another_dist))

    # Find out which envs hit the goal and update successes count
    goal_resets = torch.where(torch.abs(goal_dist) <= 0, torch.ones_like(reset_goal_buf), reset_goal_buf)
    successes = torch.where(successes == 0, 
                    torch.where(goal_dist + goal_another_dist < 0.06, torch.ones_like(successes), successes), successes)

    # Fall penalty: distance to the goal is larger than a threashold
    reward = torch.where(object_pos[:, 2] <= 0.2, reward + fall_penalty, reward)
    reward = torch.where(object_another_pos[:, 2] <= 0.2, reward + fall_penalty, reward)

    # Check env termination conditions, including maximum success number
    resets = torch.where(object_pos[:, 2] <= 0.2, torch.ones_like(reset_buf), reset_buf)
    resets = torch.where(object_another_pos[:, 2] <= 0.2, torch.ones_like(reset_buf), resets)

    if max_consecutive_successes > 0:
        # Reset progress buffer on goal envs if max_consecutive_successes > 0
        progress_buf = torch.where(torch.abs(rot_dist) <= success_tolerance, torch.zeros_like(progress_buf), progress_buf)
        resets = torch.where(successes >= max_consecutive_successes, torch.ones_like(resets), resets)
    resets = torch.where(progress_buf >= max_episode_length, torch.ones_like(resets), resets)

    # Apply penalty for not reaching the goal
    if max_consecutive_successes > 0:
        reward = torch.where(progress_buf >= max_episode_length, reward + 0.5 * fall_penalty, reward)

    cons_successes = torch.where(resets > 0, successes * resets, consecutive_successes).mean()
    goal_reach = torch.where(goal_dist + goal_another_dist <= 0.06, 
                             torch.ones_like(successes), torch.zeros_like(successes))
    heuristic_reward, task_reward = reward, goal_reach
    return heuristic_reward, task_reward, resets, goal_resets, progress_buf, successes, cons_successes
\end{minted}
\end{figure}

\begin{figure}[H]
\centering
\caption*{Bi-DexHands: \textbf{\textit{ShadowHandLiftUnderarm}}}
\begin{minted}[mathescape, breaklines,frame=single, fontsize=\scriptsize]{python}
def compute_hand_reward(
    rew_buf, reset_buf, reset_goal_buf, progress_buf, successes, consecutive_successes,
    max_episode_length: float, object_pos, object_rot, target_pos, target_rot, pot_left_handle_pos, pot_right_handle_pos, 
    left_hand_pos, right_hand_pos, 
    dist_reward_scale: float, rot_reward_scale: float, rot_eps: float,
    actions, action_penalty_scale: float,
    success_tolerance: float, reach_goal_bonus: float, fall_dist: float,
    fall_penalty: float, max_consecutive_successes: int, av_factor: float, ignore_z_rot: bool
):
    # Distance from the hand to the object
    goal_dist = torch.norm(target_pos - object_pos, p=2, dim=-1)
    
    # goal_dist = target_pos[:, 2] - object_pos[:, 2]
    right_hand_dist = torch.norm(pot_right_handle_pos - right_hand_pos, p=2, dim=-1)
    left_hand_dist = torch.norm(pot_left_handle_pos - left_hand_pos, p=2, dim=-1)

    right_hand_dist_rew = right_hand_dist
    left_hand_dist_rew = left_hand_dist

    # Total reward is: position distance + orientation alignment + action regularization + success bonus + fall penalty
    up_rew = torch.zeros_like(right_hand_dist_rew)
    up_rew = torch.where(right_hand_dist < 0.08,
                        torch.where(left_hand_dist < 0.08,
                                        3*(0.385 - goal_dist), up_rew), up_rew)
    
    reward = 0.2 - right_hand_dist_rew - left_hand_dist_rew + up_rew

    resets = torch.where(object_pos[:, 2] <= 0.3, torch.ones_like(reset_buf), reset_buf)
    resets = torch.where(right_hand_dist >= 0.2, torch.ones_like(resets), resets)
    resets = torch.where(left_hand_dist >= 0.2, torch.ones_like(resets), resets)

    # Find out which envs hit the goal and update successes count
    successes = torch.where(successes == 0, 
                    torch.where(goal_dist < 0.05, torch.ones_like(successes), successes), successes)

    resets = torch.where(progress_buf >= max_episode_length, torch.ones_like(resets), resets)

    goal_resets = torch.zeros_like(resets)

    cons_successes = torch.where(resets > 0, successes * resets, consecutive_successes).mean()
    goal_reach = torch.where(goal_dist <= 0.05, 
                             torch.ones_like(successes), torch.zeros_like(successes))
    heuristic_reward, task_reward = reward, goal_reach     return heuristic_reward, task_reward, resets, goal_resets, progress_buf, successes, cons_successes
\end{minted}
\end{figure}

\begin{figure}[H]
\centering
\caption*{Bi-DexHands: \textbf{\textit{ShadowHandDoorOpenInward}}}
\begin{minted}[mathescape, breaklines,frame=single, fontsize=\scriptsize]{python}
def compute_hand_reward(
    rew_buf, reset_buf, reset_goal_buf, progress_buf, successes, consecutive_successes,
    max_episode_length: float, object_pos, object_rot, target_pos, target_rot, door_left_handle_pos, door_right_handle_pos,
    left_hand_pos, right_hand_pos, right_hand_ff_pos, right_hand_mf_pos, right_hand_rf_pos, right_hand_lf_pos, right_hand_th_pos,
    left_hand_ff_pos, left_hand_mf_pos, left_hand_rf_pos, left_hand_lf_pos, left_hand_th_pos,
    dist_reward_scale: float, rot_reward_scale: float, rot_eps: float,
    actions, action_penalty_scale: float,
    success_tolerance: float, reach_goal_bonus: float, fall_dist: float,
    fall_penalty: float, max_consecutive_successes: int, av_factor: float, ignore_z_rot: bool
):
    # Distance from the hand to the object
    goal_dist = torch.norm(target_pos - object_pos, p=2, dim=-1)
    
    right_hand_finger_dist = (torch.norm(door_right_handle_pos - right_hand_ff_pos, p=2, dim=-1) + torch.norm(door_right_handle_pos - right_hand_mf_pos, p=2, dim=-1)
                            + torch.norm(door_right_handle_pos - right_hand_rf_pos, p=2, dim=-1) + torch.norm(door_right_handle_pos - right_hand_lf_pos, p=2, dim=-1) 
                            + torch.norm(door_right_handle_pos - right_hand_th_pos, p=2, dim=-1))
    left_hand_finger_dist = (torch.norm(door_left_handle_pos - left_hand_ff_pos, p=2, dim=-1) + torch.norm(door_left_handle_pos - left_hand_mf_pos, p=2, dim=-1)
                            + torch.norm(door_left_handle_pos - left_hand_rf_pos, p=2, dim=-1) + torch.norm(door_left_handle_pos - left_hand_lf_pos, p=2, dim=-1) 
                            + torch.norm(door_left_handle_pos - left_hand_th_pos, p=2, dim=-1))
    
    right_hand_dist_rew = right_hand_finger_dist
    left_hand_dist_rew = left_hand_finger_dist

    # Total reward is: position distance + orientation alignment + action regularization + success bonus + fall penalty
    up_rew = torch.zeros_like(right_hand_dist_rew)
    up_rew = torch.where(right_hand_finger_dist < 0.5,
                    torch.where(left_hand_finger_dist < 0.5,
                                    torch.abs(door_right_handle_pos[:, 1] - door_left_handle_pos[:, 1]) * 2, up_rew), up_rew)

    reward = 2 - right_hand_dist_rew - left_hand_dist_rew + up_rew

    resets = torch.where(right_hand_finger_dist >= 1.5, torch.ones_like(reset_buf), reset_buf)
    resets = torch.where(left_hand_finger_dist >= 1.5, torch.ones_like(resets), resets)

    # Find out which envs hit the goal and update successes count
    successes = torch.where(successes == 0, 
                    torch.where(torch.abs(door_right_handle_pos[:, 1] - door_left_handle_pos[:, 1]) > 0.5, torch.ones_like(successes), successes), successes)

    resets = torch.where(progress_buf >= max_episode_length, torch.ones_like(resets), resets)

    goal_resets = torch.zeros_like(resets)

    cons_successes = torch.where(resets > 0, successes * resets, consecutive_successes).mean()
    goal_reach = torch.where(torch.abs(door_right_handle_pos[:, 1] - door_left_handle_pos[:, 1]) >= 0.5, 
                             torch.ones_like(successes), torch.zeros_like(successes))
    heuristic_reward, task_reward = reward, goal_reach     return heuristic_reward, task_reward, resets, goal_resets, progress_buf, successes, cons_successes
\end{minted}
\end{figure}

\begin{figure}[H]
\centering
\caption*{Bi-DexHands: \textbf{\textit{ShadowHandDoorOpenOutward}}}
\begin{minted}[mathescape, breaklines,frame=single, fontsize=\scriptsize]{python}
def compute_hand_reward(
    rew_buf, reset_buf, reset_goal_buf, progress_buf, successes, consecutive_successes,
    max_episode_length: float, object_pos, object_rot, target_pos, target_rot, door_left_handle_pos, door_right_handle_pos,
    left_hand_pos, right_hand_pos, right_hand_ff_pos, right_hand_mf_pos, right_hand_rf_pos, right_hand_lf_pos, right_hand_th_pos,
    left_hand_ff_pos, left_hand_mf_pos, left_hand_rf_pos, left_hand_lf_pos, left_hand_th_pos,
    dist_reward_scale: float, rot_reward_scale: float, rot_eps: float,
    actions, action_penalty_scale: float,
    success_tolerance: float, reach_goal_bonus: float, fall_dist: float,
    fall_penalty: float, max_consecutive_successes: int, av_factor: float, ignore_z_rot: bool
):
    right_hand_finger_dist = (torch.norm(door_right_handle_pos - right_hand_ff_pos, p=2, dim=-1) + torch.norm(door_right_handle_pos - right_hand_mf_pos, p=2, dim=-1)
                            + torch.norm(door_right_handle_pos - right_hand_rf_pos, p=2, dim=-1) + torch.norm(door_right_handle_pos - right_hand_lf_pos, p=2, dim=-1) 
                            + torch.norm(door_right_handle_pos - right_hand_th_pos, p=2, dim=-1))
    left_hand_finger_dist = (torch.norm(door_left_handle_pos - left_hand_ff_pos, p=2, dim=-1) + torch.norm(door_left_handle_pos - left_hand_mf_pos, p=2, dim=-1)
                            + torch.norm(door_left_handle_pos - left_hand_rf_pos, p=2, dim=-1) + torch.norm(door_left_handle_pos - left_hand_lf_pos, p=2, dim=-1) 
                            + torch.norm(door_left_handle_pos - left_hand_th_pos, p=2, dim=-1))
    
    right_hand_dist_rew = right_hand_finger_dist
    left_hand_dist_rew = left_hand_finger_dist

    # Total reward is: position distance + orientation alignment + action regularization + success bonus + fall penalty
    up_rew = torch.zeros_like(right_hand_dist_rew)
    up_rew = torch.where(right_hand_finger_dist < 0.5,
                    torch.where(left_hand_finger_dist < 0.5,
                                    torch.abs(door_right_handle_pos[:, 1] - door_left_handle_pos[:, 1]) * 2, up_rew), up_rew)
    reward = 2 - right_hand_dist_rew - left_hand_dist_rew + up_rew

    resets = torch.where(right_hand_finger_dist >= 1.5, torch.ones_like(reset_buf), reset_buf)
    resets = torch.where(left_hand_finger_dist >= 1.5, torch.ones_like(resets), resets)
    
    # Find out which envs hit the goal and update successes count
    successes = torch.where(successes == 0, 
                    torch.where(torch.abs(door_right_handle_pos[:, 1] - door_left_handle_pos[:, 1]) > 0.5, torch.ones_like(successes), successes), successes)

    resets = torch.where(progress_buf >= max_episode_length, torch.ones_like(resets), resets)

    goal_resets = torch.zeros_like(resets)

    cons_successes = torch.where(resets > 0, successes * resets, consecutive_successes).mean()
    goal_reach = torch.where(torch.abs(door_right_handle_pos[:, 1] - door_left_handle_pos[:, 1]) >= 0.5, 
                             torch.ones_like(successes), torch.zeros_like(successes))
    heuristic_reward, task_reward = reward, goal_reach     return heuristic_reward, task_reward, resets, goal_resets, progress_buf, successes, cons_successes
\end{minted}
\end{figure}

\begin{figure}[H]
\centering
\caption*{Bi-DexHands: \textbf{\textit{ShadowHandDoorCloseInward}}}
\begin{minted}[mathescape, breaklines,frame=single, fontsize=\scriptsize]{python}
def compute_hand_reward(
    rew_buf, reset_buf, reset_goal_buf, progress_buf, successes, consecutive_successes,
    max_episode_length: float, object_pos, object_rot, target_pos, target_rot, door_left_handle_pos, door_right_handle_pos,
    left_hand_pos, right_hand_pos, right_hand_ff_pos, right_hand_mf_pos, right_hand_rf_pos, right_hand_lf_pos, right_hand_th_pos,
    left_hand_ff_pos, left_hand_mf_pos, left_hand_rf_pos, left_hand_lf_pos, left_hand_th_pos,
    dist_reward_scale: float, rot_reward_scale: float, rot_eps: float,
    actions, action_penalty_scale: float,
    success_tolerance: float, reach_goal_bonus: float, fall_dist: float,
    fall_penalty: float, max_consecutive_successes: int, av_factor: float, ignore_z_rot: bool
):
    right_hand_finger_dist = (torch.norm(door_right_handle_pos - right_hand_ff_pos, p=2, dim=-1) + torch.norm(door_right_handle_pos - right_hand_mf_pos, p=2, dim=-1)
                            + torch.norm(door_right_handle_pos - right_hand_rf_pos, p=2, dim=-1) + torch.norm(door_right_handle_pos - right_hand_lf_pos, p=2, dim=-1) 
                            + torch.norm(door_right_handle_pos - right_hand_th_pos, p=2, dim=-1))
    left_hand_finger_dist = (torch.norm(door_left_handle_pos - left_hand_ff_pos, p=2, dim=-1) + torch.norm(door_left_handle_pos - left_hand_mf_pos, p=2, dim=-1)
                            + torch.norm(door_left_handle_pos - left_hand_rf_pos, p=2, dim=-1) + torch.norm(door_left_handle_pos - left_hand_lf_pos, p=2, dim=-1) 
                            + torch.norm(door_left_handle_pos - left_hand_th_pos, p=2, dim=-1))
    
    right_hand_dist_rew = right_hand_finger_dist
    left_hand_dist_rew = left_hand_finger_dist
    
    # Total reward is: position distance + orientation alignment + action regularization + success bonus + fall penalty
    up_rew = torch.zeros_like(right_hand_dist_rew)
    up_rew = torch.where(right_hand_finger_dist < 0.5,
                    torch.where(left_hand_finger_dist < 0.5,
                                    1 - torch.abs(door_right_handle_pos[:, 1] - door_left_handle_pos[:, 1]) * 2, up_rew), up_rew)

    reward = 2 - right_hand_dist_rew - left_hand_dist_rew + up_rew

    resets = torch.where(right_hand_finger_dist >= 1.5, torch.ones_like(reset_buf), reset_buf)
    resets = torch.where(left_hand_finger_dist >= 1.5, torch.ones_like(resets), resets)

    # Find out which envs hit the goal and update successes count
    successes = torch.where(successes == 0, 
                    torch.where(torch.abs(door_right_handle_pos[:, 1] - door_left_handle_pos[:, 1]) < 0.5, torch.ones_like(successes), successes), successes)
                    
    resets = torch.where(progress_buf >= max_episode_length, torch.ones_like(resets), resets)

    goal_resets = torch.zeros_like(resets)

    cons_successes = torch.where(resets > 0, successes * resets, consecutive_successes)
    goal_reach = torch.where(torch.abs(door_right_handle_pos[:, 1] - door_left_handle_pos[:, 1]) <= 0.5, 
                             torch.ones_like(successes), torch.zeros_like(successes))
    heuristic_reward, task_reward = reward, goal_reach     return heuristic_reward, task_reward, resets, goal_resets, progress_buf, successes, cons_successes
\end{minted}
\end{figure}

\begin{figure}[H]
\centering
\caption*{Bi-DexHands: \textbf{\textit{ShadowHandDoorCloseOutward}}}
\begin{minted}[mathescape, breaklines,frame=single, fontsize=\scriptsize]{python}
def compute_hand_reward(
    rew_buf, reset_buf, reset_goal_buf, progress_buf, successes, consecutive_successes,
    max_episode_length: float, object_pos, object_rot, target_pos, target_rot, door_left_handle_pos, door_right_handle_pos,
    left_hand_pos, right_hand_pos, right_hand_ff_pos, right_hand_mf_pos, right_hand_rf_pos, right_hand_lf_pos, right_hand_th_pos,
    left_hand_ff_pos, left_hand_mf_pos, left_hand_rf_pos, left_hand_lf_pos, left_hand_th_pos,
    dist_reward_scale: float, rot_reward_scale: float, rot_eps: float,
    actions, action_penalty_scale: float,
    success_tolerance: float, reach_goal_bonus: float, fall_dist: float,
    fall_penalty: float, max_consecutive_successes: int, av_factor: float, ignore_z_rot: bool
):
    right_hand_finger_dist = (torch.norm(door_right_handle_pos - right_hand_ff_pos, p=2, dim=-1) + torch.norm(door_right_handle_pos - right_hand_mf_pos, p=2, dim=-1)
                            + torch.norm(door_right_handle_pos - right_hand_rf_pos, p=2, dim=-1) + torch.norm(door_right_handle_pos - right_hand_lf_pos, p=2, dim=-1) 
                            + torch.norm(door_right_handle_pos - right_hand_th_pos, p=2, dim=-1))
    left_hand_finger_dist = (torch.norm(door_left_handle_pos - left_hand_ff_pos, p=2, dim=-1) + torch.norm(door_left_handle_pos - left_hand_mf_pos, p=2, dim=-1)
                            + torch.norm(door_left_handle_pos - left_hand_rf_pos, p=2, dim=-1) + torch.norm(door_left_handle_pos - left_hand_lf_pos, p=2, dim=-1) 
                            + torch.norm(door_left_handle_pos - left_hand_th_pos, p=2, dim=-1))
    
    right_hand_dist_rew = right_hand_finger_dist
    left_hand_dist_rew = left_hand_finger_dist

    # Total reward is: position distance + orientation alignment + action regularization + success bonus + fall penalty
    up_rew = torch.zeros_like(right_hand_dist_rew)
    up_rew = torch.where(right_hand_finger_dist < 0.5,
                    torch.where(left_hand_finger_dist < 0.5,
                                    1 - torch.abs(door_right_handle_pos[:, 1] - door_left_handle_pos[:, 1]) * 2, up_rew), up_rew)

    reward = 6 - right_hand_dist_rew - left_hand_dist_rew + up_rew

    resets = torch.where(right_hand_finger_dist >= 3, torch.ones_like(reset_buf), reset_buf)
    resets = torch.where(left_hand_finger_dist >= 3, torch.ones_like(resets), resets)

    # Find out which envs hit the goal and update successes count
    successes = torch.where(successes == 0, 
                    torch.where(torch.abs(door_right_handle_pos[:, 1] - door_left_handle_pos[:, 1]) < 0.5, torch.ones_like(successes), successes), successes)

    resets = torch.where(progress_buf >= max_episode_length, torch.ones_like(resets), resets)

    goal_resets = torch.zeros_like(resets)

    cons_successes = torch.where(resets > 0, successes * resets, consecutive_successes).mean()
    goal_reach = torch.where(torch.abs(door_right_handle_pos[:, 1] - door_left_handle_pos[:, 1]) <= 0.5, 
                torch.ones_like(successes), torch.zeros_like(successes))
    heuristic_reward, task_reward = reward, goal_reach     return heuristic_reward, task_reward, resets, goal_resets, progress_buf, successes, cons_successes
\end{minted}
\end{figure}

\begin{figure}[H]
\centering
\caption*{Bi-DexHands: \textbf{\textit{ShadowHandSpin}}}
\begin{minted}[mathescape, breaklines,frame=single, fontsize=\scriptsize]{python}
def compute_hand_reward(
    rew_buf, reset_buf, reset_goal_buf, progress_buf, successes, consecutive_successes,
    max_episode_length: float, object_pos, object_rot, target_pos, target_rot,
    dist_reward_scale: float, rot_reward_scale: float, rot_eps: float,
    actions, action_penalty_scale: float,
    success_tolerance: float, reach_goal_bonus: float, fall_dist: float,
    fall_penalty: float, max_consecutive_successes: int, av_factor: float, ignore_z_rot: bool
):
    # Distance from the hand to the object
    goal_dist = torch.norm(object_pos - target_pos, p=2, dim=-1)

    if ignore_z_rot:
        success_tolerance = 2.0 * success_tolerance
    
    # Orientation alignment 
    # Modified so pen is symmetrical; since we only rotate around the z axis,
    quat_diff_1 = quat_mul(object_rot, quat_conjugate(target_rot))
    rot_dist_1 = 2.0 * torch.asin(torch.clamp(torch.norm(quat_diff_1[:, 0:3], p=2, dim=-1), max=1.0))
    quat_diff_2 = quat_mul(object_rot, quat_conjugate(flip_orientation(target_rot)))
    rot_dist_2 = 2.0 * torch.asin(torch.clamp(torch.norm(quat_diff_2[:, 0:3], p=2, dim=-1), max=1.0))
    rot_dist = torch.min(rot_dist_1, rot_dist_2)

    dist_rew = goal_dist * dist_reward_scale
    rot_rew = 1.0/(torch.abs(rot_dist) + rot_eps) * rot_reward_scale

    action_penalty = torch.sum(actions ** 2, dim=-1)

    # Total reward is: position distance + orientation alignment + action regularization + success bonus + fall penalty
    reward = dist_rew + rot_rew + action_penalty * action_penalty_scale

    # Find out which envs hit the goal and update successes count
    goal_resets = torch.where(torch.abs(rot_dist) <= success_tolerance, torch.ones_like(reset_goal_buf), reset_goal_buf)
    successes = successes + goal_resets

    # Success bonus: orientation is within `success_tolerance` of goal orientation
    reward = torch.where(goal_resets == 1, reward + reach_goal_bonus, reward)

    # Fall penalty: distance to the goal is larger than a threshold
    reward = torch.where(goal_dist >= fall_dist, reward + fall_penalty, reward)

    # Check env termination conditions, including maximum success number
    resets = torch.where(goal_dist >= fall_dist, torch.ones_like(reset_buf), reset_buf)
    if max_consecutive_successes > 0:
        # Reset progress buffer on goal envs if max_consecutive_successes > 0
        progress_buf = torch.where(torch.abs(rot_dist) <= success_tolerance, torch.zeros_like(progress_buf), progress_buf)
        resets = torch.where(successes >= max_consecutive_successes, torch.ones_like(resets), resets)
    resets = torch.where(progress_buf >= max_episode_length - 1, torch.ones_like(resets), resets)

    # Apply penalty for not reaching the goal
    if max_consecutive_successes > 0:
        reward = torch.where(progress_buf >= max_episode_length - 1, reward + 0.5 * fall_penalty, reward)

    num_resets = torch.sum(resets)
    finished_cons_successes = torch.sum(successes * resets.float())

    cons_successes = torch.where(num_resets > 0, av_factor*finished_cons_successes/num_resets + (1.0 - av_factor)*consecutive_successes, consecutive_successes)
    goal_reach = torch.where(torch.abs(rot_dist) <= success_tolerance, 
                             torch.ones_like(reset_goal_buf), torch.zeros_like(reset_goal_buf))
    heuristic_reward, task_reward = reward, goal_reach     return heuristic_reward, task_reward, resets, goal_resets, progress_buf, successes, cons_successes
\end{minted}
\end{figure}

\begin{figure}[H]
\centering
\caption*{Bi-DexHands: \textbf{\textit{ShadowHandUpsideDown}}}
\begin{minted}[mathescape, breaklines,frame=single, fontsize=\scriptsize]{python}
def compute_hand_reward(
    rew_buf, reset_buf, reset_goal_buf, progress_buf, successes, consecutive_successes,
    max_episode_length: float, object_pos, object_rot, target_pos, target_rot,
    dist_reward_scale: float, rot_reward_scale: float, rot_eps: float,
    actions, action_penalty_scale: float,
    success_tolerance: float, reach_goal_bonus: float, fall_dist: float,
    fall_penalty: float, max_consecutive_successes: int, av_factor: float, ignore_z_rot: bool
):
    # Distance from the hand to the object
    goal_dist = torch.norm(object_pos - target_pos, p=2, dim=-1)

    if ignore_z_rot:
        success_tolerance = 2.0 * success_tolerance

    # Orientation alignment for the cube in hand and goal cube
    quat_diff = quat_mul(object_rot, quat_conjugate(target_rot))
    rot_dist = 2.0 * torch.asin(torch.clamp(torch.norm(quat_diff[:, 0:3], p=2, dim=-1), max=1.0))

    dist_rew = goal_dist * dist_reward_scale
    rot_rew = 1.0/(torch.abs(rot_dist) + rot_eps) * rot_reward_scale

    action_penalty = torch.sum(actions ** 2, dim=-1)

    # Total reward is: position distance + orientation alignment + action regularization + success bonus + fall penalty
    reward = dist_rew + rot_rew + action_penalty * action_penalty_scale

    # Find out which envs hit the goal and update successes count
    goal_resets = torch.where(torch.abs(rot_dist) <= success_tolerance, torch.ones_like(reset_goal_buf), reset_goal_buf)
    successes = successes + goal_resets

    # Success bonus: orientation is within `success_tolerance` of goal orientation
    reward = torch.where(goal_resets == 1, reward + reach_goal_bonus, reward)

    # Fall penalty: distance to the goal is larger than a threshold
    reward = torch.where(goal_dist >= fall_dist, reward + fall_penalty, reward)

    # Check env termination conditions, including maximum success number
    resets = torch.where(goal_dist >= fall_dist, torch.ones_like(reset_buf), reset_buf)
    if max_consecutive_successes > 0:
        # Reset progress buffer on goal envs if max_consecutive_successes > 0
        progress_buf = torch.where(torch.abs(rot_dist) <= success_tolerance, torch.zeros_like(progress_buf), progress_buf)
        resets = torch.where(successes >= max_consecutive_successes, torch.ones_like(resets), resets)
    resets = torch.where(progress_buf >= max_episode_length - 1, torch.ones_like(resets), resets)

    # Apply penalty for not reaching the goal
    if max_consecutive_successes > 0:
        reward = torch.where(progress_buf >= max_episode_length - 1, reward + 0.5 * fall_penalty, reward)

    num_resets = torch.sum(resets)
    finished_cons_successes = torch.sum(successes * resets.float())

    cons_successes = torch.where(num_resets > 0, av_factor*finished_cons_successes/num_resets + (1.0 - av_factor)*consecutive_successes, consecutive_successes)
    goal_reach = torch.where(torch.abs(rot_dist) <= success_tolerance, torch.ones_like(reset_goal_buf), torch.zeros_like(reset_goal_buf))
    heuristic_reward, task_reward = reward, goal_reach     return heuristic_reward, task_reward, resets, goal_resets, progress_buf, successes, cons_successes
\end{minted}
\end{figure}

\begin{figure}[H]
\centering
\caption*{Bi-DexHands: \textbf{\textit{ShadowHandBlockStack}}}
\begin{minted}[mathescape, breaklines,frame=single, fontsize=\scriptsize]{python}
def compute_hand_reward(
    rew_buf, reset_buf, reset_goal_buf, progress_buf, successes, consecutive_successes,
    max_episode_length: float, object_pos, object_rot, target_pos, target_rot, block_right_handle_pos, block_left_handle_pos,
    left_hand_pos, right_hand_pos, right_hand_ff_pos, right_hand_mf_pos, right_hand_rf_pos, right_hand_lf_pos, right_hand_th_pos,
    left_hand_ff_pos, left_hand_mf_pos, left_hand_rf_pos, left_hand_lf_pos, left_hand_th_pos,
    dist_reward_scale: float, rot_reward_scale: float, rot_eps: float,
    actions, action_penalty_scale: float,
    success_tolerance: float, reach_goal_bonus: float, fall_dist: float,
    fall_penalty: float, max_consecutive_successes: int, av_factor: float, ignore_z_rot: bool
):
    # Distance from the hand to the object
    stack_pos1 = target_pos.clone()
    stack_pos2 = target_pos.clone()

    stack_pos1[:, 1] -= 0.1
    stack_pos2[:, 1] -= 0.1
    stack_pos1[:, 2] += 0.05

    goal_dist1 = torch.norm(stack_pos1 - block_left_handle_pos, p=2, dim=-1)
    goal_dist2 = torch.norm(stack_pos2 - block_right_handle_pos, p=2, dim=-1)
    
    right_hand_finger_dist = (torch.norm(block_right_handle_pos - right_hand_ff_pos, p=2, dim=-1) + torch.norm(block_right_handle_pos - right_hand_mf_pos, p=2, dim=-1)
                            + torch.norm(block_right_handle_pos - right_hand_rf_pos, p=2, dim=-1) + torch.norm(block_right_handle_pos - right_hand_lf_pos, p=2, dim=-1) 
                            + torch.norm(block_right_handle_pos - right_hand_th_pos, p=2, dim=-1))
    left_hand_finger_dist = (torch.norm(block_left_handle_pos - left_hand_ff_pos, p=2, dim=-1) + torch.norm(block_left_handle_pos - left_hand_mf_pos, p=2, dim=-1)
                            + torch.norm(block_left_handle_pos - left_hand_rf_pos, p=2, dim=-1) + torch.norm(block_left_handle_pos - left_hand_lf_pos, p=2, dim=-1) 
                            + torch.norm(block_left_handle_pos - left_hand_th_pos, p=2, dim=-1))
    
    right_hand_dist_rew = right_hand_finger_dist
    left_hand_dist_rew = left_hand_finger_dist

    # Total reward is: position distance + orientation alignment + action regularization + success bonus + fall penalty
    up_rew = torch.zeros_like(right_hand_dist_rew)
    up_rew = torch.where(right_hand_finger_dist < 0.5,
                    torch.where(left_hand_finger_dist < 0.5,
                        (0.24 - goal_dist1 - goal_dist2) * 2, up_rew), up_rew)

    stack_rew = torch.zeros_like(right_hand_dist_rew)
    stack_rew = torch.where(goal_dist2 < 0.07,
                    torch.where(goal_dist1 < 0.07,
                        (0.05-torch.abs(stack_pos1[:, 2] - block_left_handle_pos[:, 2])) * 50 ,stack_rew),stack_rew)

    reward = 1.5 - right_hand_dist_rew - left_hand_dist_rew + up_rew + stack_rew

    resets = torch.where(right_hand_dist_rew <= 0, torch.ones_like(reset_buf), reset_buf)
    resets = torch.where(right_hand_finger_dist >= 0.75, torch.ones_like(resets), resets)
    resets = torch.where(left_hand_finger_dist >= 0.75, torch.ones_like(resets), resets)

    # Find out which envs hit the goal and update successes count
    successes = torch.where(successes == 0, 
                    torch.where(stack_rew > 1, torch.ones_like(successes), successes), successes)

    resets = torch.where(progress_buf >= max_episode_length, torch.ones_like(resets), resets)

    goal_resets = torch.zeros_like(resets)

    cons_successes = torch.where(resets > 0, successes * resets, consecutive_successes)
    goal_reach = torch.where(stack_rew >= 1, 
                             torch.ones_like(successes), torch.zeros_like(successes))
    heuristic_reward, task_reward = reward, goal_reach     return heuristic_reward, task_reward, resets, goal_resets, progress_buf, successes, cons_successes
\end{minted}
\end{figure}

\begin{figure}[H]
\centering
\caption*{Bi-DexHands: \textbf{\textit{ShadowHandBottleCap}}}
\begin{minted}[mathescape, breaklines,frame=single, fontsize=\scriptsize]{python}
def compute_hand_reward(
    rew_buf, reset_buf, reset_goal_buf, progress_buf, successes, consecutive_successes,
    max_episode_length: float, object_pos, object_rot, target_pos, target_rot, bottle_cap_pos, bottle_pos, bottle_cap_up,
    left_hand_pos, right_hand_pos, right_hand_ff_pos, right_hand_mf_pos, right_hand_rf_pos, right_hand_lf_pos, right_hand_th_pos,
    dist_reward_scale: float, rot_reward_scale: float, rot_eps: float,
    actions, action_penalty_scale: float,
    success_tolerance: float, reach_goal_bonus: float, fall_dist: float,
    fall_penalty: float, max_consecutive_successes: int, av_factor: float, ignore_z_rot: bool
):
    right_hand_dist = torch.norm(bottle_cap_pos - right_hand_pos, p=2, dim=-1)
    left_hand_dist = torch.norm(bottle_pos - left_hand_pos, p=2, dim=-1)

    right_hand_finger_dist = (torch.norm(bottle_cap_pos - right_hand_ff_pos, p=2, dim=-1) + torch.norm(bottle_cap_pos - right_hand_mf_pos, p=2, dim=-1)
                            + torch.norm(bottle_cap_pos - right_hand_rf_pos, p=2, dim=-1) + torch.norm(bottle_cap_pos - right_hand_lf_pos, p=2, dim=-1) 
                            + torch.norm(bottle_cap_pos - right_hand_th_pos, p=2, dim=-1))
    
    right_hand_dist_rew = right_hand_finger_dist
    left_hand_dist_rew = left_hand_dist

    # Total reward is: position distance + orientation alignment + action regularization + success bonus + fall penalty
    up_rew = torch.zeros_like(right_hand_dist_rew)

    up_rew =  torch.where(right_hand_finger_dist <= 0.3, torch.norm(bottle_cap_up - bottle_pos, p=2, dim=-1) * 30, up_rew)

    reward = 2.0 - right_hand_dist_rew - left_hand_dist_rew + up_rew

    resets = torch.where(bottle_cap_pos[:, 2] <= 0.5, torch.ones_like(reset_buf), reset_buf)
    resets = torch.where(right_hand_dist >= 0.5, torch.ones_like(resets), resets)
    resets = torch.where(left_hand_dist >= 0.2, torch.ones_like(resets), resets)

    # Find out which envs hit the goal and update successes count
    successes = torch.where(successes == 0, 
                    torch.where(torch.norm(bottle_cap_up - bottle_pos, p=2, dim=-1) > 0.03, torch.ones_like(successes), successes), successes)

    resets = torch.where(progress_buf >= max_episode_length, torch.ones_like(resets), resets)

    goal_resets = torch.zeros_like(resets)

    cons_successes = torch.where(resets > 0, successes * resets, consecutive_successes)
    goal_reach = torch.where(torch.norm(bottle_cap_up - bottle_pos, p=2, dim=-1) >= 0.03, 
                             torch.ones_like(successes), torch.zeros_like(successes))
    heuristic_reward, task_reward = reward, goal_reach     return heuristic_reward, task_reward, resets, goal_resets, progress_buf, successes, cons_successes
\end{minted}
\end{figure}

\begin{figure}[H]
\centering
\caption*{Bi-DexHands: \textbf{\textit{ShadowHandGraspAndPlace}}}
\begin{minted}[mathescape, breaklines,frame=single, fontsize=\scriptsize]{python}
def compute_hand_reward(
    rew_buf, reset_buf, reset_goal_buf, progress_buf, successes, consecutive_successes,
    max_episode_length: float, object_pos, object_rot, target_pos, target_rot, block_right_handle_pos, block_left_handle_pos,
    left_hand_pos, right_hand_pos, right_hand_ff_pos, right_hand_mf_pos, right_hand_rf_pos, right_hand_lf_pos, right_hand_th_pos,
    left_hand_ff_pos, left_hand_mf_pos, left_hand_rf_pos, left_hand_lf_pos, left_hand_th_pos,
    dist_reward_scale: float, rot_reward_scale: float, rot_eps: float,
    actions, action_penalty_scale: float,
    success_tolerance: float, reach_goal_bonus: float, fall_dist: float,
    fall_penalty: float, max_consecutive_successes: int, av_factor: float, ignore_z_rot: bool
):
    # Distance from the hand to the object
    right_hand_finger_dist = (torch.norm(block_right_handle_pos - right_hand_ff_pos, p=2, dim=-1) + torch.norm(block_right_handle_pos - right_hand_mf_pos, p=2, dim=-1)
                            + torch.norm(block_right_handle_pos - right_hand_rf_pos, p=2, dim=-1) + torch.norm(block_right_handle_pos - right_hand_lf_pos, p=2, dim=-1) 
                            + torch.norm(block_right_handle_pos - right_hand_th_pos, p=2, dim=-1))
    left_hand_finger_dist = (torch.norm(block_left_handle_pos - left_hand_ff_pos, p=2, dim=-1) + torch.norm(block_left_handle_pos - left_hand_mf_pos, p=2, dim=-1)
                            + torch.norm(block_left_handle_pos - left_hand_rf_pos, p=2, dim=-1) + torch.norm(block_left_handle_pos - left_hand_lf_pos, p=2, dim=-1) 
                            + torch.norm(block_left_handle_pos - left_hand_th_pos, p=2, dim=-1))
    
    right_hand_dist_rew = torch.exp(-10 * right_hand_finger_dist)
    left_hand_dist_rew = torch.exp(-10 * left_hand_finger_dist)

    up_rew = torch.zeros_like(right_hand_dist_rew)
    up_rew = torch.exp(-10 * torch.norm(block_right_handle_pos - block_left_handle_pos, p=2, dim=-1)) * 2
    
    reward = right_hand_dist_rew + left_hand_dist_rew + up_rew

    resets = torch.where(right_hand_dist_rew <= 0, torch.ones_like(reset_buf), reset_buf)
    resets = torch.where(right_hand_finger_dist >= 1.5, torch.ones_like(resets), resets)
    resets = torch.where(left_hand_finger_dist >= 1.5, torch.ones_like(resets), resets)

    # Find out which envs hit the goal and update successes count
    successes = torch.where(successes == 0, 
                    torch.where(torch.norm(block_right_handle_pos - block_left_handle_pos, p=2, dim=-1) < 0.2, torch.ones_like(successes), successes), successes)

    resets = torch.where(progress_buf >= max_episode_length, torch.ones_like(resets), resets)

    goal_resets = torch.zeros_like(resets)

    cons_successes = torch.where(resets > 0, successes * resets, consecutive_successes).mean()
    goal_reach = torch.where(torch.norm(block_right_handle_pos - block_left_handle_pos, p=2, dim=-1) <= 0.2, 
                             torch.ones_like(successes), torch.zeros_like(successes))
    heuristic_reward, task_reward = reward, goal_reach     return heuristic_reward, task_reward, resets, goal_resets, progress_buf, successes, cons_successes
\end{minted}
\end{figure}

\begin{figure}[H]
\centering
\caption*{Bi-DexHands: \textbf{\textit{ShadowHandKettle}}}
\begin{minted}[mathescape, breaklines,frame=single, fontsize=\scriptsize]{python}
def compute_hand_reward(
    rew_buf, reset_buf, reset_goal_buf, progress_buf, successes, consecutive_successes,
    max_episode_length: float, object_pos, object_rot, target_pos, target_rot, kettle_handle_pos, bucket_handle_pos, kettle_spout_pos,
    left_hand_pos, right_hand_pos, right_hand_ff_pos, right_hand_mf_pos, right_hand_rf_pos, right_hand_lf_pos, right_hand_th_pos,
    left_hand_ff_pos, left_hand_mf_pos, left_hand_rf_pos, left_hand_lf_pos, left_hand_th_pos,
    dist_reward_scale: float, rot_reward_scale: float, rot_eps: float,
    actions, action_penalty_scale: float,
    success_tolerance: float, reach_goal_bonus: float, fall_dist: float,
    fall_penalty: float, max_consecutive_successes: int, av_factor: float, ignore_z_rot: bool
):
    # Distance from the hand to the object
    right_hand_finger_dist = (torch.norm(kettle_handle_pos - right_hand_ff_pos, p=2, dim=-1) + torch.norm(kettle_handle_pos - right_hand_mf_pos, p=2, dim=-1)
                            + torch.norm(kettle_handle_pos - right_hand_rf_pos, p=2, dim=-1) + torch.norm(kettle_handle_pos - right_hand_lf_pos, p=2, dim=-1) 
                            + torch.norm(kettle_handle_pos - right_hand_th_pos, p=2, dim=-1))
    left_hand_finger_dist = (torch.norm(bucket_handle_pos - left_hand_ff_pos, p=2, dim=-1) + torch.norm(bucket_handle_pos - left_hand_mf_pos, p=2, dim=-1)
                            + torch.norm(bucket_handle_pos - left_hand_rf_pos, p=2, dim=-1) + torch.norm(bucket_handle_pos - left_hand_lf_pos, p=2, dim=-1) 
                            + torch.norm(bucket_handle_pos - left_hand_th_pos, p=2, dim=-1))

    right_hand_dist_rew = right_hand_finger_dist
    left_hand_dist_rew = left_hand_finger_dist

    # Total reward is: position distance + orientation alignment + action regularization + success bonus + fall penalty
    up_rew = torch.zeros_like(right_hand_dist_rew)
    up_rew = torch.where(right_hand_finger_dist < 0.7,
                    torch.where(left_hand_finger_dist < 0.7,
                                    0.5 - torch.norm(bucket_handle_pos - kettle_spout_pos, p=2, dim=-1) * 2, up_rew), up_rew)

    reward = 1 + up_rew - right_hand_dist_rew - left_hand_dist_rew

    resets = torch.where(bucket_handle_pos[:, 2] <= 0.2, torch.ones_like(reset_buf), reset_buf)
    
    # Find out which envs hit the goal and update successes count
    successes = torch.where(successes == 0, 
                    torch.where(torch.norm(bucket_handle_pos - kettle_spout_pos, p=2, dim=-1) < 0.05, torch.ones_like(successes), successes), successes)

    resets = torch.where(progress_buf >= max_episode_length, torch.ones_like(resets), resets)

    goal_resets = torch.zeros_like(resets)

    cons_successes = torch.where(resets > 0, successes * resets, consecutive_successes).mean()
    goal_reach = torch.where(torch.norm(bucket_handle_pos - kettle_spout_pos, p=2, dim=-1) <= 0.05, 
                             torch.ones_like(successes), torch.zeros_like(successes))    
    heuristic_reward, task_reward = reward, goal_reach     return heuristic_reward, task_reward, resets, goal_resets, progress_buf, successes, cons_successes
\end{minted}
\end{figure}

\begin{figure}[H]
\centering
\caption*{Bi-DexHands: \textbf{\textit{ShadowHandPen}}}
\begin{minted}[mathescape, breaklines,frame=single, fontsize=\scriptsize]{python}
def compute_hand_reward(
    rew_buf, reset_buf, reset_goal_buf, progress_buf, successes, consecutive_successes,
    max_episode_length: float, object_pos, object_rot, target_pos, target_rot, pen_right_handle_pos, pen_left_handle_pos,
    left_hand_pos, right_hand_pos, right_hand_ff_pos, right_hand_mf_pos, right_hand_rf_pos, right_hand_lf_pos, right_hand_th_pos,
    left_hand_ff_pos, left_hand_mf_pos, left_hand_rf_pos, left_hand_lf_pos, left_hand_th_pos,
    dist_reward_scale: float, rot_reward_scale: float, rot_eps: float,
    actions, action_penalty_scale: float,
    success_tolerance: float, reach_goal_bonus: float, fall_dist: float,
    fall_penalty: float, max_consecutive_successes: int, av_factor: float, ignore_z_rot: bool
):
    # Distance from the hand to the object
    right_hand_finger_dist = (torch.norm(pen_right_handle_pos - right_hand_ff_pos, p=2, dim=-1) + torch.norm(pen_right_handle_pos - right_hand_mf_pos, p=2, dim=-1)
                            + torch.norm(pen_right_handle_pos - right_hand_rf_pos, p=2, dim=-1) + torch.norm(pen_right_handle_pos - right_hand_lf_pos, p=2, dim=-1) 
                            + torch.norm(pen_right_handle_pos - right_hand_th_pos, p=2, dim=-1))
    left_hand_finger_dist = (torch.norm(pen_left_handle_pos - left_hand_ff_pos, p=2, dim=-1) + torch.norm(pen_left_handle_pos - left_hand_mf_pos, p=2, dim=-1)
                            + torch.norm(pen_left_handle_pos - left_hand_rf_pos, p=2, dim=-1) + torch.norm(pen_left_handle_pos - left_hand_lf_pos, p=2, dim=-1) 
                            + torch.norm(pen_left_handle_pos - left_hand_th_pos, p=2, dim=-1))
  
    right_hand_dist_rew = torch.exp(-10 * right_hand_finger_dist)
    left_hand_dist_rew = torch.exp(-10 * left_hand_finger_dist)

    # Total reward is: position distance + orientation alignment + action regularization + success bonus + fall penalty
    up_rew = torch.zeros_like(right_hand_dist_rew)
    up_rew = torch.where(right_hand_finger_dist < 0.75,
                    torch.where(left_hand_finger_dist < 0.75,
                        torch.norm(pen_right_handle_pos - pen_left_handle_pos, p=2, dim=-1) * 5 - 0.8, up_rew), up_rew)

    reward = up_rew + right_hand_dist_rew + left_hand_dist_rew

    resets = torch.where(right_hand_dist_rew <= 0, torch.ones_like(reset_buf), reset_buf)
    resets = torch.where(right_hand_finger_dist >= 1.5, torch.ones_like(resets), resets)
    resets = torch.where(left_hand_finger_dist >= 1.5, torch.ones_like(resets), resets)

    # Find out which envs hit the goal and update successes count
    successes = torch.where(successes == 0, 
                    torch.where(torch.norm(pen_right_handle_pos - pen_left_handle_pos, p=2, dim=-1) * 5 > 1.5, torch.ones_like(successes), successes), successes)

    resets = torch.where(progress_buf >= max_episode_length, torch.ones_like(resets), resets)

    goal_resets = torch.zeros_like(resets)

    cons_successes = torch.where(resets > 0, successes * resets, consecutive_successes)
    goal_reach = torch.where(torch.norm(pen_right_handle_pos - pen_left_handle_pos, p=2, dim=-1) * 5 >= 1.5, 
                             torch.ones_like(successes), torch.zeros_like(successes))
    heuristic_reward, task_reward = reward, goal_reach     return heuristic_reward, task_reward, resets, goal_resets, progress_buf, successes, cons_successes
\end{minted}
\end{figure}

\begin{figure}[H]
\centering
\caption*{Bi-DexHands: \textbf{\textit{ShadowHandPushBlock}}}
\begin{minted}[mathescape, breaklines,frame=single, fontsize=\scriptsize]{python}
def compute_hand_reward(
    rew_buf, reset_buf, reset_goal_buf, progress_buf, successes, consecutive_successes,
    max_episode_length: float, object_pos, object_rot, left_target_pos, left_target_rot, right_target_pos, right_target_rot, block_right_handle_pos, block_left_handle_pos,
    left_hand_pos, right_hand_pos, right_hand_ff_pos, right_hand_mf_pos, right_hand_rf_pos, right_hand_lf_pos, right_hand_th_pos,
    left_hand_ff_pos, left_hand_mf_pos, left_hand_rf_pos, left_hand_lf_pos, left_hand_th_pos,
    dist_reward_scale: float, rot_reward_scale: float, rot_eps: float,
    actions, action_penalty_scale: float,
    success_tolerance: float, reach_goal_bonus: float, fall_dist: float,
    fall_penalty: float, max_consecutive_successes: int, av_factor: float, ignore_z_rot: bool
):
    # Distance from the hand to the object
    left_goal_dist = torch.norm(left_target_pos - block_left_handle_pos, p=2, dim=-1)
    right_goal_dist = torch.norm(right_target_pos - block_right_handle_pos, p=2, dim=-1)
    
    right_hand_finger_dist = (torch.norm(block_right_handle_pos - right_hand_ff_pos, p=2, dim=-1) + torch.norm(block_right_handle_pos - right_hand_mf_pos, p=2, dim=-1)
                            + torch.norm(block_right_handle_pos - right_hand_rf_pos, p=2, dim=-1) + torch.norm(block_right_handle_pos - right_hand_lf_pos, p=2, dim=-1) 
                            + torch.norm(block_right_handle_pos - right_hand_th_pos, p=2, dim=-1))
    left_hand_finger_dist = (torch.norm(block_left_handle_pos - left_hand_ff_pos, p=2, dim=-1) + torch.norm(block_left_handle_pos - left_hand_mf_pos, p=2, dim=-1)
                            + torch.norm(block_left_handle_pos - left_hand_rf_pos, p=2, dim=-1) + torch.norm(block_left_handle_pos - left_hand_lf_pos, p=2, dim=-1) 
                            + torch.norm(block_left_handle_pos - left_hand_th_pos, p=2, dim=-1))
    
    right_hand_dist_rew = 1.2-1*right_hand_finger_dist
    left_hand_dist_rew = 1.2-1*left_hand_finger_dist
    
    # Total reward is: position distance + orientation alignment + action regularization + success bonus + fall penalty
    up_rew = torch.zeros_like(right_hand_dist_rew)
    up_rew = 5 - 5*left_goal_dist - 5*right_goal_dist
    
    reward = right_hand_dist_rew + left_hand_dist_rew + up_rew

    resets = torch.where(right_hand_finger_dist >= 1.2, torch.ones_like(reset_buf), reset_buf)
    resets = torch.where(left_hand_finger_dist >= 1.2, torch.ones_like(resets), resets)

    # Find out which envs hit the goal and update successes count
    successes = torch.where(successes == 0, 
                    torch.where(torch.abs(left_goal_dist) <= 0.1, 
                        torch.where(torch.abs(right_goal_dist) <= 0.1, torch.ones_like(successes), torch.ones_like(successes) * 0.5), successes), successes)

    resets = torch.where(progress_buf >= max_episode_length, torch.ones_like(resets), resets)

    goal_resets = torch.zeros_like(resets)

    cons_successes = torch.where(resets > 0, successes * resets, consecutive_successes).mean()
    goal_reach = 0.5 * (torch.where(torch.abs(left_goal_dist) <= 0.1, 
                                    torch.ones_like(successes), torch.zeros_like(successes)) \
                        + torch.where(torch.abs(right_goal_dist) <= 0.1, 
                                      torch.ones_like(successes), torch.zeros_like(successes)))
    heuristic_reward, task_reward = reward, goal_reach     return heuristic_reward, task_reward, resets, goal_resets, progress_buf, successes, cons_successes
\end{minted}
\end{figure}

\begin{figure}[H]
\centering
\caption*{Bi-DexHands: \textbf{\textit{ShadowHandReOrientation}}}
\begin{minted}[mathescape, breaklines,frame=single, fontsize=\scriptsize]{python}
def compute_hand_reward(
    rew_buf, reset_buf, reset_goal_buf, progress_buf, successes, consecutive_successes,
    max_episode_length: float, object_pos, object_rot, target_pos, target_rot, object_another_pos, object_another_rot, target_another_pos, target_another_rot,
    dist_reward_scale: float, rot_reward_scale: float, rot_eps: float,
    actions, action_penalty_scale: float,
    success_tolerance: float, reach_goal_bonus: float, fall_dist: float,
    fall_penalty: float, max_consecutive_successes: int, av_factor: float, ignore_z_rot: bool
):
    # Distance from the hand to the object
    goal_dist = torch.norm(target_pos - object_pos, p=2, dim=-1)
    if ignore_z_rot:
        success_tolerance = 2.0 * success_tolerance

    goal_another_dist = torch.norm(target_another_pos - object_another_pos, p=2, dim=-1)
    if ignore_z_rot:
        success_tolerance = 2.0 * success_tolerance

    # Orientation alignment for the cube in hand and goal cube
    quat_diff = quat_mul(object_rot, quat_conjugate(target_rot))
    rot_dist = 2.0 * torch.asin(torch.clamp(torch.norm(quat_diff[:, 0:3], p=2, dim=-1), max=1.0))

    quat_another_diff = quat_mul(object_another_rot, quat_conjugate(target_another_rot))
    rot_another_dist = 2.0 * torch.asin(torch.clamp(torch.norm(quat_another_diff[:, 0:3], p=2, dim=-1), max=1.0))

    dist_rew = goal_dist * dist_reward_scale + goal_another_dist * dist_reward_scale
    rot_rew = 1.0/(torch.abs(rot_dist) + rot_eps) * rot_reward_scale + 1.0/(torch.abs(rot_another_dist) + rot_eps) * rot_reward_scale

    action_penalty = torch.sum(actions ** 2, dim=-1)

    # Total reward is: position distance + orientation alignment + action regularization + success bonus + fall penalty
    reward = dist_rew + rot_rew + action_penalty * action_penalty_scale

    # Find out which envs hit the goal and update successes count
    goal_resets = torch.where(torch.abs(rot_dist) < 0.1, torch.ones_like(reset_goal_buf), reset_goal_buf)
    goal_resets = torch.where(torch.abs(rot_another_dist) < 0.1, torch.ones_like(reset_goal_buf), reset_goal_buf)

    successes = successes + goal_resets

    # Success bonus: orientation is within `success_tolerance` of goal orientation
    reward = torch.where(goal_resets == 1, reward + reach_goal_bonus, reward)

    # Fall penalty: distance to the goal is larger than a threashold
    reward = torch.where(object_pos[:, 2] <= 0.2, reward + fall_penalty, reward)
    reward = torch.where(object_another_pos[:, 2] <= 0.2, reward + fall_penalty, reward)

    # Check env termination conditions, including maximum success number
    resets = torch.where(object_pos[:, 2] <= 0.2, torch.ones_like(reset_buf), reset_buf)
    resets = torch.where(object_another_pos[:, 2] <= 0.2, torch.ones_like(reset_buf), resets)

    if max_consecutive_successes > 0:
        # Reset progress buffer on goal envs if max_consecutive_successes > 0
        progress_buf = torch.where(torch.abs(rot_dist) <= success_tolerance, torch.zeros_like(progress_buf), progress_buf)
        resets = torch.where(successes >= max_consecutive_successes, torch.ones_like(resets), resets)
    resets = torch.where(progress_buf >= max_episode_length, torch.ones_like(resets), resets)

    # Apply penalty for not reaching the goal
    if max_consecutive_successes > 0:
        reward = torch.where(progress_buf >= max_episode_length, reward + 0.5 * fall_penalty, reward)

    num_resets = torch.sum(resets)
    finished_cons_successes = torch.sum(successes * resets.float())

    cons_successes = torch.where(num_resets > 0, av_factor*finished_cons_successes/num_resets + (1.0 - av_factor)*consecutive_successes, consecutive_successes)
    goal_reach = 0.5 * (torch.where(torch.abs(rot_dist) <= 0.1, torch.ones_like(reset_goal_buf), torch.zeros_like(reset_goal_buf)) \
            + torch.where(torch.abs(rot_another_dist) <= 0.1, torch.ones_like(reset_goal_buf), torch.zeros_like(reset_goal_buf)))
    heuristic_reward, task_reward = reward, goal_reach     return heuristic_reward, task_reward, resets, goal_resets, progress_buf, successes, cons_successes
\end{minted}
\end{figure}

\begin{figure}[H]
\centering
\caption*{Bi-DexHands: \textbf{\textit{ShadowHandScissors}}}
\begin{minted}[mathescape, breaklines,frame=single, fontsize=\scriptsize]{python}
def compute_hand_reward(
    rew_buf, reset_buf, reset_goal_buf, progress_buf, successes, consecutive_successes,
    max_episode_length: float, object_pos, object_rot, target_pos, target_rot, scissors_right_handle_pos, scissors_left_handle_pos, object_dof_pos,
    left_hand_pos, right_hand_pos, right_hand_ff_pos, right_hand_mf_pos, right_hand_rf_pos, right_hand_lf_pos, right_hand_th_pos,
    left_hand_ff_pos, left_hand_mf_pos, left_hand_rf_pos, left_hand_lf_pos, left_hand_th_pos,
    dist_reward_scale: float, rot_reward_scale: float, rot_eps: float,
    actions, action_penalty_scale: float,
    success_tolerance: float, reach_goal_bonus: float, fall_dist: float,
    fall_penalty: float, max_consecutive_successes: int, av_factor: float, ignore_z_rot: bool
):
    # Distance from the hand to the object
    right_hand_finger_dist = (torch.norm(scissors_right_handle_pos - right_hand_ff_pos, p=2, dim=-1) + torch.norm(scissors_right_handle_pos - right_hand_mf_pos, p=2, dim=-1)
                            + torch.norm(scissors_right_handle_pos - right_hand_rf_pos, p=2, dim=-1) + torch.norm(scissors_right_handle_pos - right_hand_lf_pos, p=2, dim=-1) 
                            + torch.norm(scissors_right_handle_pos - right_hand_th_pos, p=2, dim=-1))
    left_hand_finger_dist = (torch.norm(scissors_left_handle_pos - left_hand_ff_pos, p=2, dim=-1) + torch.norm(scissors_left_handle_pos - left_hand_mf_pos, p=2, dim=-1)
                            + torch.norm(scissors_left_handle_pos - left_hand_rf_pos, p=2, dim=-1) + torch.norm(scissors_left_handle_pos - left_hand_lf_pos, p=2, dim=-1) 
                            + torch.norm(scissors_left_handle_pos - left_hand_th_pos, p=2, dim=-1))
    
    right_hand_dist_rew = right_hand_finger_dist
    left_hand_dist_rew = left_hand_finger_dist

    # Total reward is: position distance + orientation alignment + action regularization + success bonus + fall penalty
    up_rew = torch.zeros_like(right_hand_dist_rew)
    up_rew = torch.where(right_hand_finger_dist < 0.7,
                    torch.where(left_hand_finger_dist < 0.7,
                        (0.59 + object_dof_pos[:, 0]) * 5, up_rew), up_rew)
    
    reward = 2 + up_rew - right_hand_dist_rew - left_hand_dist_rew

    resets = torch.where(up_rew < -0.5, torch.ones_like(reset_buf), reset_buf)
    resets = torch.where(right_hand_finger_dist >= 1.75, torch.ones_like(resets), resets)
    resets = torch.where(left_hand_finger_dist >= 1.75, torch.ones_like(resets), resets)
    
    # Find out which envs hit the goal and update successes count
    resets = torch.where(progress_buf >= max_episode_length, torch.ones_like(resets), resets)

    successes = torch.where(successes == 0, 
                    torch.where(object_dof_pos[:, 0] > -0.3, torch.ones_like(successes), successes), successes)

    goal_resets = torch.zeros_like(resets)

    cons_successes = torch.where(resets > 0, successes * resets, consecutive_successes).mean()
    goal_reach = torch.where(object_dof_pos[:, 0] >= -0.3, 
                             torch.ones_like(successes), torch.zeros_like(successes))
    heuristic_reward, task_reward = reward, goal_reach     return heuristic_reward, task_reward, resets, goal_resets, progress_buf, successes, cons_successes
\end{minted}
\end{figure}

\begin{figure}[H]
\centering
\caption*{Bi-DexHands: \textbf{\textit{ShadowHandSwingCup}}}
\begin{minted}[mathescape, breaklines,frame=single, fontsize=\scriptsize]{python}
def compute_hand_reward(
    rew_buf, reset_buf, reset_goal_buf, progress_buf, successes, consecutive_successes,
    max_episode_length: float, object_pos, object_rot, target_pos, target_rot, cup_right_handle_pos, cup_left_handle_pos,
    left_hand_pos, right_hand_pos, right_hand_ff_pos, right_hand_mf_pos, right_hand_rf_pos, right_hand_lf_pos, right_hand_th_pos,
    left_hand_ff_pos, left_hand_mf_pos, left_hand_rf_pos, left_hand_lf_pos, left_hand_th_pos,
    dist_reward_scale: float, rot_reward_scale: float, rot_eps: float,
    actions, action_penalty_scale: float,
    success_tolerance: float, reach_goal_bonus: float, fall_dist: float,
    fall_penalty: float, max_consecutive_successes: int, av_factor: float, ignore_z_rot: bool
):
    # Distance from the hand to the object
    right_hand_finger_dist = (torch.norm(cup_right_handle_pos - right_hand_ff_pos, p=2, dim=-1) + torch.norm(cup_right_handle_pos - right_hand_mf_pos, p=2, dim=-1)
                            + torch.norm(cup_right_handle_pos - right_hand_rf_pos, p=2, dim=-1) + torch.norm(cup_right_handle_pos - right_hand_lf_pos, p=2, dim=-1) 
                            + torch.norm(cup_right_handle_pos - right_hand_th_pos, p=2, dim=-1))
    left_hand_finger_dist = (torch.norm(cup_left_handle_pos - left_hand_ff_pos, p=2, dim=-1) + torch.norm(cup_left_handle_pos - left_hand_mf_pos, p=2, dim=-1)
                            + torch.norm(cup_left_handle_pos - left_hand_rf_pos, p=2, dim=-1) + torch.norm(cup_left_handle_pos - left_hand_lf_pos, p=2, dim=-1) 
                            + torch.norm(cup_left_handle_pos - left_hand_th_pos, p=2, dim=-1))
    
    # Orientation alignment for the cube in hand and goal cube
    quat_diff = quat_mul(object_rot, quat_conjugate(target_rot))
    rot_dist = 2.0 * torch.asin(torch.clamp(torch.norm(quat_diff[:, 0:3], p=2, dim=-1), max=1.0))

    right_hand_dist_rew = right_hand_finger_dist
    left_hand_dist_rew = left_hand_finger_dist

    rot_rew = 1.0/(torch.abs(rot_dist) + rot_eps) * rot_reward_scale - 1

    # Total reward is: position distance + orientation alignment + action regularization + success bonus + fall penalty
    up_rew = torch.zeros_like(rot_rew)
    up_rew = torch.where(right_hand_finger_dist < 0.4,
                        torch.where(left_hand_finger_dist < 0.4,
                                        rot_rew, up_rew), up_rew)
        
    reward = - right_hand_dist_rew - left_hand_dist_rew + up_rew

    resets = torch.where(object_pos[:, 2] <= 0.3, torch.ones_like(reset_buf), reset_buf)
    
    # Find out which envs hit the goal and update successes count
    successes = torch.where(successes == 0, 
                    torch.where(rot_dist < 0.785, torch.ones_like(successes), successes), successes)


    resets = torch.where(progress_buf >= max_episode_length, torch.ones_like(resets), resets)

    goal_resets = torch.zeros_like(resets)

    cons_successes = torch.where(resets > 0, successes * resets, consecutive_successes).mean()
    goal_reach = torch.where(rot_dist <= 0.785, torch.ones_like(successes), torch.zeros_like(successes))
    heuristic_reward, task_reward = reward, goal_reach     return heuristic_reward, task_reward, resets, goal_resets, progress_buf, successes, cons_successes
\end{minted}
\end{figure}

\begin{figure}[H]
\centering
\caption*{Bi-DexHands: \textbf{\textit{ShadowHandSwitch}}}
\begin{minted}[mathescape, breaklines,frame=single, fontsize=\scriptsize]{python}
def compute_hand_reward(
    rew_buf, reset_buf, reset_goal_buf, progress_buf, successes, consecutive_successes,
    max_episode_length: float, object_pos, object_rot, target_pos, target_rot, switch_right_handle_pos, switch_left_handle_pos,
    left_hand_pos, right_hand_pos, right_hand_ff_pos, right_hand_mf_pos, right_hand_rf_pos, right_hand_lf_pos, right_hand_th_pos,
    left_hand_ff_pos, left_hand_mf_pos, left_hand_rf_pos, left_hand_lf_pos, left_hand_th_pos,
    dist_reward_scale: float, rot_reward_scale: float, rot_eps: float,
    actions, action_penalty_scale: float,
    success_tolerance: float, reach_goal_bonus: float, fall_dist: float,
    fall_penalty: float, max_consecutive_successes: int, av_factor: float, ignore_z_rot: bool
):
    # Distance from the hand to the object
    right_hand_finger_dist = (torch.norm(switch_right_handle_pos - right_hand_ff_pos, p=2, dim=-1) + torch.norm(switch_right_handle_pos - right_hand_mf_pos, p=2, dim=-1)
                            + torch.norm(switch_right_handle_pos - right_hand_rf_pos, p=2, dim=-1) + torch.norm(switch_right_handle_pos - right_hand_lf_pos, p=2, dim=-1) 
                            + torch.norm(switch_right_handle_pos - right_hand_th_pos, p=2, dim=-1))
    left_hand_finger_dist = (torch.norm(switch_left_handle_pos - left_hand_ff_pos, p=2, dim=-1) + torch.norm(switch_left_handle_pos - left_hand_mf_pos, p=2, dim=-1)
                            + torch.norm(switch_left_handle_pos - left_hand_rf_pos, p=2, dim=-1) + torch.norm(switch_left_handle_pos - left_hand_lf_pos, p=2, dim=-1) 
                            + torch.norm(switch_left_handle_pos - left_hand_th_pos, p=2, dim=-1))

    right_hand_dist_rew = right_hand_finger_dist
    left_hand_dist_rew = left_hand_finger_dist

    # Total reward is: position distance + orientation alignment + action regularization + success bonus + fall penalty
    up_rew = torch.zeros_like(right_hand_dist_rew)
    up_rew = (1.4-(switch_right_handle_pos[:, 2] + switch_left_handle_pos[:, 2])) * 50

    reward = 2 - right_hand_dist_rew - left_hand_dist_rew + up_rew

    resets = torch.where(right_hand_dist_rew <= 0, torch.ones_like(reset_buf), reset_buf)

    # Find out which envs hit the goal and update successes count
    successes = torch.where(successes == 0, 
                    torch.where(1.4-(switch_right_handle_pos[:, 2] + switch_left_handle_pos[:, 2]) > 0.05, torch.ones_like(successes), successes), successes)

    resets = torch.where(progress_buf >= max_episode_length, torch.ones_like(resets), resets)

    goal_resets = torch.zeros_like(resets)

    cons_successes = torch.where(resets > 0, successes * resets, consecutive_successes).mean()
    goal_reach = torch.where(1.4 - (switch_right_handle_pos[:, 2] + switch_left_handle_pos[:, 2]) >= 0.05, 
                             torch.ones_like(successes), torch.zeros_like(successes))
    heuristic_reward, task_reward = reward, goal_reach     return heuristic_reward, task_reward, resets, goal_resets, progress_buf, successes, cons_successes
\end{minted}
\end{figure}

\newpage
\section*{NeurIPS Paper Checklist}
\begin{enumerate}

\item {\bf Claims}
    \item[] Question: Do the main claims made in the abstract and introduction accurately reflect the paper's contributions and scope?
    \item[] Answer: \answerYes{} %
    \item[] Justification: This paper aims to explore alternatives for improving task performance in finite data settings using heuristic signals. Our experiments on robotic locomotion, helicopter, and manipulation tasks demonstrate that this method consistently improves performance, regardless of the general effectiveness of the heuristic signals. We are confident that our abstract and introduction sections accurately reflect the paper's contributions and scope.

\item {\bf Limitations}
    \item[] Question: Does the paper discuss the limitations of the work performed by the authors?
    \item[] Answer: \answerYes %
    \item[] Justification: The limitations of our approach are illustrated in Section~\ref{sec:discussion}.

\item {\bf Theory Assumptions and Proofs}
    \item[] Question: For each theoretical result, does the paper provide the full set of assumptions and a complete (and correct) proof?
    \item[] Answer: \answerYes{} %
    \item[] Justification: We have shown our derivation details and limitations in Section~\ref{app:derivation} and Section~\ref{sec:discussion}.

    \item {\bf Experimental Result Reproducibility}
    \item[] Question: Does the paper fully disclose all the information needed to reproduce the main experimental results of the paper to the extent that it affects the main claims and/or conclusions of the paper (regardless of whether the code and data are provided or not)?
    \item[] Answer: \answerYes{} %
    \item[] Justification: We have provided detailed derivation and implementation descriptions (including the simulation environments, hyperparameters, and reward definitions for training and evaluations) in our Appendix section. Additionally, we have also provided our source code in our Supplementary Materials.

\item {\bf Open access to data and code}
    \item[] Question: Does the paper provide open access to the data and code, with sufficient instructions to faithfully reproduce the main experimental results, as described in supplemental material?
    \item[] Answer: \answerYes{} %
    \item[] Justification: We have provided detailed implementation descriptions (including the simulation environments, hyperparameters, and reward definitions for training and evaluations) in our Appendix section. Additionally, we have also provided our source code in our Supplementary Materials. The adopted simulation environments are all well-known and available online.

\item {\bf Experimental Setting/Details}
    \item[] Question: Does the paper specify all the training and test details (e.g., data splits, hyperparameters, how they were chosen, type of optimizer, etc.) necessary to understand the results?
    \item[] Answer: \answerYes{} %
    \item[] Justification: We have provided detailed implementation descriptions (including the simulation environments, hyperparameters, and reward definitions for training and evaluations) in the experiment and Appendix sections. Additionally, we have also provided our source code in our Supplementary Materials, including all the training and environment configurations.

\item {\bf Experiment Statistical Significance}
    \item[] Question: Does the paper report error bars suitably and correctly defined or other appropriate information about the statistical significance of the experiments?
    \item[] Answer: \answerYes{} %
    \item[] Justification: All of our experimental results were obtained by 5 random seeds. We have provided the mean and standard deviation for our experimental results.

\item {\bf Experiments Compute Resources}
    \item[] Question: For each experiment, does the paper provide sufficient information on the computer resources (type of compute workers, memory, time of execution) needed to reproduce the experiments?
    \item[] Answer: \answerNo{} %
    \item[] Justification: But each training procedure can be performed on a single GeForce RTX 2080 Ti device. The required computational resources for all the simulation benchmarks are listed on their respective websites.
    
\item {\bf Code Of Ethics}
    \item[] Question: Does the research conducted in the paper conform, in every respect, with the NeurIPS Code of Ethics \url{https://neurips.cc/public/EthicsGuidelines}?
    \item[] Answer: \answerYes{} %
    \item[] Justification: We have carefully examined the ethical guidelines and verified that our work fully adheres to all the principles and requirements.

\item {\bf Broader Impacts}
    \item[] Question: Does the paper discuss both potential positive societal impacts and negative societal impacts of the work performed?
    \item[] Answer: \answerNA{} %
    \item[] Justification: The paper does not discuss both potential positive societal impacts and negative societal impacts of the work performed.
    
\item {\bf Safeguards}
    \item[] Question: Does the paper describe safeguards that have been put in place for responsible release of data or models that have a high risk for misuse (e.g., pretrained language models, image generators, or scraped datasets)?
    \item[] Answer: \answerNA{} %
    \item[] Justification: This paper has no such risks.

\item {\bf Licenses for existing assets}
    \item[] Question: Are the creators or original owners of assets (e.g., code, data, models), used in the paper, properly credited and are the license and terms of use explicitly mentioned and properly respected?
    \item[] Answer: \answerYes{} %
    \item[] Justification: Throughout this paper, we have provided proper citations and references for all utilized repositories, benchmark simulations, and models/algorithms to uphold transparency and ensure appropriate attribution.

\item {\bf New Assets}
    \item[] Question: Are new assets introduced in the paper well documented and is the documentation provided alongside the assets?
    \item[] Answer: \answerYes{} %
    \item[] Justification: We will furnish comprehensive documentation for our released code, elucidating its usage and providing information about the original source. Additionally, we have ensured that any code modified from external sources is subject to licenses that permit modification and redistribution.

\item {\bf Crowdsourcing and Research with Human Subjects}
    \item[] Question: For crowdsourcing experiments and research with human subjects, does the paper include the full text of instructions given to participants and screenshots, if applicable, as well as details about compensation (if any)? 
    \item[] Answer: \answerNA{} %
    \item[] Justification: No, but we engaged 12 participants in devising reward functions as part of the experiments detailed in Section~\ref{subsec:exp:application}.

\item {\bf Institutional Review Board (IRB) Approvals or Equivalent for Research with Human Subjects}
    \item[] Question: Does the paper describe potential risks incurred by study participants, whether such risks were disclosed to the subjects, and whether Institutional Review Board (IRB) approvals (or an equivalent approval/review based on the requirements of your country or institution) were obtained?
    \item[] Answer: \answerNA{} %
    \item[] Justification: This paper does not involve crowdsourcing nor research with human subjects.

\end{enumerate}

\end{document}